\documentclass[10pt,journal,compsoc]{IEEEtran}
\ifCLASSINFOpdf
   \usepackage[pdftex]{graphicx}
   \graphicspath{{../pdf/}{../jpeg/}}
   \DeclareGraphicsExtensions{.pdf,.jpeg,.png}
\else
   \usepackage[dvips]{graphicx}
   \graphicspath{{../eps/}}
   \DeclareGraphicsExtensions{.eps}
\fi
\usepackage{amsmath,amssymb,amsfonts}
\usepackage{float}
\usepackage[T1]{fontenc}
\usepackage{stfloats}
\usepackage{multirow}
\usepackage{multicol}
\usepackage{booktabs}
\usepackage[colorlinks]{hyperref}
\usepackage{makecell}
\usepackage{colortbl}
\usepackage{arydshln}
\usepackage{bm}
\usepackage{diagbox}
\usepackage[numbers,sort&compress]{natbib}
\DeclareMathAlphabet{\mymathbb}{U}{BOONDOX-ds}{m}{n}
\makeatletter
\newcommand*\bigcdot{\mathpalette\bigcdot@{.5}}
\newcommand*\bigcdot@[2]{\mathbin{\vcenter{\hbox{\scalebox{#2}{$\m@th#1\bullet$}}}}}
\makeatother
\newcommand\onedot{.}
\newcommand\eg{\emph{e.g}\onedot} 
\newcommand\ie{\emph{i.e}\onedot}

\newcommand\etal{\emph{et al}\onedot}

\newcommand{\midsepremove}{\aboverulesep=0pt \belowrulesep=0pt}
\usepackage{xcolor}
\newcommand{\graycell}[1]{
  \setlength{\fboxsep}{2pt}
  \colorbox{gray!20}{\makebox[3.5em][r]{#1}}
}

\definecolor{change}{rgb}{0,0,1}

\begin{document}
\title{Revisiting Face Forgery Detection: From Facial Representation to Forgery Detection}
\author{Zonghui~Guo,~\IEEEmembership{Member,~IEEE},
        Yingjie~Liu,~\IEEEmembership{Student Member,~IEEE},
        Jie~Zhang,~\IEEEmembership{Member,~IEEE},\\
        Haiyong~Zheng,~\IEEEmembership{Senior Member,~IEEE},
        and~Shiguang~Shan,~\IEEEmembership{Fellow,~IEEE}
\IEEEcompsocitemizethanks{
\IEEEcompsocthanksitem Zonghui~Guo, Yingjie~Liu, and Haiyong~Zheng are with the Faculty of Information Science and Engineering, Ocean University of China, Qingdao 266404, China. E-mails: guozonghui@ouc.edu.cn, yingjieliu@stu.ouc.edu.cn, zhenghaiyong@ouc.edu.cn.

\IEEEcompsocthanksitem Jie~Zhang and Shiguang~Shan are with the State Key Laboratory of AI Safety, Beijing 100086, China, the Institute of Computing Technology, Chinese Academy of Sciences, Beijing 100190, China, and the University of Chinese Academy of Sciences, Beijing 100049, China. E-mails: \{zhangjie, sgshan\}@ict.ac.cn.

\IEEEcompsocthanksitem Haiyong~Zheng and Shiguang~Shan are the corresponding authors.
}

}

\markboth{Journal of \LaTeX\ Class Files,~Vol.~14, No.~8, August~2015}%
{Shell \MakeLowercase{\textit{et al.}}: Bare Demo of IEEEtran.cls for Computer Society Journals}

\IEEEtitleabstractindextext{%
\begin{abstract}
Face Forgery Detection (FFD), or Deepfake detection, aims to determine whether a digital face is real or fake. Due to different face synthesis algorithms with diverse forgery patterns, FFD models often overfit specific patterns in training datasets, resulting in poor generalization to other unseen forgeries. Existing FFD methods primarily leverage pre-trained backbones with general image representation capabilities and fine-tune them to identify facial forgery cues. However, these backbones lack domain-specific facial knowledge and insufficiently capture complex facial features, thus hindering effective implicit forgery cue identification and limiting generalization. Therefore, it is essential to revisit FFD workflow across the \textit{pre-training} and \textit{fine-tuning} stages, achieving an elaborate integration from facial representation to forgery detection to improve generalization. Specifically, we develop an FFD-specific pre-trained backbone with superior facial representation capabilities through self-supervised pre-training on real faces. We then propose a competitive fine-tuning framework that stimulates the backbone to identify implicit forgery cues through a competitive learning mechanism. Moreover, we devise a threshold optimization mechanism that utilizes prediction confidence to improve the inference reliability. Comprehensive experiments demonstrate that our method achieves excellent performance in FFD and extra face-related tasks, \ie, presentation attack detection. Code and models are available at \href{https://github.com/zhenglab/FFDBackbone}{https://github.com/zhenglab/FFDBackbone}.
\end{abstract}

}
\maketitle
\IEEEdisplaynontitleabstractindextext
\IEEEpeerreviewmaketitle

\IEEEraisesectionheading{\section{Introduction}\label{sec:introduction}}
\IEEEPARstart{H}{UMAN} faces, being unique and readily available biometric data, play a pivotal role in various research fields, including face recognition~\cite{wang2021deep}, generation~\cite{melnik2024face}, and analysis~\cite{rivera2012local}. Unfortunately, facial information leakage and theft have spawned a clandestine industry where criminals exploit forged faces to perpetrate illegal activities, posing an extreme threat to national security, social stability, and individual well-being. Meanwhile, diverse GAN-based and Diffusion Model-based methods in Artificial Intelligence Generated Content (AIGC) have drastically reduced the barriers to creating lifelike forged faces. The proliferation of these forged faces online has led to significant crises in the security and authenticity of information. Therefore, distinguishing digitally forged faces created by image/video synthesis algorithms, \ie, \textit{Face Forgery Detection} (FFD), also referred to as Deepfake detection, has become a critical issue in both research and industrial communities.

\begin{figure}[t]
\centering
\includegraphics[width=1.0\linewidth]{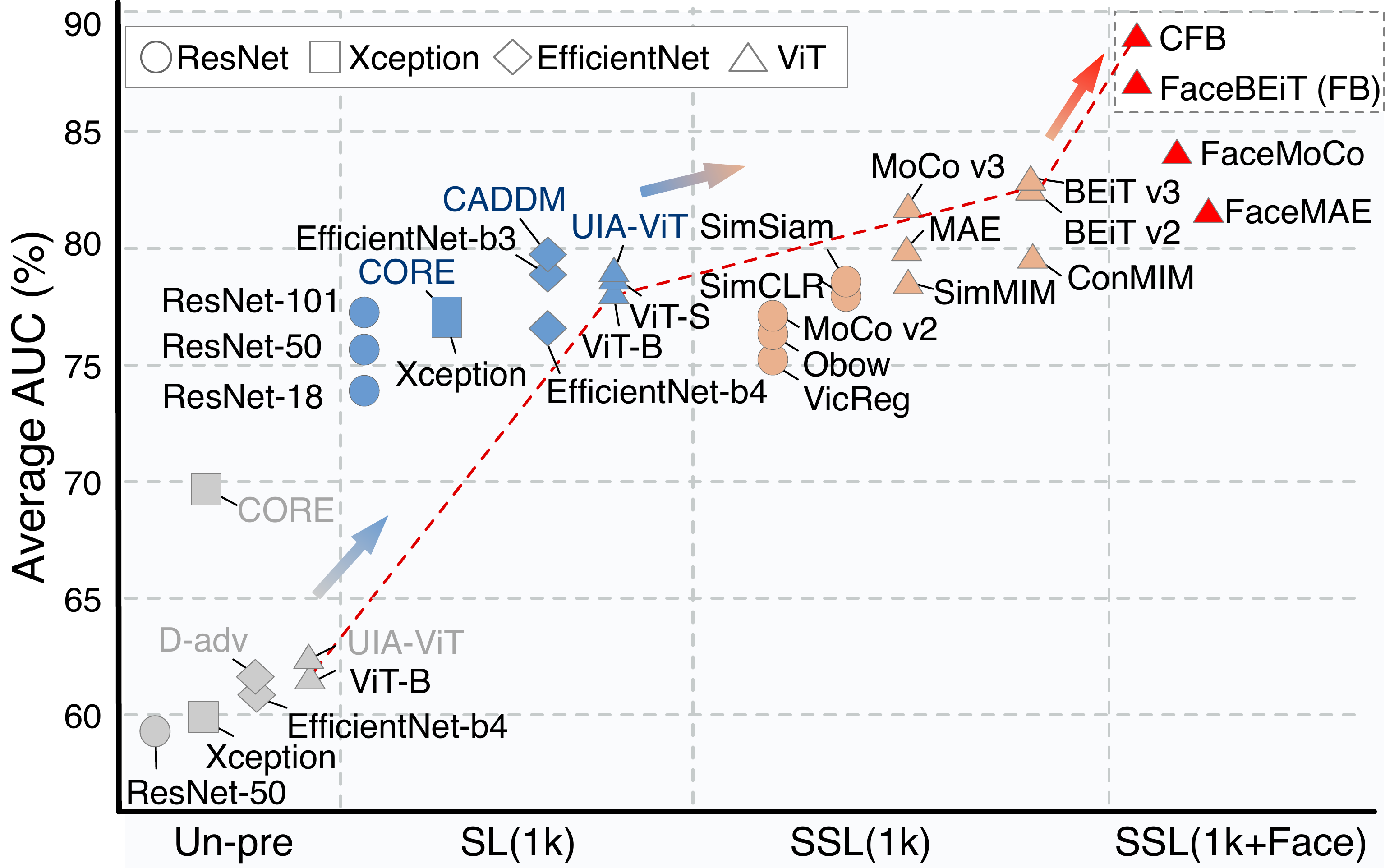}
\vspace{-15pt}
\caption{Quantitative comparison of FFD methods using various backbones without pre-training (Unpre-trained), as well as pre-training with supervised learning (SL) or self-supervised learning (SSL) on ImageNet-1k (1k) or further on real face (Face) datasets. The average AUC across three cross-datasets (Celeb-DF, DFDC, and FFIW) better reflects the generalization of FFD models.}
\vspace{-5pt}
\label{fig:compare_backbones}
\end{figure}

Unlike general image classification focusing on salient objects with specific structures and semantics, FFD requires identifying forgery cues that can appear irregularly anywhere on the face. As a result, many conventional CNN-based and Transformer-based architectures perform well on general tasks but often fail on FFD when trained end-to-end from scratch, as shown in the \textbf{Unpre-trained} results of Fig.~\ref{fig:compare_backbones}, backbones without pre-training typically perform poorly and may fail to converge, whereas those pre-trained with supervised learning on ImageNet-1k (\textbf{SL-1k}) show significant improvement, similar to the findings in DeepfakeBench~\cite{yan2023deepfakebench}. This marked difference is consistent across backbones including Xception~\cite{chollet2017xception}, EfficientNet~\cite{tan2019efficientnet}, and ViT~\cite{dosovitskiy2020vit}, substantiating why most existing FFD methods~\cite{ni2022core, zhuang2022uia,yan2023ucf, dong2023implicit, choi2024exploiting, yan2023transcending} employ pre-trained backbones as their foundational units. Meanwhile, \textbf{SL-1k} results also indicate that the performance of existing FFD methods such as CORE~\cite{ni2022core}, CADDM~\cite{dong2023implicit}, and UIA-ViT~\cite{zhuang2022uia} primarily depends on their pre-trained backbones, which provide slight improvements over Xception, EfficientNet-b3, and ViT-B, respectively.

Due to the diverse forgery patterns generated by different synthesis algorithms (\eg, entire face synthesis~\cite{Karras2021}, face swapping~\cite{liu2023fine}, face attribute editing~\cite{choi2020starganv2}, and face reenactment~\cite{prajwal2020lip}), FFD models often overfit to specific forgery patterns in training datasets, limiting their generalization to unseen data~\cite{shiohara2022detecting,yan2023ucf,yan2024transcending}. Consequently, recent FFD research primarily leverages pre-trained backbones and focuses on designing advanced fine-tuning techniques, including data augmentation to expand forgery diversity~\cite{shiohara2022detecting,li2020face,yan2023transcending}, as well as attention mechanisms and feature decoupling to assist the backbone in better capturing spatial forgery cues~\cite{zhao2021multi,fei2022learning,sun2022information,zhuang2022uia,yan2023ucf}. However, these backbones are mostly adopted from previous FFD work or selected based on strong performance in general tasks~\cite{shiohara2022detecting, zhuang2022uia, liu2024forgery}, without a comprehensive exploration of which pre-training configurations improve FFD's base performance. Although LipForensics~\cite{haliassos2021lips} and RealForensics~\cite{haliassos2022leveraging} utilize visual speech recognition and video-audio alignment as pre-training tasks, they depend on specific datasets with additional modality and are tailored to their particular approaches and applications.

Given the critical role of pre-trained backbones in FFD models, along with their extensive research in self-supervised learning (SSL) on general CV tasks~\cite{jing2020self,xie2022simmim,radford2018improving}, we argue that if SSL can be adapted to FFD and help to boost its performance. Thus, in this work, we systematically investigate pre-trained backbones with architectures such as ResNet and ViT via SSL on ImageNet-1k, including MoCo~\cite{he2020momentum}, SimSiam~\cite{chen2021exploring}, MAE~\cite{he2022masked}, and BEiT~\cite{peng2208beit}. As shown by the \textbf{SSL-1k} results in Fig.~\ref{fig:compare_backbones}, these backbones further improve FFD performance by benefiting from SSL's enhancement of image representation. Notably, MoCo v3~\cite{chen2021empirical} and BEiT v2~\cite{peng2208beit} boost ViT's image feature representational capacity and achieve substantial gains over ViT with \textbf{SL-1k}. Therefore, we dive deep to integrate SSL pre-training into FFD workflow and tailor them to tackle FFD challenges, yielding SSL-pretrained backbones for enhancing FFD models' sensitivity in identifying forgeries, as detailed in Sec.~\ref{sec:pre-training}.

Indeed, facial images differ significantly from generic natural images due to their unique characteristics, including distinct shapes, keypoint distributions, and locally consistent displacement fields~\cite{he2020deformable}. As synthesis techniques advance and fake faces become increasingly realistic, uncovering implicit forgery cues amid complex facial features becomes more challenging. However, existing FFD models primarily utilize backbones pre-trained on ImageNet~\cite{yan2023deepfakebench,yan2023transcending} or other natural image datasets such as LAION-2B~\cite{laion5b} and WIT~\cite{radford2021learning}, which lack facial domain knowledge and may insufficiently represent the complex characteristics of facial images. It is worth noting that facial features encompass comprehensive information beyond semantic attributes such as identity or gender, as overfitting to these attributes can hinder generalization~\cite{yan2023ucf,dong2023implicit}. Therefore, we further introduce real facial images to feed the SSL-pretrained backbones with face-specific knowledge, which can contribute to substantial improvements in FFD performance. As shown in the \textbf{SSL-1k\&Face} results in Fig.~\ref{fig:compare_backbones}, the backbones further pre-trained as FaceMAE, FaceMoCo, and FaceBEiT (\textbf{FB}) achieve varying degrees of improvement, with \textbf{FB} exhibiting particularly significant gains. Refer to Sec.~\ref{sec:pre-training} for more analysis.

In the above empirical studies of backbones with different pre-training configurations, we employ a basic end-to-end fine-tuning structure, \ie, \textit{Backbone and FC with Cross-Entropy loss}, which is commonly used and directly reveals the backbone’s ability to support downstream task performance~\cite{radford2018improving,chen2021empirical,he2022masked}. Moreover, in response to the subtle nature of forgery cues, we propose a competitive backbone fine-tuning framework that leverages dual backbones with an uncertainty-based fusion module to competitively identify implicit forgery cues, thereby improving the generalization performance of the pre-trained backbone. As shown by the \textbf{SSL-1k\&Face} results in Fig.~\ref{fig:compare_backbones}, our competitive fine-tuning method (\textbf{CFB}) achieves a $2.22\%$ significant performance improvement over its pre-trained backbone (\textbf{FB}), with additional experiments on other backbones presented in Table~\ref{tab:ablation2}.

In practice, existing FFD methods typically adopt an empirical threshold of $0.5$ to convert continuous prediction probabilities ($p \in [0, 1]$) into discrete labels (real or fake)~\cite{kendall2017uncertainties}. However, this classification threshold depends on the probability distribution predicted by the FFD model on the test sets, and $0.5$ is not necessarily suitable. To address this, we propose a threshold optimization mechanism that integrates prediction probability and its confidence to compute a relatively optimal threshold, thereby enhancing the practicality of FFD models during \textit{inference}.

Additionally, to explore the potential and practicality of our FFD method, we validate it from digital face to physical forgery face detection, which is known as presentation attack detection (PAD)\cite{george2019biometric}, face anti-spoofing\cite{jia2020single,wang2022domain,wang2022patchnet,long2023dual,liu2023towards,le2024grad}, and face liveness detection~\cite{chan2017face}. The PAD task focuses on identifying fake faces made from physical materials, such as printed photos, replayed images/videos, and 3D masks, all of which pose a significant threat in physical environments, particularly to face recognition systems.

In summary, the main contributions of this work are as follows: (1) we systematically analyze the crucial role of pre-trained backbones in FFD task and develop an elaborate backbone that incorporates self-supervised learning and real faces to enhance facial feature representation for better forgery cue identification; (2) we build a competitive backbone fine-tuning framework with a decorrelation constraint and uncertainty-based fusion module to stimulate the backbone’s identification of generalization-benefiting implicit forgery cues; (3) we devise a threshold optimization mechanism using prediction probabilities and their uncertainty to calculate the optimal threshold from the seen test set, which improves discrimination accuracy on unseen test sets; (4) we conduct extensive experiments to demonstrate the superiority and potential of our FFD method, achieving significantly better generalization than previous methods in FFD and extra face-related PAD tasks.

\section{Related Work}
We start by summarizing existing FFD methods, then provide an overview of supervised and self-supervised learning for backbone pre-training.

\subsection{Face Forgery Detection}

Thanks to the creation of numerous face forgery datasets such as FaceForensic++~\cite{rossler2019faceforensics++}, Celeb-DF~\cite{li2020celeb}, and DFDC~\cite{dolhansky2020deepfake}, deep learning-based FFD methods have made considerable advancements and can achieve outstanding accuracy within training datasets (exceeding $95$\%). However, their performance drops sharply when encountering unseen forgery patterns across different datasets. Recent FFD work aims to improve model generalization by implementing specific fine-tuning methods such as data augmentation, attention mechanisms, forgery feature decoupling, and temporal inconsistency, as well as spatial and frequency fusion.

Data augmentation enlarges the diversity of training data via various image transformations, which becomes a common setting in the training stage, especially for domain generalization tasks. Face X-ray~\cite{li2020face} proposed an augmentation method for generating forgery faces and boundaries by fusing source face images with similar faces obtained through nearest neighbor search. Chen~\etal~\cite{chen2022self} proposed a synthesized forgery face method from two real face images by adjusting forgery configurations, such as forgery region, blending type, and blending ratio. SBI~\cite{shiohara2022detecting} built a source-target generator to generate pseudo-source and target images through different image transformations, combined with blending masks to produce forged faces. LSDA~\cite{yan2023transcending} enriched the diversity of forged samples by fusing and transforming features of different forgery types in the latent space to learn more generalized decision boundaries.

Attention mechanisms, known for their ability to focus on task-relevant critical features, have been widely applied in most computer vision tasks. Many FFD methods also incorporate diverse attention networks into the middle or later layers of the backbone to assist in identifying forgery clues or localizing forged regions within faces~\cite{dang2020detection,miao2021learning,wang2021representative,zhao2021multi,fei2022learning,sun2022information,zhuang2022uia}. Some methods utilized convolutional or linear transformations for adaptive learning and estimating attention maps of forged regions~\cite{dang2020detection,zhao2021multi}, while others employed attention maps to erase conspicuous and specific forgery traces within the training dataset, compelling the detection model further to explore generic forgery features~\cite{miao2021learning,wang2021representative}. Subsequently, attention mechanisms have become increasingly sophisticated. Fei~\etal~\cite{fei2022learning} proposed a second-order local anomaly learning module to mine local region anomalies by calculating fine-grained local anomalies of the first and second order. Sun~\etal~\cite{sun2022information} found that face forgery is closely associated with high-information content and designed the Self-Information Attention to utilize self-information as a theoretical guide for capturing forgery cues while highlighting more informative regions. Zhuang~\etal~\cite{zhuang2022uia} proposed an Unsupervised Inconsistency-Aware method based on Vision Transformer, \ie, UIA-ViT, to extract inconsistency cues without pixel-level forged location annotations.

Due to the presence of redundant forgery-independent features in forged faces (\eg, identity, natural artifacts), the distinction of forgery-dependent features is disrupted and hindered. Researchers leveraged disentangled representation learning~\cite{wang2022disentangled} or game-theoretical approaches to enhance the backbone's ability for extracting forgery clues. Specifically, disentangled representation learning-based FFD methods primarily adopted a training process that involved exchanging decoupled forgery-independent and forgery-dependent features to reconstruct pairs between real and forged faces, enhancing the discriminative performance of the classification branch with the backbone encoder~\cite{cozzolino2018forensictransfer,hu2021improving,yan2022deepfake,cao2022end,yan2023ucf}. Game-theoretical-based FFD methods~\cite{dong2022explaining,yao2023towards} utilized the knowledge of game theory to decouple unrelated regions from source and target images, promoting FFD performance with the support of Shapley value~\cite{shapley1953value}.

FTCN~\cite{zheng2021exploring} and AltFreezing~\cite{wang2023altfreezing} explored temporal inconsistencies in forged videos. FTCN reduced the spatial convolution kernel, while AltFreezing alternated between spatial and temporal freezing, both aiming to encourage ResNet3D to explore generic features of forged face videos.   Choi~\etal~\cite{choi2024exploiting} employed StyleGRU and a style attention module to exploit the temporal low-variance properties  of style latent vectors in each frame for detecting forged videos. TFCU~\cite{guo2025face} identifies spatially discriminative features in fine detail by unraveling temporal forgery cues. Additionally, Wang~\etal~\cite{wang2022m2tr} and Miao~\etal~\cite{miao2023f} leveraged frequency domain information to assist the backbone in capturing more fine-grained artifact features.

Benefiting from the strong transfer capabilities of large-scale pre-trained models like CLIP in downstream tasks, recent studies have explored designing adapters tailored for FFD task. ForensicsAdapter~\cite{cui2025forensics} designs a CLIP adapter to transfer knowledge and enhance CLIP visual tokens for FFD generalization. Yan~\etal~\cite{yan2025generalizing} proposed a lightweight spatio-temporal adapter for CLIP, effectively enabling spatio-temporal joint learning. M2F2-Det~\cite{M2F2_Det_xiao} employs face-forgery prompt learning with CLIP to enhance generalization and integrates an LLM to generate detailed textual explanations of its detection decisions. DFD-FCG~\cite{han2025towards} introduces a facial component guidance mechanism to improve the generalization of spatial learning by prompting the model to focus on key facial regions. RepDFD~\cite{lin2025standing} reprograms CLIP model for FFD task by operating solely on its image and text inputs without tuning internal parameters. Additionally, UNITE~\cite{kundu2025towards} introduces an attention-diversity loss and employs SigLIP model for fully synthetic video detection. Zhang~\etal~\cite{zhang2024common} introduce a text and image-aware feature alignment formulation to enhance multi-modal representation learning for Deepfake Detection VQA task.

After reviewing previous FFD methods and their implementations, we find that these methods consistently equip various fine-tuning techniques to backbones, such as ResNet~\cite{he2016deep}, Xception~\cite{chollet2017xception}, EfficientNet~\cite{tan2019efficientnet}, and ViT~\cite{dosovitskiy2020vit}, which are pre-trained on natural image datasets. These backbones are often used directly in the \textit{fine-tuning} while overlooking the crucial roles of both the \textit{pre-training} and \textit{inference}, as shown in Fig.~\ref{fig:framework}(top). In contrast, we endeavor to offer comprehensive research insights and guidance for FFD generalization by revisiting the complete FFD workflow.

\subsection{Supervised and Self-Supervised Learning}

Early backbone research utilized supervised learning (SL) with an image classification task as the pre-training approach, focusing on designing more sophisticated network architectures to improve backbone capacity and capability. During this period, numerous well-designed networks developed for the ImageNet Large-Scale Visual Recognition Challenge demonstrated outstanding performance, such as VGG, ResNet, and SENet. Benefiting from Transformers' global contextual attention design, backbone network architectures have gradually transitioned from CNNs to Transformers. In particular, Vision Transformers (ViTs)~\cite{dosovitskiy2020vit} have come to dominate the field of computer vision (CV), achieving superior performance compared to ResNet, Xception, and EfficientNet.

Since SL with the classification pretext task relies on annotated datasets, it somewhat restricts the capability of pre-trained backbones. Recent research on backbones has gradually shifted from SL to self-supervised learning (SSL)~\cite{ozbulak2023know, jing2020self}, which can flexibly employ different pretext tasks and larger-scale training datasets without annotations. SSL has significantly improved backbones' feature representation capabilities, surpassing SL methods~\cite{he2016deep,dosovitskiy2020vit} in most CV tasks. SSL is usually divided into discriminative and generative approaches based on different technical implementations and pretext tasks. With technological advancements, contrastive learning (CL) and masked image modeling (MIM) have gradually become the mainstream discriminative and generative SSL paradigms, respectively.

Discriminative SSL typically ensures representation consistency across different augmented views of the same image using Siamese architectures, which can be divided into clustering, CL, distillation, and information maximization. Clustering SSL methods, such as Deep Cluster~\cite{caron2018deep}, SCAN~\cite{van2020scan}, and Swav~\cite{caron2020unsupervised}, are among the early SSL methods and used techniques like K-Means, advanced neighbor search, and contrastive elements. CL methods, such as MoCo~\cite{chen2003improved, chen2021empirical, he2020momentum} and SimCLR~\cite{chen2020simple, chen2020big}, achieved instance discrimination by pulling positive samples closer in the latent space while pushing negative samples further apart. Distillation SSL methods, such as BYOL~\cite{grill2020bootstrap}, SimSiam~\cite{chen2021exploring}, and DINO~\cite{caron2021emerging}, trained a student network to predict the outputs of a teacher network, preventing collapse into identical representations and eliminating the need for negative samples. VICReg~\cite{bardes2021vicreg} is an information maximization SSL method that eliminates correlations between different samples in the feature space to learn independent features.

The success of autoregressive modeling in GPT~\cite{radford2018improving} and masked autoencoding in BERT~\cite{devlin2018bert} for NLP has driven the development of generative SSL with MIM in computer vision. The MIM approach involves masking part of the input image and training the backbone to predict the missing content. Based on different pretext tasks, MIM can be divided into predicting pixel values (\eg, MAE~\cite{he2022masked} and SimMIM~\cite{xie2022simmim}) and predicting discrete visual tokens (\eg, BEiT~\cite{bao2021beit, peng2208beit, wang2023image} and CAE~\cite{chen2024context}).

Given the significance of learning approaches with different pretext tasks for backbones in the \textit{pre-training} stage, we conduct a comprehensive empirical study to investigate their performance in FFD task and provide valuable guidelines for their selection and optimization.

\section{FFD Workflow and Method}

\subsection{Research Overview}

\begin{figure}[t]
\centering
\includegraphics[width=1.0\linewidth]{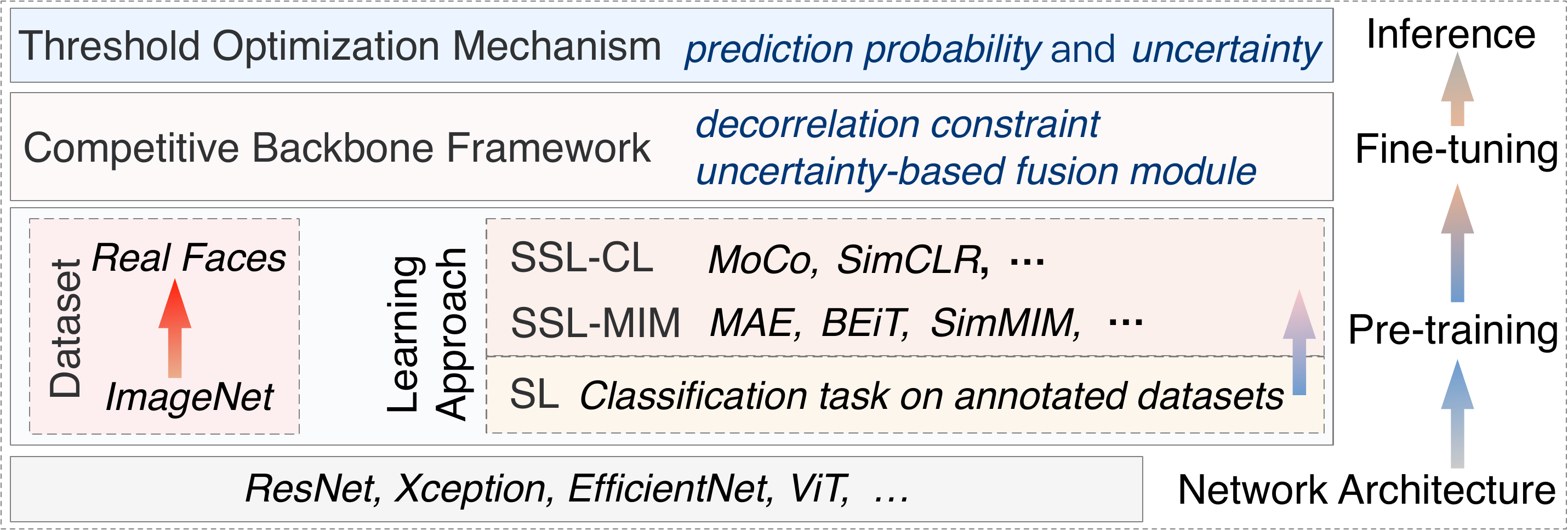}

\caption{Overview of our research to revisit the complete FFD workflow from \textit{pre-training} and \textit{fine-tuning} to \textit{inference} in discriminant results. This includes network architectures, datasets, learning approaches with supervised learning (SL) and self-supervised learning (SSL), which encompass contrastive learning (CL) and masked image modeling (MIM) pretext tasks, as well as our competitive backbone fine-tuning framework and threshold optimization mechanism.}
\label{fig:overview}
\end{figure}

Our work leverages backbone pre-training as the starting point, and following our investigation of different pre-training configurations, we construct a comprehensive research overview, as shown in Fig.~\ref{fig:overview}. This overview covers the complete FFD workflow: (1) the network architectures, (2) the learning approaches with SL and SSL, as well as the datasets such as ImageNet and real faces during \textit{pre-training}, (3) the adaptation methods to downstream tasks during \textit{fine-tuning}, and (4) the application mechanisms to maximize performance during \textit{inference}.

Specifically, we explore the critical role of backbones with different configurations in FFD task and seek insightful solutions that align with FFD challenges during the \textit{pre-training} stage (Section~\ref{sec:pre-training}). We then introduce our backbone fine-tuning framework with a decorrelation constraint and an uncertainty-based fusion module, which improves the backbone's ability to extract forgery cues during the \textit{fine-tuning} stage (Section~\ref{sec:fine-tuning}). Moreover, we analyze the limitations of FFD methods in discriminant results and propose a threshold optimization mechanism by integrating prediction probability and confidence to enhance the practicality of FFD models during the \textit{inference} stage (Section~\ref{sec:inference}).

\begin{figure*}[t]
\centering
\includegraphics[width=1.0\linewidth]{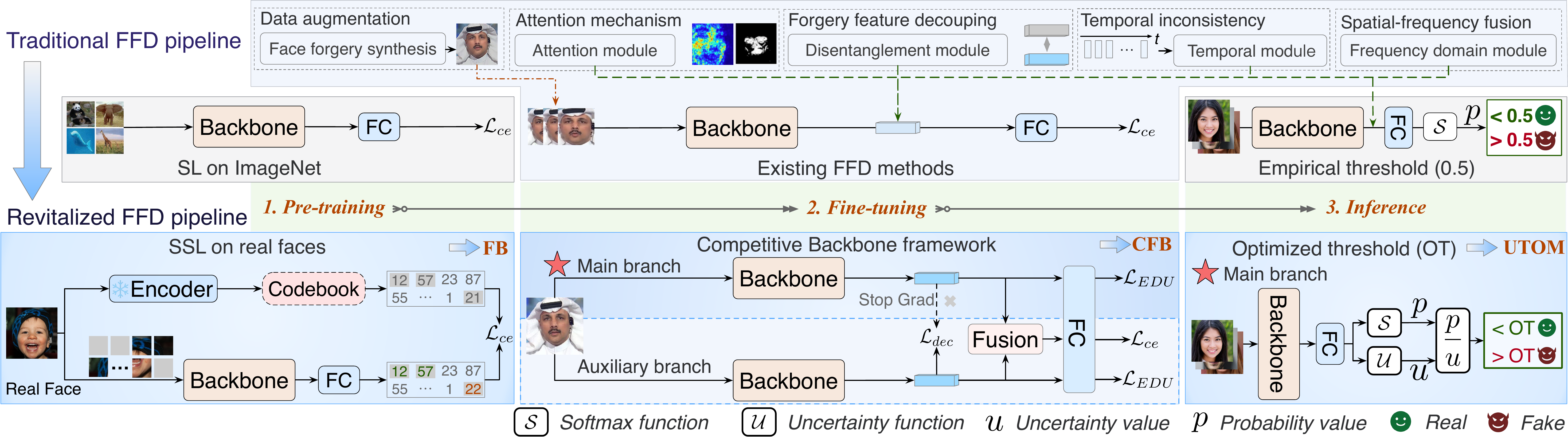}
\caption{FFD workflow evolves from the traditional pipeline to our revitalized FFD pipeline. Existing FFD methods primarily rely on backbones pre-trained with supervised learning on ImageNet, applying various techniques during \textit{fine-tuning}, and using empirical classification thresholds during \textit{inference}. In contrast, our FFD pipeline offers a more proficient, promising, and reliable solution by incorporating self-supervised learning on real faces, a competitive backbone framework, and an uncertainty-based threshold optimization mechanism across the three stages.} 
\label{fig:framework}
\end{figure*}

\subsection{FFD with Pre-trained Backbones}
\label{sec:pre-training}

We abstract a traditional pipeline of existing FFD methods to highlight the vital role of backbones, as shown in Fig.~\ref{fig:framework}(top). These methods directly use backbones pre-trained with SL on ImageNet, implement fine-tuning networks or strategies to equip the backbone for better generalization, and apply an empirical classification threshold ($0.5$) to output the final discriminant result (real or fake)~\cite{shiohara2022detecting,miao2021learning,wang2021representative}.

Relying on our research overview, we first explore the impact of pre-training backbones with different network architectures and learning approaches in FFD task. Technically, we utilize a basic fine-tuning structure (\textit{Backbone and FC with Cross-entropy loss}) and prioritizing various mainstream backbones pre-trained on ImageNet to evaluate FFD performance. Meanwhile, we select CNN-based architectures such as ResNet, Xception, and EfficientNet, which are common in previous FFD methods, as well as Transformer-based architectures like ViT, which are superior in most other CV tasks. For learning approaches, we employ SL and SSL with different pretext tasks, including contrastive learning (CL) tasks such as SimSiam~\cite{chen2021exploring} and MoCo~\cite{chen2003improved,chen2021empirical}, as well as masked image modeling (MIM) tasks such as MAE~\cite{he2022masked}, BEiT~\cite{peng2208beit,wang2023image}, and ConMIM~\cite{yi2022masked}.

In our comprehensive comparison (refer to Section~\ref{chap:compare backbone}), we statistically find that backbones pre-trained with SSL outperform those with SL under the same network architecture, and Transformer-based backbones outperform CNN-based ones with the same learning approach. These results highlight the crucial role of SSL in pre-training ViT-based backbones to enhance FFD model generalization, with the representation capabilities of SSL approaches directly influencing FFD performance. Notably, BEiT~\cite{peng2208beit,wang2023image} achieves superior performance, benefiting from its ability to independently represent visual tokens of image patches, thus better extracting differential cues between facial local components.

Since most mainstream backbones are pre-trained on ImageNet or other natural image datasets such as LAION-2B~\cite{laion5b} and WIT~\cite{radford2021learning}, which contains content significantly different from facial images, this training data gap will also restrict the performance of general-purpose backbones in facial-related tasks, particularly FFD, which requires identifying subtle forgery cues within complex facial features. Furthermore, compared to forgery faces, we can easily collect large-scale and diverse real-face datasets~\cite{liu2015faceattributes,karras2019style}. Although these datasets may lack labels, we can still leverage SSL with appropriate pretext tasks to pre-train a robust face representation backbone with face-specific knowledge tailored to facial tasks. In the technical implementation of the backbone, we adopt ViT as the network architecture and employ SSL methods such as MoCo v3~\cite{chen2021empirical}, MAE~\cite{he2022masked}, and BEiT v2~\cite{peng2208beit}, which are concise and classical in CL and MIM.

As evident in \textit{SSL-Face} and \textit{SSL-1k}\&\textit{Face} of Table~\ref{tab:backbones}, the backbone pre-trained on real-face datasets yields superior FFD generalization compared to one on ImageNet. Notably, MoCo v3 and BEiT v2 are superior to MAE, whose performance is relatively weaker. These results stem from the close relationship between FFD core challenges and the backbone's pretext tasks. On the one hand, forgery cues typically manifest in the local components of fake faces (\eg, eyebrows, eyes, lips) and the contrast between facial components, involving both low-level features and semantic errors, as depicted in Fig.~\ref{fig:local}. On the other hand, MoCo v3, with its CL pretext task that compares the similarity of high-level features from differently augmented images, can better focus on forgery cues in local regions between these images. Meanwhile, with a classification pretext task based on encoding image patches, BEiT v2 excels at discerning the authenticity of local facial components. The excellent performance of MoCo v3 and BEiT v2 further demonstrates the indispensability of equipping backbones with both general and specific knowledge through appropriate SSL approaches to enhance FFD generalization.

Based on our comprehensive study of backbones in the \textit{pre-training} stage, we revitalize the traditional FFD pipeline by integrating SSL on real-face datasets, a competitive backbone framework, and an uncertainty-based threshold optimization mechanism in the \textit{pre-training}, \textit{fine-tuning}, and \textit{inference} stages. As shown in Fig.~\ref{fig:framework} (bottom), our revitalized FFD pipeline enables a more proficient and promising step-by-step enhancement of the backbone's baseline and task-specific capabilities, as well as greater reliability in FFD discriminant results. Additionally, given the superior performance of BEiT v2, we use its SSL method and ViT architecture as the representative pre-training configuration in our pipeline, as shown in Fig.~\ref{fig:framework}(left-bottom).

\subsection{Competitive Backbone Fine-tuning Framework}
\label{sec:fine-tuning}

\begin{figure}[t]
  \centering
  \includegraphics[width=1.0\linewidth]{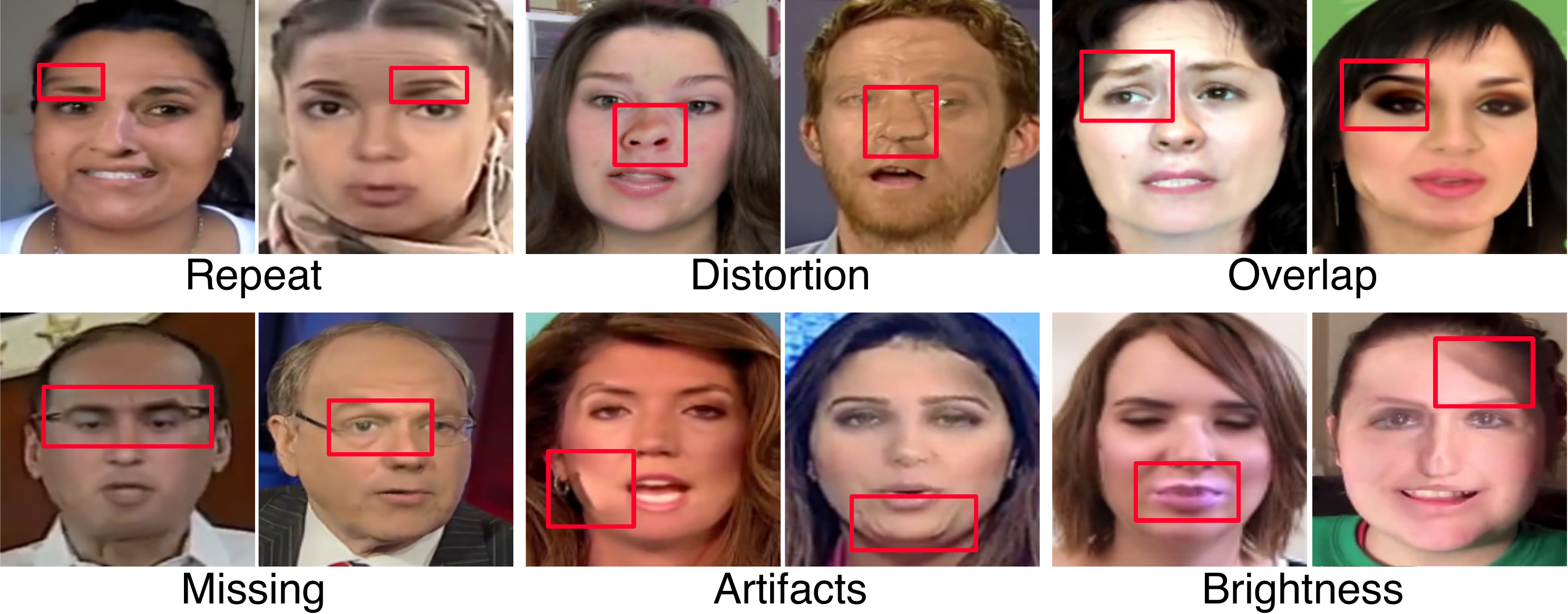}
\vspace{-15pt}
  \caption{Representative forged face examples with forgery cues primarily in facial components such as eyebrows, nose, eyes, and lips.}
  \label{fig:local}
\vspace{-10pt}
\end{figure}

Building on the pre-trained backbone with enhanced facial representations, we further explore effective fine-tuning methods with improved generalization to identify implicit forgery cues in the represented facial features. Nevertheless, apparent and implicit forgery cues in the latent feature space of training sets are even more diverse and entangled, making it nearly impossible to explicitly disentangle these two types of cues due to the lack of available definitive constraints. In practice, humans tend to detect forgeries by noticing semantic repetition or inconsistencies in facial components and contrast in appearance between them, as illustrated in Fig.~\ref{fig:local}. Motivated by this, we consider leveraging two unshared backbones to competitively mine different forgery cues in the latent space to enhance generalization.

Referring to the commonly used dual-branch Siamese architecture in contrastive learning (\eg, MoCo), we propose building a dual-branch framework that cultivates two backbones to extract different forgery cues within a competitive adversarial learning mechanism. However, contrastive learning approaches primarily constrains the backbones to represent the semantic similarity of two differently augmented images, which involve changes in both high-level (\eg, crop, resize) and low-level (\eg, color distortion, Gaussian blur) features. These different augmentations of the same image may potentially lead to the loss or introduction of subtle forgery traces related to low-level features, such as artifacts and appearance. Thus, we remove the image augmentations and contrastive loss while retaining only the dual-branch architecture and incorporating cross-entropy loss $\mathcal{L}_{ce}$ for FFD. Furthermore, we fuse the output features of both backbones to enable their interaction, thereby forming our basic dual-branch backbone fine-tuning architecture.

Since the dual branches receive the same inputs (without distinct augmentations) and are optimized with the same loss ($\mathcal{L}_{ce}$), the two unshared backbones tend to learn similar features despite independent weight updates. To address this, we design a decorrelation constraint and an uncertainty-based fusion module to enforce dissimilarity between features extracted by the two unshared backbones, thereby enabling competitive learning within the dual-branch architecture, distinct from the joint or contrastive paradigms used in prior work~\cite{chen2021exploring,he2020momentum,radford2021learning}.




\textbf{Decorrelation Constraint.}
We employ the Pearson correlation coefficient as the decorrelation constraint $\mathcal{L}_{dec}$ to measure and minimize the linear correlation between class tokens of the dual-branch backbones. Under the combined constraints of $\mathcal{L}_{dec}$ and $\mathcal{L}_{ce}$ on the dual branches, both backbones achieve competitive learning of different forgery cues. Moreover, we utilize a gradient-stopping backpropagation strategy on one branch under the constraint of $\mathcal{L}_{dec}$ to particularly foster the forgery cue extraction capability of that branch's backbone, while enabling both backbones to compete in the fusion module for their contributions to this task. Additionally, we refer to the fostered branch as the \textbf{main} branch and the other as the \textbf{auxiliary} branch.

At this point, the dual-branch architecture with decorrelation constraint is preliminarily established, as shown in Fig.~\ref{fig:framework}(middle-bottom), namely \textit{competitive backbone fine-tuning framework}. It includes two unshared backbones processing the same face image, feature fusion module, and three $\mathcal{L}_{ce}$, which constrain both backbones to competitively mine different forgery cues to enhance the main branch's generalization. $\mathcal{L}_{dec}$ and $\mathcal{L}_{ce}$ are defined as follows:
\begin{align}
\vspace{-10pt}
\label{eq:method_loss}
    \mathcal{L}_{dec} &= \frac{\sum_{i=1}^C (f_M^i - \overline{f}_M)(f_A^i - \overline{f}_A)}{\sqrt{\sum_{i=1}^C (f_M^i - \overline{f}_M)^2 \sum_{i=1}^C (f_A^i - \overline{f}_A)^2}}, \\
    \mathcal{L}_{ce} &= -\left(y \log(p) + (1 - y) \log(1 - p)\right),
    \vspace{-10pt}
\end{align}
where $f$ represents the feature vectors from both main ($M$) and auxiliary ($A$) backbones with dimensions $C$, while $y$ and $p$ are the label and predicted probability, respectively. Notably, we adopt $[\texttt{CLS}]$ token as $f$, and utilizing all tokens from final Transformer layer yields similar performance, provided consistency with classification head input is maintained. Further analysis is available in \textit{Supp. File}.

\textbf{Uncertainty-based Fusion Module.}\label{sec:uncertainty} Indeed, directly summing the feature vectors of both backbones implies equal contribution, which will limit their competition in a naive fusion operation. To stimulate the competitive detection of forgery cues by both backbones, we further explore assigning different fusion weights to each branch based on the confidence of their discrimination results, dynamically giving greater weight to the more confident branch.

\begin{figure}[t]
\centering
\includegraphics[width=\linewidth]{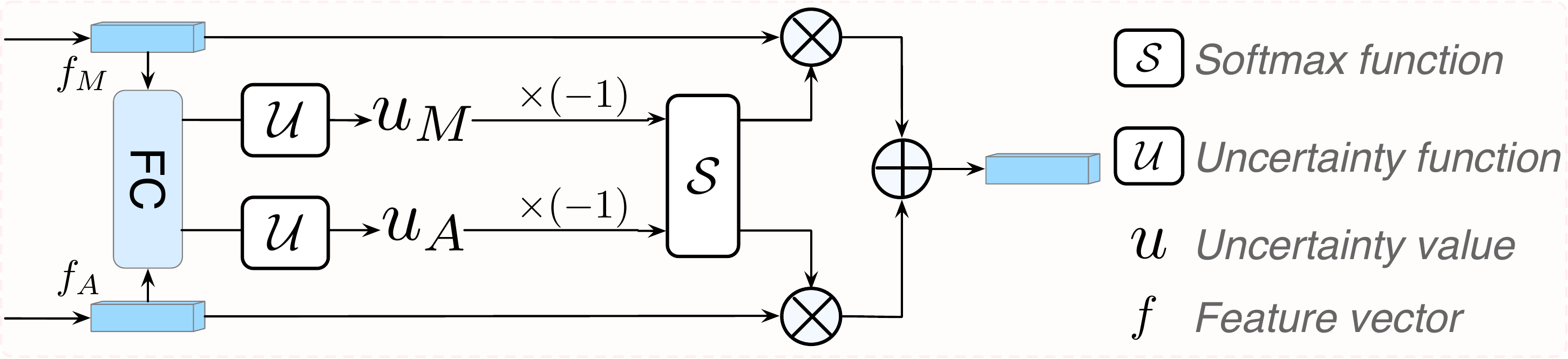}
\vspace{-15pt}
\caption{Our uncertainty-based fusion module, calculates uncertainty using features from the main (M) and auxiliary (A) branches, then inputs these into \textit{Softmax} to obtain weights for fusing the features as output.}
\label{fig:fusionmodule}
\end{figure}

FFD models typically use \textit{Softmax} in their final layer to convert the continuous output values into class probabilities. However, \textit{Softmax} tends to exaggerate the probabilities due to its exponential function, which does not accurately reflect the model's prediction confidence. Despite minimizing the $\mathcal{L}_{ce}$ of predicted classes during training, the classification model lacks awareness of its prediction confidence. To overcome this issue, uncertainty estimation methods have been proposed to assess the model's confidence in its predictions~\cite{kendall2017uncertainties}. Particularly, Evidence Deep Learning (EDL) simultaneously predicts a model's class probability and its uncertainty by assuming that the predicted class probabilities follow a prior Dirichlet distribution~\cite{sensoy2018evidential,bao2021evidential}. Therefore, we utilize the uncertainties ($u$) from both branches as weights ($w$) to fuse their feature vectors, as shown in Fig.~\ref{fig:fusionmodule}, enabling flexible and dynamic competition between both backbones. The fusion process is as follows:
\begin{align}
\label{eq:uncertainty_formula}
&S=\sum\nolimits_{k=1}^K(e_k+1), ~\text{where}~~e = \sigma\Big(\Phi\big(f\big)\Big),\\
\label{eq:uncertainty_fusion}
&w_M, w_A = softmax(-u_M, -u_A),~\text{where}~~u= K/S,\\
&F = w_M f_M + w_A f_A,
\end{align}
where $\Phi$ denotes the classification header (\eg, FC), $\sigma$ is a non-negative evidence function ensuring that evidence $e\geq0$ (\eg, exp, ReLU), and $K$ denotes the total number of classes, \ie, $K=2$ in FFD task.

Furthermore, we employ EDL and Evidential Uncertainty Calibration (EUC) losses~\cite{bao2021evidential} to improve the accuracy of predicted probability and uncertainty, as follows:
\begin{align}
\label{eq:loss_edl}
\mathcal{L}_{EDL} =& \sum\nolimits^K_{k=1}y_k\big(\log S-\log (e_k+1)\big),\\
\mathcal{L}_{EUC} =&-\lambda_{t} \sum_{i \in\left\{\hat{y}=y\right\}} p \log \left(1-u\right) \\
&-\left(1-\lambda_{t}\right) \sum_{i \in\left\{\hat{y} \neq y\right\}}\left(1-p\right) \log \left(u\right),
\end{align}
where $y$ denotes a one-hot $K$-dimensional label, $\hat{y}$ represents the class with the highest probability $p$ for the input image, and $p$ is calculated as $p=\max((e_{k}+1)/S)$. Additionally, the annealing factor $\lambda_t$ is defined as $\lambda_{t}=\lambda_{0} \exp \left\{-\left(\ln \lambda_{0} / T\right) t\right\}$, where $t$ and $T$ denote the current training epoch and the total number of training epochs, respectively. For further details, please refer to Bao~\etal\cite{bao2021evidential}.

As shown in Fig.~\ref{fig:framework}(middle-bottom), we replace the $\mathcal{L}_{ce}$ with $\mathcal{L}_{EDL}$ and $\mathcal{L}_{EUC}$ in both the main and auxiliary branches. Thus, the loss function $\mathcal{L}$ of our competitive backbone fine-tuning framework for FFD task is as follows:
\begin{align}
\label{eq:all_loss}
\mathcal{L} =& \mathcal{L}_{EDU}^M+\mathcal{L}_{EDU}^A+\mathcal{L}_{ce}^F+\mathcal{L}_{dec},
\end{align}
where $\mathcal{L}_{EDU}=\mathcal{L}_{EDL}+\mathcal{L}_{EUC}$.

Thanks to the dedicated design of our decorrelation constraint and uncertainty-based fusion module in our competitive backbone fine-tuning framework, the main backbone is better fostered and significantly enhances its generalization under the stimulation of the auxiliary backbone. Consequently, only the main branch is required during \textit{inference}, also saving half of the computational resources.

\subsection{FFD Inference with Optimized Threshold}
\label{sec:inference}

In FFD applications, a threshold $\tau$ is required as an anchor value to classify the FFD model's continuous probability outputs ($p \in [0,1]$) into discrete results, such as $p \geq \tau$ indicates fake and $p < \tau$ indicates real. Due to factors such as network architecture, training strategies, and dataset, prediction probability distributions for real and fake faces vary across FFD models. As a result, empirically setting $\tau=0.5$ is not suitable as a universal threshold for all FFD models. Particularly, FFD models exhibit limited generalization on unseen datasets, yielding predicted probabilities $p$ that are more evenly distributed between real and fake faces, so different choices of $\tau$ can greatly affect discriminative accuracy. In real-world scenarios, forged face images are produced by ever-emerging unseen manipulation algorithms, for which ground-truth labels are unknown to the detector. Therefore, estimating a relatively optimal threshold $\tau_{ot}$ to improve discriminative accuracy is both necessary and challenging.


Before estimating $\tau_{ot}$, as mentioned in the Uncertainty-based Fusion Module of Section~\ref{sec:uncertainty}, the prediction probability $p$ produced by the \textit{Softmax} layer in most FFD models can cause the model to be overly confident in its predictions, even when it is wrong. To mitigate this, we introduce uncertainty $u$ to adjust $p$ via division, yielding $p' = p/u$, where both $p$ and $u$ are predicted by our FFD model. Such adjustment increases $p'$ under high confidence (low uncertainty) and decreases $p'$ under low confidence (high uncertainty). Meanwhile, we apply $\tanh$ to map scores into $[0, 1)$ via $p' = \tanh( \lambda p / 2(u+\epsilon) )$, using $\epsilon = 10^{-7}$ and scaling factor $\lambda = 0.02$, which is further explained in \textit{Supp. File}.

%

Accordingly, we propose an Uncertainty-based Threshold Optimization Mechanism (UTOM) that computes the optimal threshold $\tau_{ot}^{UTOM}$ on one cross-dataset and applies it as the relatively optimal classification threshold for other unseen datasets, which also applies in real-world scenarios where ground-truth labels for face images are unavailable.
Specifically, we iterate through the adjusted probabilities $p'$ of all faces in one cross-dataset, treating each probability as a potential threshold and calculating the corresponding ACC for the dataset. The threshold with the highest ACC is determined as $\tau_{ot}^{UTOM}$, and its calculation is as follows:
\begin{align}
\label{eq:uncertainty}
\tau_{ot}^{UTOM} &= \underset{\tau}{\arg\max} \, \text{ACC}(\tau), \\
\label{eq:ACC}
\text{ACC}(\tau) &= \frac{1}{N} \sum_{i=1}^{N} \left[ \mathbb{I}(p'_i \geq \tau) y_i + \mathbb{I}(p'_i < \tau) (1 - y_i) \right],
\end{align}
where $N$ is the total number of face images, $y_i$ is the label for the $i$-th image (with $1$ for fake and $0$ for real), and $\mathbb{I}$ is the indicator function returning $1$ if true and $0$ otherwise. Moreover, comparison of $\tau=0.5$ and $\tau=\tau_{ot}^{UTOM}$ in Table~\ref{tab:inference} indicates that $\tau_{ot}^{UTOM}$ from one cross-dataset can significantly improve ACC of other unseen datasets, underscoring the critical role of optimized thresholds in FFD applications.




\section{Experiment}
\subsection{ Experimental Setup}
\subsubsection{Datasets}
We conduct experiments by \textit{pre-training} on real-face dataset, \textit{fine-tuning} on c23 of Faceforensics++~\cite{rossler2019faceforensics++} (FF++), and evaluating on cross-datasets with various forgery algorithms during \textit{inference}. Specifically, our real-face dataset contains $550k$ images from CelebA~\cite{liu2015faceattributes}, CelebV-Text~\cite{yu2022celebvtext}, and FFHQ~\cite{karras2019style}. FF++ contains $1k$ real videos and $4k$ manipulated videos obtained by four forging methods: DeepFakes~\cite{deepfakes2018}, FaceSwap~\cite{faceswap2020}, Face2Face~\cite{thies2016face2face}, and NeuralTexture~\cite{thies2019deferred}. Cross-datasets consist of Celeb-DF~\cite{li2020celeb} (CDF), DFDC~\cite{dolhansky2020deepfake}, FFIW~\cite{zhou2021face}, DFDCP~\cite{dolhansky1910deepfake}, and DFD~\cite{Deepfake21}. CDF includes real videos of $59$ celebrities from YouTube and fakes generated via an improved synthesis process~\cite{li2020celeb}. DFDC is a large-scale face-changing video dataset, created using various Deepfake and GAN-based methods, with a total of $5k$ test videos. FFIW is a challenging multi-face forgery dataset with $10k$ videos. DFDCP is a preview version of DFDC with $5k$ videos. DFD contains over $3k$ manipulated videos featuring $28$ actors across various scenes.

Furthermore, we evaluate the generalization ability of our method on DF40~\cite{yan2024df40}, a new comprehensive FFD benchmark. Additionally, we construct $9$ new cross-datasets generated by $6$ mainstream GAN-based methods and $3$ Diffusion Model-based methods, encompassing four types of forgeries: entire face synthesis, face attribute editing, face swapping, and face reenactment. Each cross-dataset contains $5k$ real faces and $5k$ forged faces.

\subsubsection{Implementation Details}
Following MoCo v3, MAE, and BEiT v2 settings, we pre-train ViT on our real-face dataset using $4$x A100 GPUs until full convergence. We use RetinaFace~\cite{deng2019retinaface} for face cropping, resize the cropped faces to $224$×$224$, and sample $8$ frames at equal intervals from each video. Fine-tuning is performed on $2$x GTX 3090 GPUs with batch size $64$, using the Lamb optimizer~\cite{you2019large} and a learning rate $5e-5$ with cosine decay for $60$ epochs. Notably, backbones pre-trained on real faces with more robust facial feature representation capabilities converge in just $10$ epochs.
In comparisons, we use $64$ frames per video for temporal-based FFD methods because they require more frames, and $8$ frames per video for spatial-based FFD methods, which are independent of video length.


\subsubsection{Evaluation Metrics and Protocols}

\textbf{Evaluation Metrics.} Similar to existing FFD and PAD works, we use the Area Under the Receiver Operating Characteristic Curve (AUC) and Half Total Error Rate (HTER) as evaluation metrics, respectively. Our work presents \textbf{video-level} AUC results, where frame prediction probabilities are averaged within each video to obtain its final probability, followed by AUC calculation over the test set.

\textbf{Evaluation Protocols.}
Existing FFD works typically evaluate the model's generalization performance on unseen cross-datasets, such as training on FF++ and evaluating on CDF, DFDC, and FFIW. In real-world scenarios, it is possible to obtain a small number of images from unseen datasets. Further fine-tuning the FFD model with these few images can better demonstrate its adaptability, as a highly adaptable model can effectively counter emerging forgery algorithms while providing significant real-world utility. Therefore, we evaluate the FFD model's practicality and robustness using two protocols: generalization and adaptation.

\subsection{Comparison of FFD with Pre-trained Backbones}
\label{chap:compare backbone}

We evaluate the importance of pre-trained backbones for FFD models by analyzing three core pre-training configurations: learning approaches, network architectures, and datasets. Table~\ref{tab:backbones} presents the quantitative results, with the unpre-trained backbone as the baseline (Unpre-trained).

\textbf{Unpre-trained Backbones.} We utilize unpre-trained backbones with different architectures, including ResNet, Xception, EfficientNet, and ViT, and fine-tune them on FF++ dataset. These models yield an average AUC of around $61$\%, indicating predictions close to random guessing.

\begin{table}[!htbp]

    \footnotesize
    \setlength{\tabcolsep}{0.001mm}
    \caption{Quantitative comparison of FFD with different backbones. The average AUC (Avg.) across three cross-datasets is provided for reference. ${\textit{\textbf{1K}}}$ and ${\textit{\textbf{Face}}}$ refer to ImageNet-1K and our real-face dataset, respectively. $\textbf{\textit{1K}\&{\textit{Face}}}$ denotes a backbone pre-trained on ${\textit{\textbf{1K}}}$ and then further pre-trained on ${\textit{\textbf{Face}}}$. \colorbox{gray!20}{\textbf{Bold}}, \textbf{bold}, and \textbf{\textit{italics}} indicate the best overall, second-best, and best in each section, respectively.}

	\centering
    \renewcommand{\arraystretch}{1.05}
	\midsepremove
    
		\begin{tabular}{l c!{\vrule height 1.05\ht\strutbox depth 3pt} c c c c}
        \hline
  
\multirow{2}{*}{Backbone} & \multirow{1}{*}{In-set} &\multicolumn{4}{c}{\multirow{1}{*}{Cross-set AUC$\uparrow$ (\%)}}  \\
  \cline{2-6}
\rule{0pt}{7.5pt} 
 &   FF++ & CDF & DFDC & FFIW & Avg.\\
		\midrule
\multicolumn{5}{l}{\multirow{1}{*}{\textit{\textbf{1. Unpre-trained}}}}\\
ResNet-18~\cite{he2016deep} & $72.87$ & $58.23$ & $61.53$ & $60.05$ & $59.94$\\
ResNet-50~\cite{he2016deep} & $75.74$ & $57.86$ & $61.72$ & $59.99$ & $59.86$\\
ResNet-101~\cite{he2016deep} & $76.54$ & $57.58$ & $63.74$ & $56.84$ & $59.39$\\
Xception~\cite{chollet2017xception} & $\textbf{\textit{80.18}}$ & $61.26$ & $63.29$ & $57.41$ & $60.65$\\
EfficientNet-b0~\cite{tan2019efficientnet} & $71.77$ & $54.55$ & $61.61$ & $62.89$ & $59.68$\\
EfficientNet-b3~\cite{tan2019efficientnet} & $73.57$ & $\textbf{\textit{64.14}}$ & $62.11$ & $\textbf{\textit{66.09}}$ & $\textbf{\textit{64.11}}$\\
EfficientNet-b4~\cite{tan2019efficientnet} & $73.07$ & $62.60$ & $57.38$ & $57.73$ & $59.24$\\
ViT-S~\cite{dosovitskiy2020vit} & $70.71$ & $58.41$ & $\textbf{\textit{64.80}}$ & $54.60$ & $59.27$\\
ViT-B~\cite{dosovitskiy2020vit} & $72.05$ & $61.85$ & $63.61$ & $57.91$ & $61.12$\\
\hline
\hline
\multicolumn{5}{l}{\multirow{1}{*}{\textit{\textbf{2. SL-1K}}}}\\
ResNet-18~\cite{he2016deep}  & $97.73$ & $79.09$ & $72.44$ & $71.23$ & $74.25$\\
ResNet-50~\cite{he2016deep}  & $97.95$ & $81.24$ & $72.23$ & $74.21$ & $75.89$\\
ResNet-101~\cite{he2016deep}  & $97.97$ & $82.58$ & $72.16$ & $75.10$ & $76.61$\\
Xception~\cite{chollet2017xception}  & $97.39$ & $82.06$ & $74.05$ & $72.91$ & $76.34$\\ 
 
EfficientNet-b0~\cite{tan2019efficientnet}  & $97.84$ & $82.77$ & $73.57$ & $76.76$ & $77.70$\\
EfficientNet-b3~\cite{tan2019efficientnet}  & $\textbf{\textit{98.35}}$ & $80.91$ & $74.05$ & $\textbf{\textit{79.08}}$ & $\textbf{\textit{78.01}}$\\
EfficientNet-b4~\cite{tan2019efficientnet}  & $97.55$ & $81.87$ & $74.42$ & $72.89$ & $76.39$\\
ViT-S~\cite{dosovitskiy2020vit}  & $97.61 $ & $82.96 $ & $\textbf{\textit{77.33}}$ & $71.14 $ & $77.14 $\\
ViT-B~\cite{dosovitskiy2020vit}  & $97.09$ & $\textbf{\textit{83.39}}$ & $76.77$ & $71.17$ & $77.11$\\

\hline
\textit{ResNet-50~(384)}  & $98.28$ & $82.20$ & $71.56$ & $77.27$ & $77.01$\\
\textit{Xception~(384)}  & $97.71$ & $79.29$ & $73.49$ & $77.33$ & $76.68$\\ 
\textit{EfficientNet-b4~(384)}  & $98.22$ & $78.98$ & $75.42$ & $73.44$ & $75.95$\\
\textit{Xception~(384)}  & $97.71$ & $79.29$ & $73.49$ & $77.33$ & $76.68$\\
\textit{ViT-B~(384)}  & $97.36$ & $84.50$ & $77.71$ & $72.56$ & $78.26$\\
\hline
\hline
\multicolumn{5}{l}{\multirow{1}{*}{\textit{\textbf{3.1 SSL-1K (ResNet-50)}}}}\\
\textcolor{gray}{Supervised}  & $\textcolor{gray}{97.95}$ & $\textcolor{gray}{81.24}$ & $\textcolor{gray}{72.23}$ & $\textcolor{gray}{74.21}$ & $\textcolor{gray}{75.89}$\\
Obow~\cite{gidaris2021obow}   & $98.39$ & $76.38$ & $\textbf{\textit{74.07}}$ & $78.54$ & $76.33$\\
VicReg~\cite{bardes2021vicreg}   & $98.20$ & $77.89$ & $72.47$ & $75.47$ & $75.28$\\
BYOL~\cite{grill2020bootstrap}    & $98.33$ & $81.88$ & $73.72$ & $70.39$ & $75.33$\\
SimSiam ~\cite{chen2021exploring}  & $\textbf{\textit{98.53}}$ & $81.37$ & $73.16$ & $\textbf{\textit{79.26}}$ & $77.93$\\
SimCLR ~\cite{chen2020simple}  & $98.50$ & $82.08$ & $73.33$ & $77.09$ & $77.50$\\
MoCo v2~\cite{chen2003improved}   & $98.37$ & $80.43$ & $73.16$ & $77.49$ & $77.03$\\
MoCo v3~\cite{chen2021empirical}   & $98.08$ & $\textbf{\textit{84.17}}$ & $73.64$ & $76.90$ & $\textbf{\textit{78.24}}$\\
\hline
\multicolumn{5}{l}{\multirow{1}{*}{\textit{\textbf{3.2 SSL-1K (ViT-B)}}}}\\
\textcolor{gray}{Supervised}   & $\textcolor{gray}{97.09}$ & $\textcolor{gray}{83.39}$ & $\textcolor{gray}{76.77}$ & $\textcolor{gray}{71.17}$ & $\textcolor{gray}{77.11}$\\
Moco V3~\cite{chen2021empirical}   & $98.29$ & $88.19$ & $79.28$ & $78.56$ & $82.01$\\
MAE~\cite{he2022masked}   & $98.31$ & $85.09$ & $76.62$ & $75.96$ & $79.22$\\
BEiT v2~\cite{peng2208beit}   &  $\textbf{\textit{99.16}}$ & \graycell{$\bm{89.48} \ $} & $\bm{81.00} $ & $77.61$ & $82.70$\\
BEiT v3~\cite{wang2023image}  & $98.76$ & $87.99$ & $80.02$ & $\textbf{\textit{80.61}}$ & $\textbf{\textit{82.87}}$\\
SimMIM~\cite{xie2022simmim}   & $98.34$ & $84.27$ & $77.29$ & $71.56$ & $77.71$ \\
DINO~\cite{caron2021emerging}   & $98.24$ & $84.56$ & $76.77$ & $74.48$ & $ 78.60$\\
ConMIM~\cite{yi2022masked}   & $98.40$ & $82.74$ & $78.14$ & $76.09$ & $78.99$ \\
\hline
\multicolumn{5}{l}{\multirow{1}{*}{\textbf{\textit{3.3 VLM and STM (ViT-B)}}}} \\
CLIP~\cite{radford2021learning} &$98.87$ & \graycell{$\bm{89.48} \ $} & $\textbf{78.86}$ & $77.43$ & $ \textbf{81.92}$ \\
BLIP~\cite{li2022blip} &$98.60$ & $87.90 $ & $78.21$ & $\textbf{77.82}$& $81.31$\\
SigLIP~\cite{10377550} &$98.71$ & $84.07$ & $78.34$ & $75.69$& $79.37$\\
VideoSwin~\cite{liu2021video} & $\bm{99.37}$ & $80.62$ & $77.58$ & $76.74$ & $78.31$  \\
VideoMAE~\cite{tong2022videomae}& $99.32 $ & $84.43$ & $74.24$ & $77.41$ & $78.69$\\
UMT~\cite{liu2022umt} & \graycell{$\bm{99.43} \ $} & $85.22$& $74.91$ & $76.15$ & $78.76$\\
Uniformer v2~\cite{li2022uniformerv2} & $99.19$ & $ 84.42$ & $75.94$ & $75.72$ & $78.69$ \\
\hline
\multicolumn{5}{l}{\multirow{1}{*}{\textbf{\textit{3.4 SSL-Face (ViT-B)}}}}\\
MoCo\_Face   & $98.41$  & $86.30$  & $78.51$  & $\textbf{\textit{83.38}} $ & \ $82.73 $\\
MAE\_Face   & $98.22$  & $85.43$  & $78.33$  & $80.02$ &\  $81.26$\\
BEiT\_Face   & $\textbf{\textit{98.93}} $  & $\textbf{\textit{88.62}} $  & $\textbf{\textit{80.99}}$  & $81.71$ & $\bm{83.77}$ \\

\hline
\multicolumn{5}{l}{\multirow{1}{*}{\textbf{\textit{3.5 SSL-1K}\&\textit{Face (ViT-B)}}}}\\
FaceMoCo   & $98.65$  & $84.98$  & $79.13$  & $\bm{85.72} $ & \ $ 83.28 $\\
FaceMAE   & $98.37$  & $85.85$  & $78.04$  & $78.07$ & \ $80.65$\\
FaceBEiT (FB)   & $\textbf{\textit{99.14}} $  & $\bm{89.18} $  & \graycell{$\bm{84.02} \ $}  & \graycell{$\bm{86.48} \ $} & \graycell{$\bm{86.56} \ $} \\

		\bottomrule
		\end{tabular}
	\label{tab:backbones}
\end{table}

\textbf{Learning Approaches.} We employ mainstream backbones pre-trained with SL and SSL on ImageNet-1K (\ie, \textbf{\textit{SL-1K}} and \textbf{\textit{SSL-1K}}) to evaluate the impact of learning approaches in FFD. As shown in Table~\ref{tab:backbones}, \textbf{\textit{SL-1K}} achieves over a $10$\% performance improvement compared to unpre-trained backbones, demonstrating the indispensability of pre-trained backbones with better image feature representation. Besides, as shown by \textit{ViT-B (384)} in \textbf{\textit{SL-1K}}, increasing the input image resolution from $224$ to $384$ benefits FFD performance but also increases computational cost.

Moreover, we employ ResNet-50 and ViT-B as backbone architectures to evaluate the impact of pre-training with different SSL methods. The comparison results in \textbf{\textit{3.1}} and \textbf{\textit{3.2}} of Table~\ref{tab:backbones} indicate that SSL gradually surpasses SL in FFD performance with the improvement of pre-training methods, such as the momentum encoder/contrast in CL equipped in SimSiam, SimCLR, and MoCo, and the MIM pretext task equipped in MAE, BEiT, and ConMIM. The results of \textbf{\textit{SSL-1K}} highlight the effectiveness of introducing SSL pre-training backbones and the necessity of exploring image feature representation approaches tailored for FFD.

\textbf{Network Architectures.}
We then evaluate the effect of different network architectures, which serve as the backbone's infrastructure. The comparison results in \textbf{\textit{SL-1K}} of Table~\ref{tab:backbones} show that the size of the network parameters under the same architecture has little impact on FFD performance. Meanwhile, ViT significantly outperforms ResNet in \textbf{\textit{SSL-1K}}, \eg, MoCo v3 with ViT-B surpasses ResNet-50 by $3.77$\% on average AUC across three cross-datasets. Thus, an excellent network architecture combined with an appropriate learning approach will significantly promote the backbone's baseline performance in FFD.

\textbf{Pre-training Datasets.}
We construct a large-scale real-face dataset to pre-train backbones, aiming to enhance their ability to represent facial features by alleviating the gap between nature images and facial images. We choose pre-training methods of MoCo v3, MAE, and BEiT v2, which are prominent in CL and MIM for single-modal pre-training and also perform excellently in FFD. We did not use BEiT v3 because its pre-training requires an image-text dataset.

In addition, we evaluate vision–language backbones (CLIP~\cite{radford2021learning}, BLIP~\cite{li2022blip}, SigLIP~\cite{10377550}) pre-trained on larger-scale datasets such as WIT~\cite{radford2021learning} and WebLI~\cite{10377550}, as well as spatio-temporal backbones (VideoSwin~\cite{liu2021video}, VideoMAE~\cite{tong2022videomae}, UMT~\cite{liu2022umt}, Uniformer v2~\cite{li2022uniformerv2}) with supervised pre-training on Kinetics~\cite{kay2017kinetics}.

The comparison results in Table~\ref{tab:backbones} indicate that FFD performance of all three methods is superior when pre-trained on the real-face dataset (\textbf{\textit{3.4 SSL-Face}}) compared to ImageNet-1K (\textbf{\textit{3.2 SSL-1K}}) and larger web-scale datasets (\textbf{\textit{3.3 VLM and STM}}), despite the real-face dataset being half the size of ImageNet-1K. Meanwhile, FaceMoCo and FaceBEiT pre-trained on ImageNet-1K achieve more significant improvements when further pre-trained on the real-face dataset (\textbf{\textit{3.5 SSL-1K}\&\textit{Face}}). Remarkably, FaceBEiT (\textbf{FB}) improves from $82.70$\% to $86.56$\% (Avg. AUC$\uparrow$: \textbf{\textit{3.2}} vs. \textbf{\textit{3.5}}), demonstrating that its tokenizer classification pretext task can better represent local facial features within faces for identifying forgery cues.

Extensive empirical studies on backbones demonstrate that embedding both general and face-specific knowledge through SSL pre-training methods is crucial for enhancing facial representation capabilities, thereby improving the fundamental generalization in FFD models.

\subsection{Comparison with State-of-the-arts}

We move on to our competitive backbone fine-tuning framework, which enhances the backbone's forgery detection capabilities through the competitive learning mechanism. We compare our model's generalization and adaptability with state-of-the-art FFD methods, such as FTCN~\cite{zheng2021exploring}, SBI~\cite{shiohara2022detecting}, RECCE~\cite{cao2022end}, CORE~\cite{ni2022core}, UIA-ViT~\cite{zhuang2022uia}, LSDA~\cite{yan2023transcending}, Choi~\etal~\cite{choi2024exploiting}, LLA+SBI~\cite{Nguyen_2024_CVPR}, Yan~\etal~\cite{yan2025generalizing}, DFD-FCG~\cite{han2025towards}, M2F2-Det~\cite{M2F2_Det_xiao}, and TFCU~\cite{guo2025face}. Furthermore, our framework fine-tunes FB and FB++ to obtain Competitive FaceBEiT (\textbf{CFB}) and Competitive FaceBEiT++ (\textbf{CFB++}), respectively. FB is pre-trained on ImageNet-1K and our real-face dataset, whereas FB++ extends FB with additional real-face pre-training. We also introduce SBI to verify its data-augmentation performance.


\subsubsection{Generalization Evaluation}

We first evaluate the generalizability of various FFD methods on $5$ classic cross-datasets and present the average AUC across these datasets for a clear comparison. Table~\ref{tab:sota} shows the quantitative results, as can be seen: (1) previous FFD methods exhibit varying and unsatisfactory performance on cross-datasets with diverse forgery patterns, indicating that generalization to unseen cross-datasets remains a significant challenge. (2) Both CFB and TFCU~\cite{guo2025face} utilize FB as the backbone and achieve relatively stable performance across $5$ common cross-datasets (all exceeding $84$\%), significantly outperforming existing methods. These results highlight the importance of backbone facial representation capability for effective fine-tuning in FFD methods. It is worth noting that TFCU relies on video temporal information, while our models only requires spatial information from a single frame, enabling support for both image and video inputs. (3) CFB++ achieves further performance gains by leveraging FB++ pre-trained on a larger real-face dataset and achieves state-of-the-art performance when combined with the SBI data augmentation strategy (CFB++ $w/$ SBI), especially on the more diverse and complex forgery patterns in DFDC and FFIW, which pose greater challenges and differ significantly from the within-dataset FF++. These results demonstrate that our fine-tuning framework effectively equips the backbone with a powerful ability to uncover implicit forgery cues across different face synthesis algorithms. For more analytical experiments, see Section~\ref{sec:facial-representation}.

\begin{table}[t]
    \footnotesize
    \setlength{\tabcolsep}{1.0mm}
    \caption{Quantitative generalization comparison of different FFD methods. ``†'' denotes the author's trained model, ``‡''indicates re-implementation using public code, and ``*'' signifies results from the original paper. \textbf{Bold} and \underline{underline} represent the best and second-best, respectively.}
    \renewcommand{\arraystretch}{1.1}
	\centering
	\midsepremove
		\begin{tabular}{l| c| c |c |c |c |c }
        \hline
\multirow{2}{*}{Method} &\multicolumn{6}{c}{\multirow{1}{*}{Cross-set AUC$\uparrow$ (\%)}}  \\
\cline{2-7}
  & CDF & DFDC & FFIW & DFDCP & DFD & Avg.\\ 
  \midrule  
FTCN\textsuperscript{†}~\cite{zheng2021exploring}$^{21'}$  &  $87.88$  & $71.31$  & $87.62$  &$ 70.16 $ & $ 89.73 $ & $ 81.34  $ \\
SBI\textsuperscript{†}~\cite{shiohara2022detecting}$^{22'}$  & $92.36$  & $71.30$  & $85.51$ & $ 83.83 $ & $ 88.11 $ &  $   84.22$\\
RECCE\textsuperscript{‡}~\cite{cao2022end}$^{22'}$ & $71.58$  & $64.53$  & $73.51$ & $ 67.14$ & $ 75.08 $ & $  70.37 $\\
CORE\textsuperscript{‡}~\cite{ni2022core}$^{22'}$ & $80.10$  & $70.53$  & $78.42$ & $ 72.19 $ &  $ 92.05$ &  $ 78.66  $\\
UIA-ViT\textsuperscript{‡}~\cite{zhuang2022uia}$^{22'}$  & $83.75$  & $74.34$  & $75.82$ & $79.11$ & $ 87.25 $ &  $ 80.05   $\\
D-adv\textsuperscript{‡}~\cite{wang2022deepfake}$^{22'}$  & $82.77$  & $73.91$  & $72.84$ &$ 77.53$ &$83.74 $ &  $ 78.16 $\\
UCF\textsuperscript{†}~\cite{yan2023ucf}$^{23'}$  & $77.45$  & $70.68$  & $66.16$ & $71.77 $ & $ 85.19 $ &  $ 74.25 $\\
CADDM\textsuperscript{†}~\cite{dong2023implicit}$^{23'}$  & $84.39$  & $70.46$  & $85.72$ & $ 63.06$ & $89.09$ &  $ 79.07 $\\
AltFreezing\textsuperscript{†}~\cite{wang2023altfreezing}$^{23'}$  & $89.23$  & $72.38$  & $75.79$ &$ 74.92 $ &$ 87.87 $ &  $ 80.04 $ \\
CFM\textsuperscript{†}~\cite{luo2023beyond}$^{23'}$  & $83.94$ & $ 74.17$ & $ 70.39$ & $79.25$ & $92.57$ &   $ 80.07$\\
LLA $w/$ SBI~\textsuperscript{†}~\cite{Nguyen_2024_CVPR}$^{24'}$  & $ 92.51 $  & $ 70.21$  & $ 69.04 $ &$ 83.44 $ & $ 88.07 $ &  $ 80.65$\\
LSDA\textsuperscript{*}~\cite{yan2023transcending}$^{24'}$  & $91.10$  & $77.00$  & $-$ & $-$ & $-$& $-$\\
Choi~\etal\textsuperscript{*}~\cite{choi2024exploiting}$^{24'}$  & $89.00$  & $-$  & $-$ & $-$& $96.10$ & $-$ \\
Yan~\etal\textsuperscript{*}~\cite{yan2025generalizing}$^{25'}$ & $ 94.70$& $ 84.30$  & $-$ & $-$ & $-$ & $-$ \\
DFD-FCG\textsuperscript{*}~\cite{han2025towards}$^{25'}$ & $ 95.00$& $ 81.80 $& $-$& $-$& $-$& $-$ \\
M2F2-Det\textsuperscript{‡}~\cite{M2F2_Det_xiao}$^{25'}$ & $ 83.22$& $  76.46 $& $ 81.78 $&  $ 71.22 $& $88.98 $& $ 80.33  $ \\

TFCU\textsuperscript{†}~\cite{guo2025face}$^{25'}$ & $ 92.27$& $  85.81 $& $ 90.35 $&  $ 86.25$& $95.85 $& $ 90.10  $ \\


 \hline

CFB & $90.46$  & $84.90$  & $90.97$ & $ 87.16 $ & $ \underline{96.21} $  & $ 89.94 $\\
CFB++ & $90.99$ & $\underline{86.79}$ & $90.72$ & $87.17$ & $\textbf{96.52}$ & $ 90.44$   \\
CFB \textit{w/} SBI   & $\underline{95.35}$  & $85.45$  & $\underline{91.82}$ & $\underline{94.13}$ & $96.04$ &  $ \underline{92.56} $\\
CFB++ \textit{w/} SBI & $\textbf{95.46}$ & $\textbf{88.66}$ & $\textbf{92.31}$ & $\textbf{94.42}$ & $95.92$ & $\textbf{93.35}$  \\

		\bottomrule
		\end{tabular}
	\label{tab:sota}
\end{table}

\begin{table}[t]
    \footnotesize
    \setlength{\tabcolsep}{0.40mm}
    \caption{Quantitative generalization comparison of FFD methods across various forgery synthesis algorithms, including representative GAN-based and Diffusion Model-based methods covering four types of forgeries. Results at the bottom of this table come from FFD methods fine-tuned on $20$ StyleGAN3-generated forged faces.}
	\centering
	\midsepremove
    \renewcommand{\arraystretch}{1.0}
		\begin{tabular}{ 
        l !{\vrule height 1\ht\strutbox depth 2.75pt}  
        *{6}{c !{\vrule height 1\ht\strutbox depth 2.75pt}}
        c 
    }
        \hline
    Forged Faces & RECCE & CORE & D-adv & CADDM & BEiT v2 & FB & CFB  \\
        \hline
    
    \multicolumn{8}{l}{\multirow{1}{*}{\textit{\textbf{1. Entire Face Synthesis}}}}\\
    DCGAN~\cite{radford2015unsupervised} & $85.28$ & $81.03$ & $90.87$ & $85.67$  & $91.93$ & $\textbf{99.94}$ & $\underline{99.48}$ \\
    StyleGAN3~\cite{Karras2021} & $\underline{57.25}$ & $35.69$ & $33.06$ & $45.15$ & $51.13$ & $57.24$ & $\textbf{60.75}$ \\
    SD-1.5~\cite{rombach2022high} & $82.46$ & $66.15$ & $66.23$ & $49.49$ & $80.15$ & $\underline{83.18}$ & $\textbf{86.84}$ \\
    
\hline
    \multicolumn{8}{l}{\multirow{1}{*}{\textit{\textbf{2. Face Attribute Editing}}}}\\
    StarGAN2~\cite{choi2020starganv2} & $54.71$ & $26.62$ & $28.20$ & $55.92$ & $66.03$ & $\underline{73.56}$ & $\textbf{79.33}$ \\
    AttnGAN~\cite{xu2018attngan} & $60.35$ & $52.13$ & $66.42$ & $64.99$ & $69.87$ & $\textbf{81.15}$ & $\underline{80.11}$ \\
    \hline
    \multicolumn{8}{l}{\multirow{1}{*}{\textit{\textbf{3. Face Swapping}}}}\\
    E4S~\cite{liu2023fine} & $65.32$ & $32.43$ & $37.03$ & $46.45$ & $66.15$ & $\underline{81.30}$ & $\textbf{85.02}$ \\
    DiffFace~\cite{kim2022diffface} & $50.07$ & $55.08$ & $49.66$ & $53.42$ & $61.28$ & $\underline{81.27}$ & $\textbf{82.94}$ \\
    \hline
    \multicolumn{8}{l}{\multirow{1}{*}{\textit{\textbf{4. Face Reenactment}}}}\\
    Wav2Lip~\cite{prajwal2020lip} & $72.71$ & $78.43$ & $73.00$ & $69.48$ & $96.92$ & $\underline{98.96}$ & $\textbf{99.00}$ \\
     Diffused & \multirow{2}{*}{$79.12$} & \multirow{2}{*}{$76.55$} & \multirow{2}{*}{$84.08$} & \multirow{2}{*}{$85.76$} & \multirow{2}{*}{$94.46$} & \multirow{2}{*}{$\textbf{97.58}$} & \multirow{2}{*}{$\underline{96.44}$} \\
    Heads~\cite{stypulkowski2024diffused} & & & & & & &  \\
    
\hline
\textbf{Avg. AUC (\%)} & $66.82$ & $53.21$ & $56.94$ & $60.86$ & $72.63$ & $\underline{81.90}$ & $\textbf{83.86}$ \\
    \hline
    Forged Faces & \textit{RECCE} & \textit{CORE} & \textit{D-adv} & \textit{CADDM} & \textit{BEiT v2} & \textit{FB} & \textit{CFB}  \\
    \midrule  
    StyleGAN3~\cite{Karras2021} & $\underline{72.09}$ & $45.21$ & $49.05$ & $-$ & $57.12$ & $71.09$ & $\textbf{82.20}$ \\
    \bottomrule
		\end{tabular}
	\label{tab:gan-sd-based}
\end{table}

We evaluate different FFD models on the $9$ cross-datasets we created. Table~\ref{tab:gan-sd-based} shows that both our baseline FB and CFB outperform other FFD methods, emphasizing the importance of a backbone with real face knowledge and the effectiveness of competitive forgery cue extraction. However, all listed FFD methods in Table~\ref{tab:gan-sd-based} exhibit poor generalization performance on StyleGAN3-generated faces, mainly due to unseen entire face forgeries during training and the high realism of these fakes. Therefore, we employ a small number of StyleGAN3-generated forged faces ($20$ images) to fine-tune different methods further, enabling verification of their rapid adaptability to unseen forgery techniques. The bottom row of Table~\ref{tab:gan-sd-based} presents the adaptability performance, where \textit{RECCE}~\cite{cao2022end} and \textit{FB} show greater improvement by modeling real face features, while \textit{CFB} demonstrates even more impressive performance with a $21.45$\% improvement, proving its superior practicality. Additionally, we also validate the superiority of our method on DF40 across various face forgery algorithms. As shown in Table~\ref{tab:sota2}, our method achieves state-of-the-art performance with an average AUC improvement of $2.91$\%.

\begin{table*}[t]
    \footnotesize
    \setlength{\tabcolsep}{1.1mm}
    \caption{Quantitative generalization comparison of various FFD methods on the DF40 dataset with different face forgery algorithms~\cite{yan2024df40}. ``†'' denotes the author's trained model, and ``‡''indicates re-implementation using public code.}
    
    \renewcommand{\arraystretch}{1.1}
	\centering
	\midsepremove
  \begin{tabular}{l| c c c c c c c |c}
    \hline
   Method& SadTalker\cite{zhang2023sadtalker} & FOMM\cite{siarohin2019first} & FaceDancer\cite{rosberg2023facedancer} & MobileSwap\cite{xu2022mobilefaceswap} & SimSwap\cite{chen2020simswap} & InSwapper\cite{InSwapper} & UniFace\cite{xu2022designing}& Avg. \\
    \hline
FTCN\textsuperscript{†}~\cite{zheng2021exploring}  & $87.76$ & $ 87.23$ & $ 85.08 $ & $ 77.46 $ & $ 94.56 $ & $ 97.12 $ & $ 98.63 $ & $ 89.69 $\\
SBI\textsuperscript{†}~\cite{shiohara2022detecting}    & $78.51$ & $99.12$ & $ 76.43 $ & $ \underline{99.23}$& $ 95.54$ & $ 89.54$ & $ 93.73$ & $ 90.30 $\\
RECCE\textsuperscript{‡}~\cite{cao2022end}   & $ 86.24$ & $ 98.52 $ & $ 81.94$ & $ 96.87 $  & $ 80.28$  & $ 93.39 $  & $ 93.59 $ & $ 90.12 $ \\
CORE\textsuperscript{‡}~\cite{ni2022core}    & $ 87.59$  & $ 98.48$ & $ 86.05$ & $ 99.12$ & $ 91.19$ & $ 93.83$ & $ 95.40$& $ 93.09$\\
UIA-ViT\textsuperscript{‡}~\cite{zhuang2022uia}  & $ 76.50$ & $ 93.64 $ & $ 84.07$ & $ 89.87 $  & $ 72.37$  & $ 89.77 $  & $ 89.78 $  & $ 85.14$    \\
D-adv\textsuperscript{‡}~\cite{wang2022deepfake}  & $86.19$ & $98.88$ &  $80.23$ & $95.44$ &$81.62$ &$88.36$ &$93.47$ &$89.17$\\
UCF\textsuperscript{†}~\cite{yan2023ucf}   & $ 71.27$ & $ 85.10$ & $ 83.87$ & $ 90.47$ & $ 61.64$ & $ 80.72$ &$ 80.18$ &$ 79.04$ \\
CADDM\textsuperscript{†}~\cite{dong2023implicit}  & $54.64$ & $79.36$ & $71.56$ & $96.62$ & $92.32$ & $75.73$& $89.84$& $80.01$ \\
AltFreezing\textsuperscript{†}~\cite{wang2023altfreezing}   & $88.28$  & $76.47$ & $87.75$  & $87.66$ &$82.84$ & $88.88$ & $86.92$& $ 85.54$ \\
CFM\textsuperscript{†}~\cite{luo2023beyond}  & $83.87$ & $ 98.66$ & $92.65 $& $98.63 $& $89.37 $& $93.60 $& $96.40 $& $93.31 $ \\

LLA+SBI~\textsuperscript{†}~\cite{Nguyen_2024_CVPR}  & $ 82.14$& $ 92.85$& $ 87.14$ & $ 97.15$& $ 92.12$& $ 90.00$& $ 90.71$& $ 90.30$ \\
M2F2-Det\textsuperscript{‡}~\cite{M2F2_Det_xiao}
& $ 83.61$& $ 97.53$& $ 93.55$&$ 98.33$&$ 92.18$& $ 94.83$ & $ 97.35$& $93.91$\\
TFCU\textsuperscript{‡}~\cite{guo2025face}  & $\underline{88.34}$ & $\underline{99.29}$ & $94.66$ & $98.84$& $94.99$ & $96.61$ & $98.77$ & $95.93$  \\

     \midrule
     CFB   & $ 87.94 $ & $ 99.15 $ & $ \textbf{95.23} $ & $ 99.18 $ & $ \textbf{95.92} $ & $ \textbf{97.22} $ & $ \textbf{98.93} $ & $ \underline{96.22}$\\

     CFB++   & $ \textbf{88.98} $ & $ \textbf{99.39} $ & $ \underline{94.93} $ & $ \textbf{99.52} $ & $ \underline{95.76} $ & $ \underline{97.14} $ & $ \underline{98.91} $ & $ \textbf{ 96.38 }$\\     
    \bottomrule

  \end{tabular}
  \label{tab:sota2}
\end{table*}

\subsubsection{Adaptation Evaluation}

\begin{table*}[t]
    \footnotesize
    \setlength{\tabcolsep}{0.9mm}
    \renewcommand\arraystretch{0.75}
    \caption{Quantitative adaptation comparison of different FFD models by further fine-tuning on a few images (\textit{$10$-Shot}, \textit{$20$-Shot}) from unseen datasets. These models have been fine-tuned on the source dataset (FF++). \colorbox{gray!20}{\textbf{bold}} means the best, and \textbf{bold} means the second-best. $\uparrow$ and $\downarrow$ indicate an increase and decrease relative to the baseline (\textcolor{gray}{gray}), respectively. Higher values on the diagonal represent greater performance and adaptation.}

	\centering
	\midsepremove
		\begin{tabular}{c| c | c c c c c|c c c c c c}
		\toprule
        \multirow{3}{*}{Method} & \multirow{1}{*}{Training Set} &\multicolumn{5}{c|}{\multirow{1}{*}{\textit{$10$-Shot} \ \ Testing Set AUC$\uparrow$ (\%)}}   &\multicolumn{5}{c}{\multirow{1}{*}{\textit{$20$-Shot} \ \  Testing Set AUC$\uparrow$ (\%)}}  \\[0.5ex]
                    \cline{3-12}
         \rule{0pt}{7.5pt}                
         & (\textit{$10$}/\textit{$20$-Shot}) &FF++& CDF & DFDC & FFIW & Avg.  & FF++ & CDF & DFDC & FFIW & Avg.  \\
	    \hline
\multirow{5}{*}{RECCE~\cite{cao2022end}} 
\rule{0pt}{7.5pt}
&\textcolor{gray}{FF++} &   $ \textcolor{gray}{97.91}\ \ \ $  & $\textcolor{gray}{71.58}\ \ \ $ &  $\textcolor{gray}{64.53}\ \ \ $  & $\textcolor{gray}{73.51}\ \ \ $ & $\textcolor{gray}{76.88}\ \ \ $  &  $\textcolor{gray}{97.91}\ \ \  $  & $\textcolor{gray}{71.58}\ \ \ $ &  $\textcolor{gray}{64.53} \ \ \ $  & $\textcolor{gray}{73.51}\ \ \ $ & \ $\textcolor{gray}{76.88}\ \ \ $ \\[0.5ex]
&CDF & $89.25\downarrow$ & $ 67.78\downarrow$ & $ 65.45\uparrow$ & $ 66.59\downarrow$ & $ 72.27 \downarrow$  & $93.58\downarrow $ & $70.45\downarrow $ & $61.95\downarrow $ & $67.50\downarrow $ & $ 73.37\downarrow$\\[0.5ex]
&DFDC & $92.36\downarrow $ & $67.00 \downarrow$ & $67.34 \uparrow$ & $77.36 \uparrow$ & $ 76.02\downarrow$  & $ 90.31\downarrow$ & $68.15\downarrow $ & $67.00 \uparrow$ & $77.36\uparrow $ & $ 75.71\downarrow$\\[0.5ex]
&FFIW & $97.66\downarrow$ & $67.73\downarrow $ & $65.22 \uparrow$ & $73.96\uparrow $ & $ 76.14\downarrow$  & $94.62\downarrow $ & $70.31 \downarrow$ & $66.55 \uparrow$ & $77.49\uparrow $ & $ 77.24\uparrow$\\
	    \hline 
\multirow{5}{*}{CORE~\cite{ni2022core}} 
\rule{0pt}{7.5pt}
&\textcolor{gray}{FF++} &  $\textcolor{gray}{97.66}\ \ \ $  & $\textcolor{gray}{80.10}\ \ \ $ &  $\textcolor{gray}{70.53}\ \ \ $  & $\textcolor{gray}{78.42}\ \ \ $ & $ \textcolor{gray}{81.68}\ \ \ $  &  $\textcolor{gray}{97.66}\ \ \ $  & $\textcolor{gray}{80.10}\ \ \ $ &  $\textcolor{gray}{70.53}\ \ \ $  & $\textcolor{gray}{78.42}\ \ \ $ & $\  \textcolor{gray}{81.68}\ \ \ $  \\[0.5ex]

&CDF & $97.77\uparrow$ & $81.49\uparrow $ & $68.91\downarrow $ & $77.78\downarrow $ & $81.49\downarrow $  & $97.74\downarrow$ & $ 81.77\uparrow$ & $68.99\downarrow $ & $79.85 \uparrow$ & $82.09\uparrow$\\[0.5ex]
&DFDC & $97.22 \downarrow$ & $79.06\downarrow $ & $74.33\uparrow $ & $79.85 \uparrow$ & $82.62\uparrow $  & $97.14\downarrow $ & $78.77 \downarrow$ & $74.93\uparrow $ & $79.92\uparrow $ & $82.69 \uparrow$\\[0.5ex]
&FFIW & $97.51\downarrow$ & $81.85 \uparrow$ & $71.21\uparrow $ & $78.32\downarrow $ & $82.22\uparrow$  & $97.54\downarrow $ & $80.82\uparrow $ & $71.34\uparrow $ & $ 79.14\uparrow$ & $82.21\uparrow$\\[0.5ex]
	    \hline
\multirow{5}{*}{D-adv~\cite{wang2022deepfake}} 
\rule{0pt}{7.5pt}
&\textcolor{gray}{FF++} &  $\textcolor{gray}{98.31}\ \ \ $  & $ \textcolor{gray}{82.77}\ \ \ $ & $ \textcolor{gray}{73.91}\ \ \ $  & $\textcolor{gray}{72.84}\ \ \ $ & $\textcolor{gray}{81.96} \ \ \ $  &  $\textcolor{gray}{98.31} \ \ \ $  &$ \textcolor{gray}{82.77}\ \ \ $ &  $ \textcolor{gray}{73.91}\ \ \ $  & $\textcolor{gray}{72.84}\ \ \ $ & $\ \textcolor{gray}{81.96}\ \ \ $  \\[0.5ex]
&CDF & $98.40\uparrow$ & $82.89\uparrow$ & $ 71.42\downarrow$ & $ 68.35\downarrow$ & $80.27\downarrow$  & $98.41\uparrow$ & $ 85.05\uparrow$ & $ 71.21\downarrow$ & $67.91\downarrow $ & $80.65\downarrow$\\[0.5ex]
&DFDC & $98.47\uparrow$ & $ 80.80\downarrow$ & $ 75.05\uparrow$ & $ 69.76\downarrow$ & $81.02\downarrow$  & $ 98.34\uparrow$ & $ 81.29\downarrow$ & $ 75.27\uparrow$ & $ 68.60\downarrow$ & $80.88\downarrow$\\[0.5ex]
&FFIW & $98.47\uparrow$ & $82.90\uparrow $ & $73.45\downarrow $ & $68.25\downarrow$ & $80.77\downarrow$ & $ 98.35\uparrow$  & $82.04\downarrow $ & $73.80\downarrow $ & $69.46 \downarrow$ & $ \ 80.91\downarrow$ \\[0.5ex]
	    \hline

\multirow{5}{*}{M2F2-Det~\cite{M2F2_Det_xiao}} 
\rule{0pt}{7.5pt}
&\textcolor{gray}{FF++} & $\textcolor{gray}{99.13} \ \ \ $  & $\textcolor{gray}{83.22}\ \ \ $ & $\textcolor{gray}{76.46}\ \ \ $ & $\textcolor{gray}{81.78}\ \ \ $ & $\textcolor{gray}{85.15}\ \ \ $& $\bm{\textcolor{gray}{99.13}} \ \ \ $  & $\textcolor{gray}{83.22}\ \ \ $ & $\textcolor{gray}{76.46}\ \ \ $ & $\textcolor{gray}{81.78}\ \ \ $ & $\textcolor{gray}{85.15}\ \ \ $\\ [0.5ex]
&CDF & $98.26\downarrow$ & $80.22\downarrow$ & $70.64\downarrow$ & $72.67\downarrow$ & $80.40\downarrow$ & $98.00\downarrow$ & $78.69\downarrow$ & $68.02\downarrow$ & $71.05\downarrow$ & $78.94\downarrow$\\[0.5ex]
&DFDC & $96.91\downarrow $ & $81.68\downarrow$ & $72.54\downarrow$ & $77.71\downarrow$ & $82.21\downarrow$ & $96.52\downarrow$ & $81.15\downarrow$ & $70.94\downarrow$ & $81.06\downarrow$ & $82.42\downarrow$\\[0.5ex]
&FFIW & $94.53\downarrow$ & $78.98\downarrow$ & $74.28\downarrow$ & $89.66\uparrow$ & $84.36\downarrow$ & $92.93\downarrow$ & $77.99\downarrow$ & $74.02\downarrow$ & $90.06\uparrow$ & $83.75\downarrow$\\[0.25ex]
	    \hline

\multirow{5}{*}{BEiT v2} 
\rule{0pt}{7.5pt}
&\textcolor{gray}{FF++} & $\textcolor{gray}{99.16} \ \ \ $  & $\textcolor{gray}{89.48}\ \ \ $ & $\textcolor{gray}{81.00}\ \ \ $ & $\textcolor{gray}{77.61}\ \ \ $ & $\textcolor{gray}{86.81}\ \ \ $& $\bm{\textcolor{gray}{99.16}} \ \ \ $  & $\textcolor{gray}{89.48}\ \ \ $ & $\textcolor{gray}{81.00}\ \ \ $ & $\textcolor{gray}{77.61}\ \ \ $ & $\textcolor{gray}{86.81}\ \ \ $\\ [0.5ex]
&CDF & $99.09\downarrow$ & $89.26\downarrow$ & $79.88\downarrow$ & $73.21\downarrow$ & $85.36\downarrow$ & $98.88\downarrow$ & $91.94\uparrow$ & $80.81\downarrow$ & $76.52\downarrow$ & $87.04\uparrow$\\[0.5ex]
&DFDC & $99.15\downarrow $ & $90.21\uparrow$ & $82.27\uparrow$ & $79.06\uparrow$ & $87.67\uparrow$ & $99.15\downarrow$ & $90.50\uparrow$ & $82.91\uparrow$ & $79.31\uparrow$ & $87.97\uparrow$\\[0.5ex]
&FFIW & $99.10\downarrow$ & $90.73\uparrow$ & $82.27\uparrow$ & $85.51\uparrow$ & $89.40\uparrow$ & $98.95\downarrow$ & $91.34\uparrow$ & $82.17\uparrow$ & $90.67\uparrow$ & $90.78\uparrow$\\[0.25ex]
	    \hline        
     
\multirow{5}{*}{FB} 
\rule{0pt}{7.5pt}
&\textcolor{gray}{FF++}  & $\textcolor{gray}{99.14} \ \ \ $  & $\textcolor{gray}{89.13} \ \ \ $ & $\textcolor{gray}{84.02}\ \ \ $ &  $\textcolor{gray}{86.48}\ \ \ $ &$\textcolor{gray}{89.71} \ \ \ $  & $ \textcolor{gray}{99.14} \ \ \ $  &$\textcolor{gray}{89.13}\ \ \ $ & $\textcolor{gray}{84.02}\ \ \ $ &  $\textcolor{gray}{86.48}\ \ \ $ & $\textcolor{gray}{89.71}\ \ \ $\\ 
&CDF & $99.07\downarrow$ & $90.83\uparrow$ & $84.69\uparrow$ & $85.97\downarrow$ & $90.14\uparrow$ & $98.70\downarrow$ & \graycell{$\bm{92.25}\uparrow$} & $85.63\uparrow$ & $89.98\uparrow$ & $91.64\uparrow$\\[0.5ex]
&DFDC & $98.79\downarrow$ & $\bm{91.17}\uparrow$ & $\bm{86.63}\uparrow$ & $91.23\uparrow$ & $91.96\uparrow$ & $98.58\downarrow$ & $90.11\uparrow$ & $\bm{87.44}\uparrow$ & $89.94\uparrow$ & $91.52\uparrow$\\[0.5ex]
&FFIW & $98.98\downarrow$ & $89.21\downarrow$ & $86.31\uparrow$ & $91.83\uparrow$ & $91.58\uparrow$  & $98.97\downarrow$ & $90.60\uparrow$ & $87.27\uparrow$ & $\bm{95.14}\uparrow$ & $\bm{93.00}\uparrow$\\[0.25ex]
	    \hline
\multirow{5}{*}{CFB} 
&\textcolor{gray}{FF++} & \graycell{$\bm{\textcolor{gray}{99.36} }\ \ \ $}  & $\textcolor{gray}{90.46}\ \ \ $ & $\textcolor{gray}{84.90}\ \ \ $ & $\textcolor{gray}{90.97}\ \ \ $ & $\textcolor{gray}{91.42}\ \ \ $  & \graycell{$\bm{\textcolor{gray}{99.36} }\ \ \ $} & $\textcolor{gray}{90.46}\ \ \ $ & $\textcolor{gray}{84.90}\ \ \ $ & $\textcolor{gray}{90.97}\ \ \ $ & $\textcolor{gray}{91.42}\ \ \ $\\
&CDF & $\bm{99.25}\downarrow$ & \graycell{$\bm{91.33}\uparrow$} & $86.17\uparrow$ & $91.74\uparrow$ & $92.12\uparrow$  & $99.10\downarrow$ & $\bm{91.31}\uparrow$ & $85.48\uparrow$ & $91.82\uparrow$ & $91.93\uparrow$\\ 
&DFDC & $99.10\downarrow$ & $91.16\uparrow$ & \graycell{$\bm{87.72}\uparrow$} & $\bm{93.23}\uparrow$ & \graycell{$\bm{92.80}\uparrow$}  & $98.92\downarrow$ & $91.08\uparrow$ & \graycell{$\bm{88.07}\uparrow$} & $91.80\uparrow$ & $92.47\uparrow$\\
&FFIW & $99.22\downarrow$ & $90.14\uparrow$ & $85.78\uparrow$ & \graycell{$\bm{93.65}\uparrow$} & $\bm{92.20}\uparrow$  & $99.09\downarrow$ & $91.11\uparrow$ & $86.78\uparrow$ & \graycell{$\bm{96.65}\uparrow$} & \graycell{$\bm{93.41}\uparrow$}\\
		\bottomrule
		\end{tabular}
	\label{tab:few}
\end{table*}

Next, we evaluate the adaptation capability of various FFD methods by further fine-tuning FFD models using a few images from previously unseen common datasets, having initially trained these models on the FF++ dataset. Specifically, we conduct $10$-shot and $20$-shot experiments with an equal number of real and fake faces, fine-tuning the different FFD models for $2$ epochs with a learning rate of $2e-5$.

Table~\ref{tab:few} shows the quantitative comparison results, indicating that FB pre-trained on real-face dataset not only adapts well to the target dataset but also maintains stable performance in the source dataset (FF++), whereas previous methods suffer varying degrees of performance degradation. Meanwhile, our CFB performs best in target and unseen cross-datasets, demonstrating that our fine-tuning framework has superior practicability and robustness. Notably, CFB improves by $2.82$\% (AUC$\uparrow$: $84.90$\% vs. $87.72$\%) on DFDC with $5,000$ test images under the $10$-shot condition, and by $3.17$\% ($84.90$\% vs. $88.07$\%) on DFDC and $5.68$\% ($90.97$\% vs. $96.65$\%) on FFIW under the $20$-shot condition, highlighting the effectiveness of our expert knowledge-equipped backbone in identifying forgery cues and efficiently adapting to unknown forgery patterns.

\begin{table}[t]
    \footnotesize
    \setlength{\tabcolsep}{1.0mm}
    \caption{Quantitative comparison of our framework with different components. Without UFM, we sum features from dual branches. When $\mathcal{L}_{EDU}$ is not used, $\mathcal{L}_{ce}$ is used instead. The results of our auxiliary branch (CFB\_A) and both branches (CFB\_M+A) are presented as reference.}
	\centering
	\midsepremove
		\begin{tabular}{c| c c c| c c c| c c c  c }
		\toprule
           \multirow{2}{*}{Method} &   \multirow{2}{*}{$\mathcal{L}_{dec}$}&  \multirow{2}{*}{UFM} &\multirow{2}{*}{$\mathcal{L}_{EDU}$} &\multicolumn{3}{c|}{\multirow{1}{*}{Cross-set AUC$\uparrow$ (\%)}} 
           &\multicolumn{2}{c}{\multirow{1}{*}{ Avg.$\uparrow$ (\%)}}
           \\  
           \cline{5-10}
              &  &  &      & CDF & DFDC & FFIW & AUC & ACC  \\
	    \hline
 \multirow{2}{*}{FB}& & &   & $89.18$ & $84.02$ & $86.48$ & $86.56$ & $79.37$\\
  & &  & \checkmark&  $89.34$ & $84.19$ & $88.76$ & $87.43$ & $80.96$\\
\hline
\multirow{4}{*}{CFB}           &  &  & &  $89.77$ & $83.67$ & $87.69$ & $87.04$ & $79.26$\\
   &\checkmark                  &  &  &  $90.31$   & $84.89$ & $89.21$ & $88.14$ & $79.42$\\
    &\checkmark &\checkmark   & &  $\textbf{91.16}$ & $84.54$ & $90.22$ & $88.64$ & $80.84$\\
   &\checkmark &\checkmark &\checkmark &  $90.46$ & $\textbf{84.90}$ & $\textbf{90.97}$ & $\textbf{88.78}$ & $\textbf{82.29}$\\
\hline
CFB\_A&\checkmark &\checkmark &\checkmark &  $79.83$ & $68.29 $ & $68.97$ & $72.36$ & $ 67.96$\\
CFB\_M+A&\checkmark &\checkmark &\checkmark &  $90.45$ & $83.10$ & $89.72$ & $87.76$ & $ 82.04 $\\
		\bottomrule
		\end{tabular}
	\label{tab:ablation}
\end{table}

\subsection{Analysis of Backbone Fine-tuning Framework}
        \label{chap:ablation}

We then analyze the efficacy of our competitive backbone fine-tuning framework through the following aspects: (1) decorrelation loss $\mathcal{L}_{dec}$, (2) Uncertainty-based Fusion Module (UFM), (3) Evidence Deep Learning and Evidential Uncertainty Calibration losses $\mathcal{L}_{EDU}$, (4) our framework with different backbones, (5) attention map and feature visualizations, and (6) additional on real-face datasets.

Table~\ref{tab:ablation} shows the quantitative comparison results of core components in our framework, as can be seen: (1) without $\mathcal{L}_{dec}$, the dual-branch backbone tends to learn similar forgery cues, resulting in only a slight improvement over FB under the fusion module with sum operation; (2) incorporating $\mathcal{L}_{dec}$ can better promote the generalization of the main branch, with the average AUC improving from $87.04$\% to $88.14$\%; (3) our framework with UFM provides dynamic weights for feature fusion based on the prediction uncertainties of the main and auxiliary branches, enhancing their competition and improving backbone performance; (4) $\mathcal{L}_{EDU}$ further enhances the model's FFD capability by constraining prediction uncertainty and class probability.

Additionally, although the Avg. AUC$\uparrow$ performance improvement of our framework with UFM and $\mathcal{L}_{EDU}$ is not very large compared to not using them ($88.78$\% vs. $88.14$\%), we are the first to introduce model uncertainty estimation in FFD solutions and propose a novel uncertainty-based threshold optimization mechanism for FFD models during the \textit{inference} stage, thereby significantly enhancing the accuracy of FFD models in discriminant results (Avg. ACC$\uparrow$: $79.42$\% vs. $82.29$\%). Please refer to Section~\ref{sec:inference-experiment} for more analysis. Besides, we also present a comparison of the auxiliary branch (CFB\_A) and both branches (CFB\_M+A) in our framework. Both CFB\_A and CFB\_M+A perform lower than the main branch (CFB), demonstrating the effectiveness of our competitive learning mechanism, which significantly fosters the main backbone's generalization.

We further validate the flexibility of our competitive backbone fine-tuning framework by using backbones pre-trained with different learning approaches and datasets, including ViT with SL on ImageNet-1k (SL on 1K), FaceMAE, FaceMoCo, and FaceBEiT (FB). The quantitative comparison in Table~\ref{tab:ablation2} shows that our framework with different backbones achieves around $2$\% performance improvement, further demonstrating the robustness and potential of our competitive learning mechanism with the decorrelation constraint and uncertainty-based fusion module.

\begin{table}[t]
    \footnotesize
    \setlength{\tabcolsep}{1.0mm}
    \caption{Quantitative comparison of our framework with different backbones.}
	\centering
    \renewcommand\arraystretch{1.1}
	\midsepremove
		\begin{tabular}{l| c| c c c c c c }
		\toprule
\multicolumn{1}{l|}{\multirow{2}{*}{Method}}  & \multicolumn{1}{l|}{\multirow{1}{*}{In-set}} &\multicolumn{4}{c}{\multirow{1}{*}{Cross-set AUC$\uparrow$ (\%)}}  \\ 
\cline{2-6}
                  &  FF++  &  CDF  & DFDC &  FFIW & Avg.   \\
\hline

ViT (SL on 1k) &$97.09$ & $83.39$ & $76.77$ & $71.17$ & $77.11$\\
Competitive ViT (SL on 1k) &$97.10$ & $83.41$ & $78.33$ & $75.09$ & $78.94$\\
\hline
FaceMAE (1k\&Face) &$98.37$ & $85.85$ & $78.04$ & $78.07$ & $80.65$\\
Competitive FaceMAE (1k\&Face) &$97.97$ & $85.91$ & $79.05$ & $81.16$ & $82.04$\\
\hline
FaceMoCo (1k\&Face) &$98.65$ & $84.98$ & $79.13$ & $85.72$ & $83.28$\\
Competitive FaceMoCo (1k\&Face) &$98.48$ & $88.02$ & $79.17$ & $89.19$ & $85.46$\\
\hline
FaceBEiT (1k\&Face) &$99.14$ & $89.18$ & $84.02$ & $86.48$ & $86.56$\\
Competitive FaceBEiT (1k\&Face) &$\textbf{99.36}$ & $\textbf{90.46}$ & $\textbf{84.90}$ & $\textbf{90.97}$ & $\textbf{88.78}$\\
		\bottomrule
		\end{tabular}
	\label{tab:ablation2}
\end{table}

\begin{figure}[t]
\centering
\includegraphics[width=1.0\linewidth]{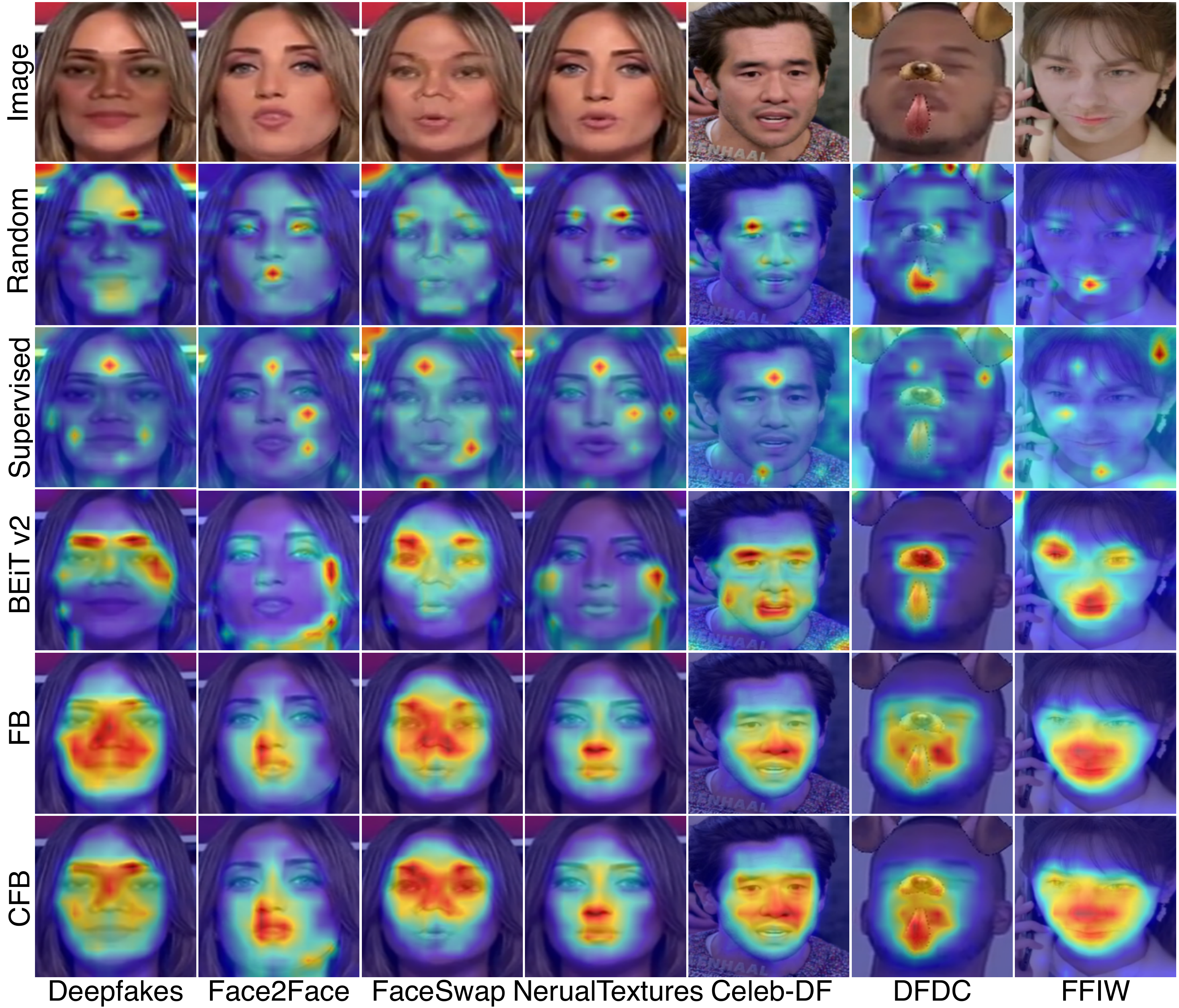}
\vspace{-15pt}
\caption{Visualization of attention maps for FFD methods on different forgery techniques. Highlighted areas indicate higher model attention.}
\label{fig:FFD-attnmap}
\end{figure}

\begin{figure}[t]
\centering
\includegraphics[width=\linewidth]{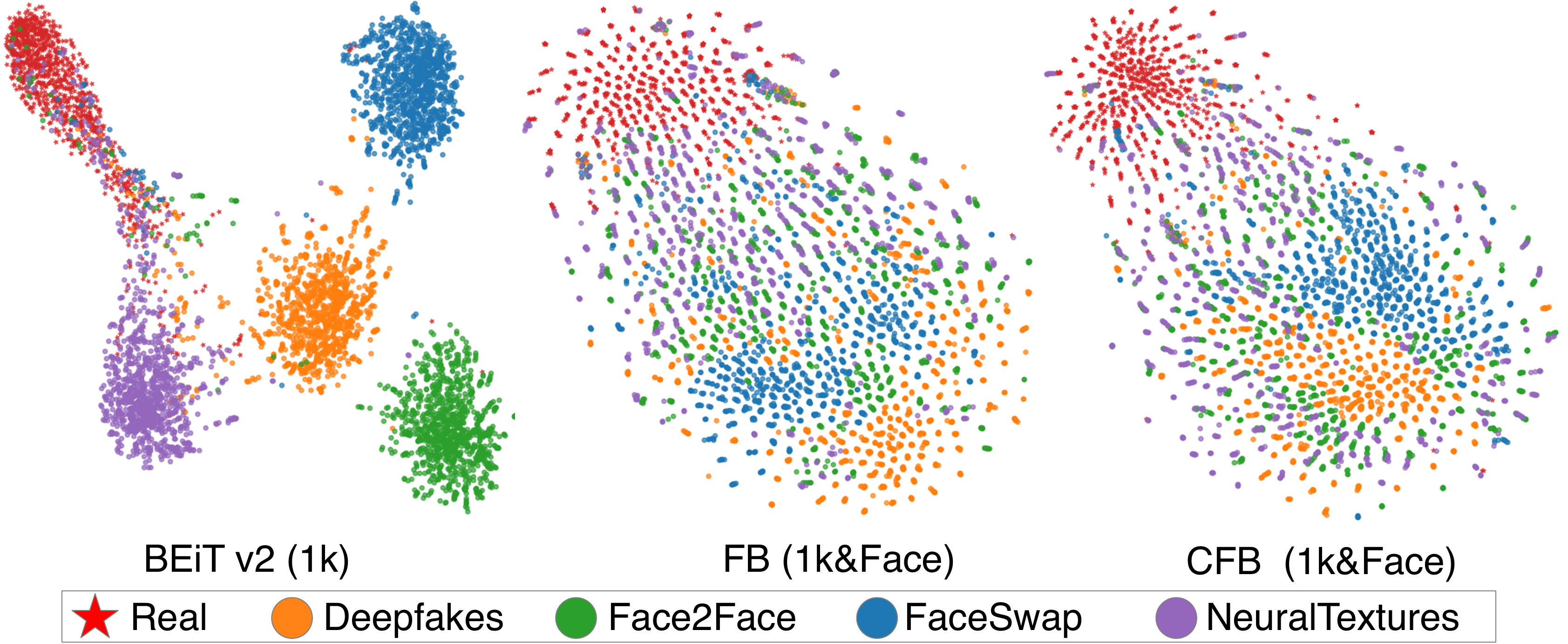}
\vspace{-15pt}
\caption{The t-SNE visualization of last-layer feature vectors from different backbones. Better generalization of the FFD model is indicated by more clustered real points and more dispersed other points.}
\label{fig:tsne}
\end{figure}

To explain the performance improvement, we visualize attention maps to highlight the model's focus areas on the face and conduct cluster analysis in the feature space.
For attention map visualization, we utilize Attention Rollout~\cite{dosovitskiy2020vit}, which averages the attention weights across all heads and recursively multiplies them by the weight matrices of all layers in the backbone. Fig.~\ref{fig:FFD-attnmap} illustrates that BEiT v2 better focuses on local facial components compared to the supervised learning approach. After real-face pre-training, FB effectively targets local forgery areas, while our model is even more precise, leading to more accurate and reliable detection results. For cluster analysis, we use t-SNE~\cite{van2008visualizing} to present the feature vectors of the last layer of the backbone. The isolated features of real and four forgery algorithms (Deepfakes, Face2Face, FaceSwap, NeuralTextures) in BEiT v2, as shown in Fig.~\ref{fig:tsne}, indicate that BEiT v2's limited ability to recognize general forgery cues constrains its generalization performance. In contrast, FB primarily distinguishes real and fake faces thanks to its robust facial representation capabilities from real face pre-training. Meanwhile, our model better clusters real faces and separates them from fake faces, indicating that it more effectively extracts implicit forgery cues for generalization.

\subsection{Analysis of Facial Representation in FFD}
\label{sec:facial-representation}

Given the importance of specific knowledge from real faces for backbones in representing facial features, we introduce a larger-scale real-face dataset to validate its performance. We build a $7500k$ real-face dataset based on MSCelebV1~\cite{guo2016ms}, MegaFace~\cite{miller2015megaface}, and LAION-Face~\cite{zheng2022general} to further pre-train FB, \ie, FB++. Table~\ref{tab:ablation3} shows that adding more real faces can further improve FFD performance, with our framework demonstrating the best results, despite the difficulty of further improvement at higher performance levels.

\begin{figure}[t]
\centering
\includegraphics[width=1.0\linewidth]{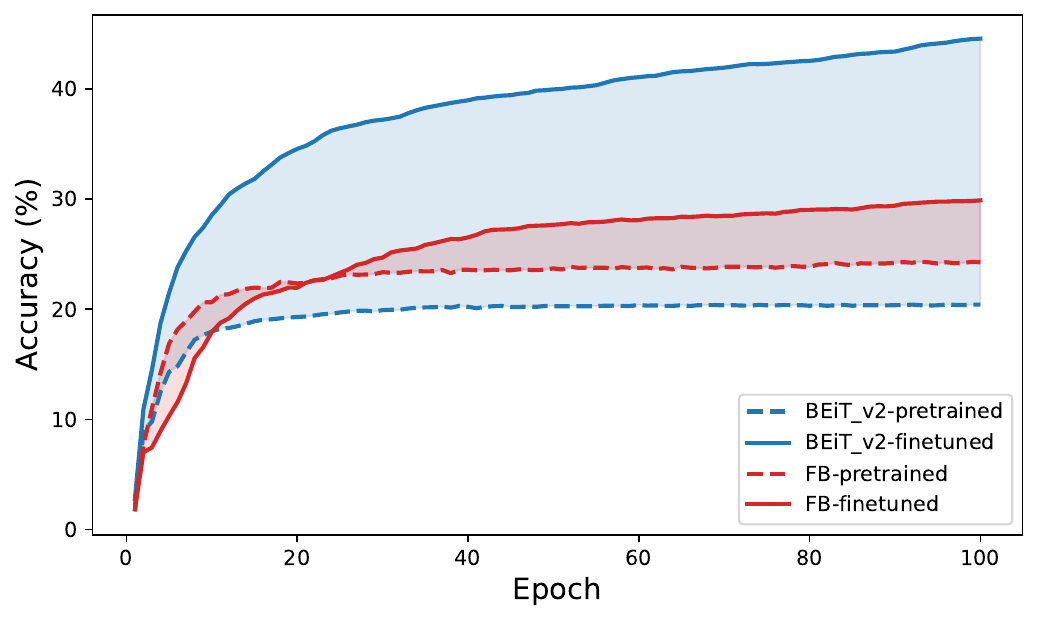}
\vspace{-20pt}
\caption{ID linear classification on frozen features of different backbone.}
\label{fig:Face_re}
\end{figure}

Previous studies~\cite{yan2023ucf, dong2023implicit} propose that FFD models tend to focus on forgery-irrelevant features during training, such as facial identity and gender information, which hinders the FFD model's generalization. Following CADDM~\cite{dong2023implicit}, we freeze the model parameters and use its features to train a linear classifier for face recognition to analyze the retained facial identity features. The experimental results in Fig.~\ref{fig:Face_re} show that: (1) \textit{FB-pretrained} (red dashed line) outperforms \textit{BEiT v2-pretrained} (blue dashed line), demonstrating that pre-training on real faces improves facial representation and enhances performance in both face forgery detection and recognition tasks; (2) After backbone fine-tuning on the FFD dataset, \textit{BEiT v2-finetuned} (blue solid line) increases its focus on facial identity features, while \textit{FB-finetuned} (red solid line) significantly reduces this focus and improves FFD generalization, which suggests that the real-face pre-trained backbone captures comprehensive facial features rather than overfitting to facial identity information.

\begin{table}[t]
    \footnotesize
    \setlength{\tabcolsep}{1.5mm}
    \caption{Quantitative comparison of pre-training on different scales of real faces.}
	\centering
	\midsepremove
		\begin{tabular}{l |c| c| c c c c c c }
		\toprule
\multicolumn{1}{l|}{\multirow{2}{*}{Method}}&\multicolumn{1}{l|}{\multirow{2}{*}{Real Faces}} &  In-set  &\multicolumn{4}{c}{\multirow{1}{*}{Cross-set AUC $\uparrow$(\%)}}  \\ 
\cline{3-7}
              &     & FF++  &  CDF  & DFDC &  FFIW & Avg.   \\
\hline
FB & $550k$ &$99.14$ & $89.18$ & $84.02$ & $86.48$ & $86.56$\\
FB++ &$7500k$ &$99.21$ & $90.76$ & $86.77$ & $88.82$ & $88.78$\\
\hline
CFB &$550k$ &$\textbf{99.36}$ & $90.46$ & $84.90$ & $90.97$ & $88.78$\\
CFB++ &$7500k$ &$99.13$ & $90.99$ & $86.79$ & $90.72$ & $89.50$\\
\hline
CFB $w/$ SBI&$550k$ &$89.70$ & $95.35$ & $85.45$ & $91.82$ & $90.87$\\
CFB++ $w/$ SBI &$7500k$ &$91.96$ & $\textbf{95.46}$ & $\textbf{88.66}$ & $\textbf{92.31}$ & $\textbf{92.14}$\\
		\bottomrule
		\end{tabular}
	\label{tab:ablation3}
\end{table}

\begin{table*}[t]
    \footnotesize
    \setlength{\tabcolsep}{1.1mm}
    
    \caption{Quantitative ACC$\uparrow$ (\%) comparison of different classification thresholds in FFD models with $\mathcal{L}_{EDU}$ or $\mathcal{L}_{ce}$. $\tau_{0.5}$ is the empirical threshold. $\tau_{ot}$ and $\tau_{ot}^{UTOM}$ denote optimal thresholds calculated from the model's default prediction probability and its adjustment by the uncertainty value, respectively. \textbf{Bold} and \underline{underline} indicate the best and second-best results.}

	\centering
	\midsepremove
		\begin{tabular}{c| c| l| c c c c| c c c c |c c c c | c }
		\toprule
  \rule{0pt}{7.5pt} 
        \multirow{2}{*}{Method} & \multirow{2}{*}{Loss} & \multirow{2}{*}{Threshold}& \multicolumn{4}{c|}{\multirow{1}{*}{ CDF to CDF\&DFDC\&FFIW }}   &\multicolumn{4}{c|}{\multirow{1}{*}{DFDC to CDF\&DFDC\&FFIW}}&\multicolumn{4}{c|}{\multirow{1}{*}{FFIW to CDF\&DFDC\&FFIW}}  &\multicolumn{1}{c}{\multirow{2}{*}{Avg.}} \\
                    \cline{4-15}
         \rule{0pt}{7.5pt}                
        & & &CDF&DFDC&FFIW&Avg.&CDF& DFDC&FFIW&Avg.&CDF&DFDC&FFIW&Avg.  \\
	    \hline
     
\multirow{7}{*}{FB} 
&\multirow{3}{*}{$\mathcal{L}_{ce}$} & $\tau_{0.5}$ & $77.61$ & $72.00$ & $76.00$ & $75.20$ & $77.61$ & $72.00$ & $76.00$ & $75.20$ & $77.61$ & $72.00$ & $76.00$ & $75.20$ & $75.20$\\

& & $\tau_{ot}$ & $83.40$ & $75.40$ & $78.00$ & $78.93$ & $80.69$ & $76.02$ & $76.60$ & $77.77$ & $81.47$ & $75.00$ & $80.20$ & $78.89$ & $78.53$\\

& & $\tau_{ot}^{UTOM}$ & $83.01$ & $75.70$ & $79.60$ & $79.44$ & $81.85$ & $75.80$ & $80.20$ & $79.28$ & $82.05$ & $75.76$ & $80.40$ & $79.40$ & $79.37$\\

\cline{2-16}
&\multirow{3}{*}{ $\mathcal{L}_{EDU}$}  & $\tau_{0.5}$ & $80.12$ & $73.42$ & $78.40$ & $77.31$ & $80.12$ & $73.42$ & $78.40$ & $77.31$ & $80.12$ & $73.42$ & $78.40$ & $77.31$ & $77.31$\\

& & $\tau_{ot}$ & $81.85$ & $76.12$ & $77.60$ & $78.52$ & $81.27$ & $76.32$ & $78.60$ & $78.73$ & $80.31$ & $75.70$ & $82.00$ & $79.34$ & $78.86$\\

& & $\tau_{ot}^{UTOM}$ & $82.05$ & $\underline{76.94}$ & $\underline{82.00}$ & $80.33$ & $80.89$ & $\underline{77.56}$ & {$\bm{84.00}$\ } & $81.22$ & $80.89$ & $\underline{77.52}$ & {$\bm{85.60}$\ } & $\underline{81.34}$ & $\underline{80.96}$\\
\hline

\multirow{7}{*}{CFB} 
&\multirow{3}{*}{$\mathcal{L}_{ce}$}  & $\tau_{0.5}$ & $83.40$ & $75.70$ & $81.40$ & $80.17$ & $83.40$ & $75.70$ & $81.40$ & $80.17$ & $\bm{83.40}$ & $75.70$ & $81.40$ & $80.17$ & $80.17$\\

& & $\tau_{ot}$ & $83.98$ & $75.84$ & $81.20$ & $80.34$ & $83.01$ & $76.68$ & $81.20$ & $80.30$ & $82.24$ & $76.16$ & $82.60$ & $80.33$ & $80.32$\\

& & $\tau_{ot}^{UTOM}$ & $\underline{84.36}$ & $76.54$ & $80.00$ & $80.30$ & $\underline{83.59}$ & $77.50$ & $\underline{82.60}$ & $\underline{81.23}$ & $82.82$ & $77.38$ & $82.80$ & $81.00$ & $80.84$\\

\cline{2-16}
&\multirow{3}{*}{ $\mathcal{L}_{EDU}$}  & $0.5$ & $83.20$ & $76.00$ & $81.20$ & $80.13$ & $83.20$ & $76.00$ & $81.20$ & $80.13$ & $\underline{83.20}$ & $76.00$ & $81.20$ & $80.13$ & $80.13$\\

& & $\tau_{ot}$ & $83.78$ & $75.94$ & $81.40$ & $\underline{80.37}$ & $81.66$ & $77.06$ & $82.20$ & $80.31$ & $82.63$ & $76.76$ & $84.20$ & $81.20$ & $80.63$\\

& & $\tau_{ot}^{UTOM}$ & {$\bm{85.33}$\ } & {$\bm{77.54}$\ } & {$\bm{83.80}$\ } & {$\bm{82.22}$\ } & {$\bm{84.75}$\ } & {$\bm{78.66}$\ } & $ \bm{84.00}  $ & {$\bm{82.47}$\ } & {$\bm{83.40}$\ } & {$\bm{78.30}$\ } & $ \underline{84.80} $ & {$\bm{82.17}$\ } & {$\bm{82.29}$\ }\\
		\bottomrule
		\end{tabular}
	\label{tab:inference}
\end{table*}


\subsection{Analysis of Threshold Optimization Mechanism}
\label{sec:inference-experiment}
We validate the critical role of classification threshold $\tau$ in FFD final discriminant results. Our study includes empirical threshold $\tau_{0.5}$, optimal threshold $\tau_{ot}^{\text{UTOM}}$ computed on one cross-dataset and applied to other cross-datasets after adjusting prediction probabilities with our Uncertainty-based Threshold Optimization Mechanism (UTOM), and $\tau_{ot}$ obtained without probability adjustment as a comparison.




Table~\ref{tab:inference} reports the quantitative comparisons, showing that $\tau_{ot}$ from one cross-dataset achieves better ACC on other cross-datasets compared to $\tau_{0.5}$, and $\tau_{ot}^{UTOM}$ performs the best. $\mathcal{L}_{EDU}$ also achieves better ACC compared to $\mathcal{L}_{ce}$, indicating that constraining uncertainty and predicted probability improves the model's confidence in its predictions. Remarkably, $\tau_{ot}^{UTOM}$ with $\mathcal{L}_{EDU}$ achieves a significant ACC improvement over $\tau_{ot}$ with $\mathcal{L}_{ce}$ in both FB and CFB, with an average ACC increase of more than $2$\%, fully demonstrates the reliability of UTOM in FFD applications, as well as the importance of our fine-tuning framework with the uncertainty-based fusion module and $\mathcal{L}_{EDU}$.

Moreover, we gradually increase the number of visible cross-datasets (Source Dataset) to calculate $\tau_{ot}$ and $\tau_{ot}^{UTOM}$, and validate the applicability of our threshold optimization mechanism on fixed unseen cross-datasets (Target Dataset). Results in Table~\ref{tab:ablation-acc} show that the optimal threshold calculated across multiple datasets significantly improves FFD model accuracy, highlighting that leveraging more datasets with diverse forgeries helps compute better classification thresholds and enhances FFD model's applicability. Besides, while UTOM proves effective overall, difficult samples with large $u$ may cause fake-to-real misclassifications. Severe class imbalance can compromise its robustness, requiring minority class re-sampling for cross-domain stability, which is further analyzed in \textit{Supp. File}.

\begin{table}[h]
    \footnotesize
    \setlength{\tabcolsep}{0.15mm}
    \caption{Quantitative ACC$\uparrow$ (\%) comparison of performance on unseen cross-datasets (Target Dataset) by calculating the optimal threshold through the addition of cross-datasets (Source Dataset).}
	\centering
	\midsepremove
    \renewcommand{\arraystretch}{1.1}
		\begin{tabular}{c|  c c c| c c c | c }
		\toprule
           \multirow{1}{*}{Target Dataset} &  \multicolumn{3}{c|}{\multirow{1}{*}{Source Dataset}} 
           &\multicolumn{2}{c}{\multirow{1}{*}{ Threshold}}
           \\  
           \hline
            & CDF & WDF~\cite{zi2020wilddeepfake} & DFD & $\tau_{ot}$ &  $\tau_{ot}^{UTOM}$ \\
           \cline{2-6}         
        \multirow{1}{*}{\shortstack{DFDC+FFIW+DFDCP \\ +AIFace~\cite{Lin_2025_CVPR}\\+UADFV~\cite{UADFV}}}
        
        & \checkmark & & & $76.52$ & $77.00$ \\
        & \checkmark & \checkmark & & $76.52$ & $77.68$ \\
        & \checkmark & \checkmark & \checkmark & $\underline{78.28}$ & $\textbf{79.08}$ \\
        \hline
        & SadTalker & SimSwap & Inswap & $\tau_{ot}$ & $\tau_{ot}^{UTOM}$  \\
        \cline{2-6} 
        \multirow{1}{*}{\shortstack{FOMM+FaceDancer \\ +MobileSwap+UniFace \\ +LIA~\cite{LIA}+FSGAN~\cite{FSGAN}}} 
        & \checkmark &  &  &$89.94$ &$92.08$ \\
        & \checkmark & \checkmark & &$90.00$ &$\underline{92.74}$ \\
        & \checkmark & \checkmark & \checkmark & $92.56$ & $\textbf{92.86}$\\
        
	    \bottomrule

		\end{tabular}
	\label{tab:ablation-acc}
\end{table}

\begin{figure}[t]
\centering
\includegraphics[width=1.0\linewidth]{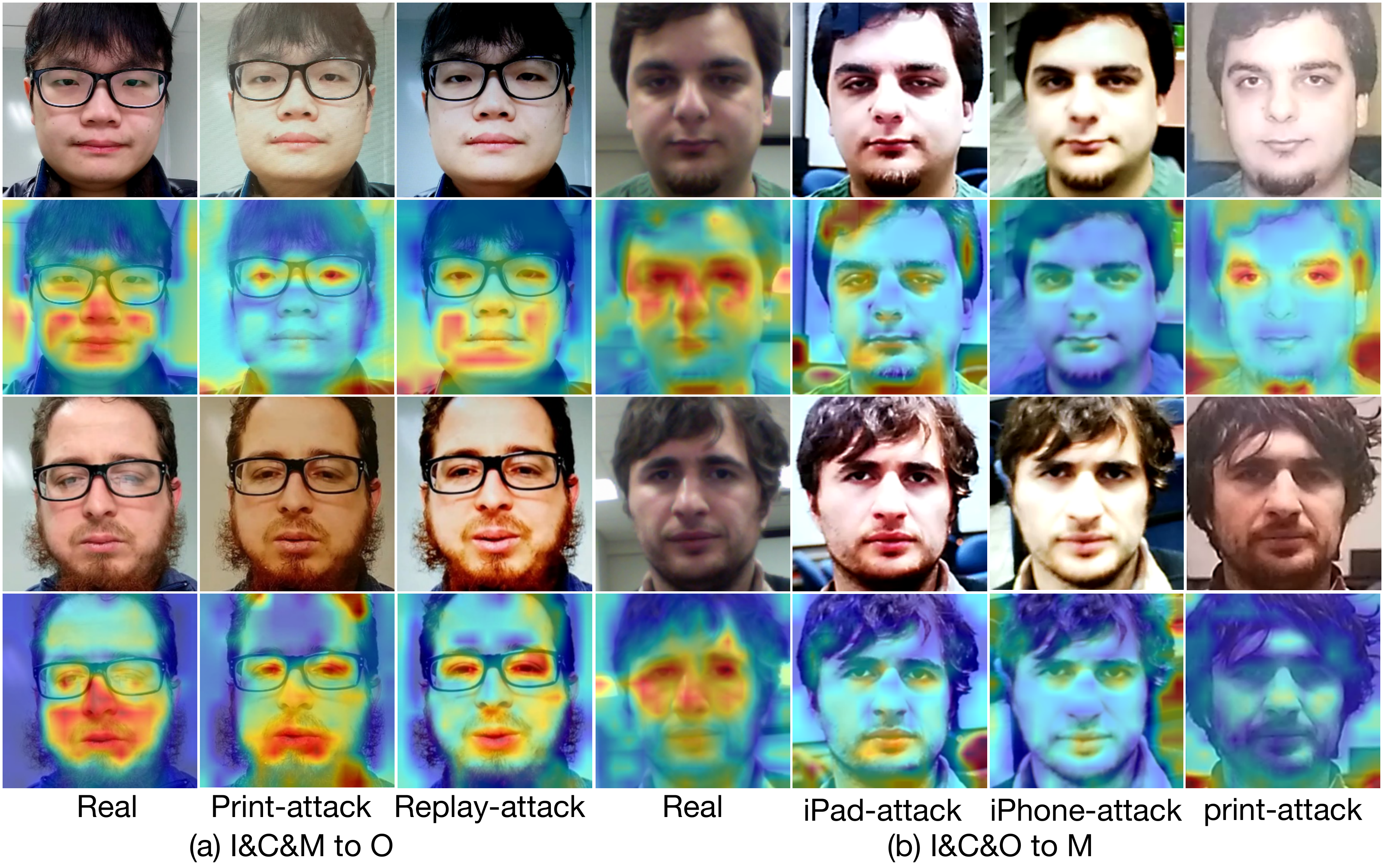}
\vspace{-15pt}
\caption{Visualization of attention maps from our PAD model on different target datasets. Highlighted areas indicate higher model attention.}
\label{fig:Attentionmap-FAS}
\end{figure}

\begin{table*}[ht]
    \footnotesize
    \setlength{\tabcolsep}{0.45mm}
    \caption{Quantitative comparison of different PAD methods from three source datasets to one target dataset.}
    \renewcommand{\arraystretch}{1.1}
	\centering
	\midsepremove
		\begin{tabular}{c| c|  c c | c c | c c | c c | c c}
		\toprule
  \rule{0pt}{7.5pt} 
  \multirow{2}{*}{Method} & \multirow{2}{*}{Venue} &\multicolumn{2}{c| }{\multirow{1}{*}{O\&C\&I\ to\ M}} &\multicolumn{2}{c| }{\multirow{1}{*}{O\&M\&I\ to\ C}} &\multicolumn{2}{c| }{\multirow{1}{*}{O\&C\&M\ to\ I}} &\multicolumn{2}{c| }{\multirow{1}{*}{I\&C\&M\ to\ O}} &\multicolumn{2}{c}{\multirow{1}{*}{Avg.}}\\
  \cline{3-12}
  \rule{0pt}{7.5pt} 
		&  & HTER$\downarrow$ (\%) & AUC$\uparrow$ (\%) &  HTER$\downarrow$ (\%) & AUC$\uparrow$ (\%) &  HTER$\downarrow$ (\%) & AUC$\uparrow$ (\%) &  HTER$\downarrow$ (\%) & AUC$\uparrow$ (\%) &  HTER$\downarrow$ (\%)  & AUC$\uparrow$ (\%) \\
		\midrule
SSDG~\cite{jia2020single} & CVPR 2020 & $7.38$ &$97.17$ & $10.44$ & $95.94$ & $11.71$ & $96.59$ & $15.61$ & $91.54$ & $11.29$ & $95.31$\\
PatchNet~\cite{wang2022patchnet} & CVPR 2022 & $7.10$ &$98.46$ & $11.33$ & $94.58$ & $13.40$ & $95.67$ & $11.82$ & $95.07$ & $10.91$ & $95.95$\\
SSAN~\cite{wang2022domain} & CVPR 2022 & $6.67$ &$98.75$ & $10.00$ & $96.67$ & $8.88$ & $96.79$ & $13.72$ & $93.63$ & $9.82$ & $96.46$\\
DSCI~\cite{long2023dual} & TIFS 2023 & $5.48$ & $97.39$ & $8.00$ & $97.50$ & $5.71$ & $98.44$ & $12.59$ & $94.57$ & $7.95$ & $96.98$\\
UDG~\cite{liu2023towards} & ICCV 2023 & $5.95$ & $98.47$ & $9.82$ & $96.76$ & $5.86$ & $98.62$ & $10.97$ & $95.36$ & $8.15$ & $97.30$\\
GAC-FAS~\cite{le2024grad} & CVPR 2024 & $5.00$ & $97.56$ & $8.20$ & $95.16$ & $\underline{4.29}$ & $\underline{98.87}$ & $8.60$ & $97.16$ & $6.52$ & $97.19$\\
GFPL-FAS~\cite{Liu_2024_CVPR} & CVPR 2024 & $\underline{3.09}$ & $\underline{99.45}$ & $\textbf{2.56}$ & $\underline{99.10}$ & $5.43$ & $98.41$ & $\underline{3.33}$ & $\underline{99.05}$ & $ \underline{3.60} $ & $ \underline{99.00} $\\
	    \hline
CFB-PAD & $-$ & $\textbf{2.62}$ & $\textbf{99.74}$ & $\underline{3.44}$ & $\textbf{99.63}$ & $\textbf{2.93}$ & $\textbf{99.58}$ & $\textbf{3.30}$ & $\textbf{99.53}$ & $\textbf{3.07}$ & $\textbf{99.62}$\\

		\bottomrule
		\end{tabular}
	\label{tab:fas-sota}
\end{table*}

\section{Beyond The Face Forgery Detection}

We extend our exploration of the backbone's \textit{pre-training} and \textit{fine-tuning} stages to the face-related task of presentation attack detection (PAD), which also aims to distinguish the authenticity of input faces. We conduct experiments on four commonly used public datasets: OULU-NPU~\cite{boulkenafet2017oulu} (O), CASIA-FASD~\cite{zhang2012face} (C), Idiap Replay-Attack~\cite{chingovska2012effectiveness} (I), and MSU-MFSD~\cite{wen2015face} (M). We follow the evaluation protocol and data preprocessing process of SSDG~\cite{jia2020single}. One dataset is used as the target domain for evaluation, while the other three serve as source domains for fine-tuning, \ie, O\&C\&I to M, O\&M\&I to C, O\&C\&M to I, and I\&C\&M to O.

We evaluate our PAD model against the state-of-the-art methods: SSDG~\cite{jia2020single}, PatchNet~\cite{wang2022patchnet}, SSAN~\cite{wang2022domain}, DSCI~\cite{long2023dual}, UDG~\cite{liu2023towards}, and GAC-FAS~\cite{le2024grad}. Table~\ref{tab:fas-sota} presents the comparison results, showing that our model performs excellently on HTER and AUC metrics. Especially, our model outperforms GFPL-FAS~\cite{Liu_2024_CVPR} on the challenging OULU-NPU dataset (I\&C\&M to O) with an improvement of $0.48$\% in AUC$\uparrow$ ($99.53$\% vs. $99.05$\%). Additionally, the AUC$\uparrow$ of our model on the CASIA-FASD (O\&M\&I to C) and Idiap Replay-Attack (O\&C\&M to I) datasets exceeds $99$\%. Moreover, we also visualize our model's attention map on PAD in Fig.~\ref{fig:Attentionmap-FAS}, indicating that our model can represent real faces by focusing on the facial region, while fake face images pay attention to inconsistent clues such as light, texture, and background.

\section{CONCLUSION}
In this paper, we present insightful guidance and practical solutions for FFD research through a complete revisiting of FFD workflow and an empirical study of backbones with various configurations. We revitalize the traditional FFD pipeline by introducing SSL on real faces, a competitive backbone framework, and a threshold optimization mechanism in the \textit{pre-training}, \textit{fine-tuning}, and \textit{inference} stages, respectively, thereby developing an elaborate backbone tailored for FFD generalization. Moreover, we leverage uncertainty estimation and evidential deep learning to design an uncertainty-based fusion module for fine-tuning and an adjusted prediction probability strategy to handle magnified confidence issues, enhancing the backbone's generalization and reliability. We conduct comprehensive experiments to demonstrate the practicality of our solution and its method. Additionally, we extend our method to PAD task, further illustrating the superiority of our method. We hope our work opens up new avenues for FFD and PAD tasks.

\ifCLASSOPTIONcompsoc
  \section*{Acknowledgments}
\else
  \section*{Acknowledgment}
\fi

This work was supported by the National Natural Science Foundation of China (No. 62276249 and No. 62571503), the Strategic Priority Research Program of the Chinese Academy of Sciences (No. XDB0680000), the TaiShan Scholars Youth Expert Program of Shandong Province (No. tsqn202507108 and No. tsqn202306096), the Strategic Priority Research Program of CAS (No. XDB0680202), the Beijing Nova Program (No. 20230484368), and the Youth Innovation Promotion Association CAS.

\ifCLASSOPTIONcaptionsoff
  \newpage
\fi

\bibliographystyle{IEEEtran}

\bibliography{deepfake.bib}

@String(PAMI  = {IEEE Trans. Pattern Anal. Mach. Intell.})

@String(IJCV  = {Int. J. Comput. Vis.})

@String(CVPR_IEEE_CVF  = {IEEE/CVF Conf. Comput. Vis. Pattern Recog.})

@String(CVPR_IEEE  = {IEEE Conf. Comput. Vis. Pattern Recog.})

@String(CVPR  = {Conf. Comput. Vis. Pattern Recog.})

@String(ICCV_IEEE_CVF  = {IEEE/CVF Int. Conf. Comput. Vis.})

@String(ICCV_IEEE  = {IEEE Int. Conf. Comput. Vis.})

@String(ICCV  = {Int. Conf. Comput. Vis.})

@String(ECCV  = {Eur. Conf. Comput. Vis.})

@String(NIPS  = {Adv. Neural Inf. Process. Syst.})

@String(ICML  = {Int. Conf. Mach. Learn.})

@String(WACV  = {IEEE Winter  Conf. Appl. Comput. Vis.})

@String(TOG   = {ACM Trans. Graph.})

@String(TIP   = {IEEE Trans. Image Process.})

@String(TIFS   = {IEEE Trans.Inf. Forensics Security.})

@String(TMLR  = {Mach. learn. res.})

@String(ACMMM = {ACM MM})

@String(ICASSP=	{ICASSP})

@String(ICIP  = {ICIP})

@String(ICLR  = {Int. Conf. Learn. Represent.})

@String(PR = {PR})

@String(AAAI = {AAAI})

@article{george2019biometric,
  title={Biometric face presentation attack detection with multi-channel convolutional neural network},
  author={George, Anjith and Mostaani, Zohreh and Geissenbuhler, David and Nikisins, Olegs and Anjos, Andr{\'e} and Marcel, S{\'e}bastien},
  journal=TIFS,
  volume={15},
  pages={42--55},
  year={2019},
  publisher={IEEE}
}

@article{chan2017face,
  title={Face liveness detection using a flash against 2D spoofing attack},
  author={Chan, Patrick PK and Liu, Weiwen and Chen, Danni and Yeung, Daniel S and Zhang, Fei and Wang, Xizhao and Hsu, Chien-Chang},
  journal=TIFS,
  volume={13},
  number={2},
  pages={521--534},
  year={2017},
  publisher={IEEE}
}

@article{wang2021deep,
  title={Deep face recognition: A survey},
  author={Wang, Mei and Deng, Weihong},
  journal={Neurocomputing},
  volume={429},
  pages={215--244},
  year={2021},
  publisher={Elsevier}
}

@article{melnik2024face,
  title={Face generation and editing with stylegan: A survey},
  author={Melnik, Andrew and Miasayedzenkau, Maksim and Makaravets, Dzianis and Pirshtuk, Dzianis and Akbulut, Eren and Holzmann, Dennis and Renusch, Tarek and Reichert, Gustav and Ritter, Helge},
  journal=PAMI,
  year={2024},
  publisher={IEEE}
}

@article{rivera2012local,
  title={Local directional number pattern for face analysis: Face and expression recognition},
  author={Rivera, Adin Ramirez and Castillo, Jorge Rojas and Chae, Oksam Oksam},
  journal=TIP,
  volume={22},
  number={5},
  pages={1740--1752},
  year={2012},
  publisher={IEEE}
}

@inproceedings{karras2019style,
  title={A style-based generator architecture for generative adversarial networks},
  author={Karras, Tero and Laine, Samuli and Aila, Timo},
  booktitle=CVPR_IEEE_CVF,
  pages={4401--4410},
  year={2019}
}

@inproceedings{dang2020detection,
  title={On the detection of digital face manipulation},
  author={Dang, Hao and Liu, Feng and Stehouwer, Joel and Liu, Xiaoming and Jain, Anil K},
  booktitle=CVPR_IEEE_CVF,
  pages={5781--5790},
  year={2020}
}

@inproceedings{wang2021representative,
  title={Representative forgery mining for fake face detection},
  author={Wang, Chengrui and Deng, Weihong},
  booktitle=CVPR_IEEE_CVF,
  pages={14923--14932},
  year={2021}
}

@inproceedings{fei2022learning,
  title={Learning second order local anomaly for general face forgery detection},
  author={Fei, Jianwei and Dai, Yunshu and Yu, Peipeng and Shen, Tianrun and Xia, Zhihua and Weng, Jian},
  booktitle=CVPR_IEEE_CVF,
  pages={20270--20280},
  year={2022}
}

@inproceedings{zhuang2022uia,
  title={UIA-ViT: Unsupervised inconsistency-aware method based on vision transformer for face forgery detection},
  author={Zhuang, Wanyi and Chu, Qi and Tan, Zhentao and Liu, Qiankun and Yuan, Haojie and Miao, Changtao and Luo, Zixiang and Yu, Nenghai},
  booktitle=ECCV,
  pages={391--407},
  year={2022},
  organization={Springer}
}

@article{cozzolino2018forensictransfer,
  title={Forensictransfer: Weakly-supervised domain adaptation for forgery detection},
  author={Cozzolino, Davide and Thies, Justus and R{\"o}ssler, Andreas and Riess, Christian and Nie{\ss}ner, Matthias and Verdoliva, Luisa},
  journal={arXiv preprint arXiv:1812.02510},
  year={2018}
}

@inproceedings{hu2021improving,
  title={Improving the generalization ability of deepfake detection via disentangled representation learning},
  author={Hu, Jiashang and Wang, Shilin and Li, Xiaoyong},
  booktitle=ICIP,
  pages={3577--3581},
  year={2021},
}

@article{yan2022deepfake,
  title={Deepfake Detection via Joint Unsupervised Reconstruction and Supervised Classification},
  author={Yan, Bosheng and Li, Chang-Tsun and Lu, Xuequan},
  journal={arXiv preprint arXiv:2211.13424},
  year={2022}
}

@inproceedings{cao2022end,
  title={End-to-end reconstruction-classification learning for face forgery detection},
  author={Cao, Junyi and Ma, Chao and Yao, Taiping and Chen, Shen and Ding, Shouhong and Yang, Xiaokang},
  booktitle=CVPR_IEEE_CVF,
  pages={4113--4122},
  year={2022}
}

@inproceedings{wang2022m2tr,
  title={M2tr: Multi-modal multi-scale transformers for deepfake detection},
  author={Wang, Junke and Wu, Zuxuan and Ouyang, Wenhao and Han, Xintong and Chen, Jingjing and Jiang, Yu-Gang and Li, Ser-Nam},
  booktitle={ICMR},
  pages={615--623},
  year={2022}
}

@article{miao2023f,
  title={F 2 trans: High-frequency fine-grained transformer for face forgery detection},
  author={Miao, Changtao and Tan, Zichang and Chu, Qi and Liu, Huan and Hu, Honggang and Yu, Nenghai},
  journal=TIFS,
  volume={18},
  pages={1039--1051},
  year={2023},
  publisher={IEEE}
}

@inproceedings{he2016deep,
  title={Deep residual learning for image recognition},
  author={He, Kaiming and Zhang, Xiangyu and Ren, Shaoqing and Sun, Jian},
  booktitle=CVPR_IEEE,
  pages={770--778},
  year={2016}
}

@inproceedings{chollet2017xception,
  title={Xception: Deep learning with depthwise separable convolutions},
  author={Chollet, Fran{\c{c}}ois},
  booktitle=CVPR_IEEE,
  pages={1251--1258},
  year={2017}
}

@inproceedings{tan2019efficientnet,
  title={Efficientnet: Rethinking model scaling for convolutional neural networks},
  author={Tan, Mingxing and Le, Quoc},
  booktitle=ICML,
  pages={6105--6114},
  year={2019},
  organization={PMLR}
}

@article{dosovitskiy2020vit,
  title={An Image is Worth 16x16 Words: Transformers for Image Recognition at Scale},
  author={Dosovitskiy, Alexey and Beyer, Lucas and Kolesnikov, Alexander and Weissenborn, Dirk and Zhai, Xiaohua and Unterthiner, Thomas and  Dehghani, Mostafa and Minderer, Matthias and Heigold, Georg and Gelly, Sylvain and Uszkoreit, Jakob and Houlsby, Neil},
  journal={ICLR},
  year={2021}
}

@article{miao2021learning,
  title={Learning forgery region-aware and ID-independent features for face manipulation detection},
  author={Miao, Changtao and Chu, Qi and Li, Weihai and Li, Suichan and Tan, Zhentao and Zhuang, Wanyi and Yu, Nenghai},
  journal={IEEE Transactions on Biometrics, Behavior, and Identity Science},
  volume={4},
  number={1},
  pages={71--84},
  year={2021},
  publisher={IEEE}
}

@inproceedings{liu2015faceattributes,
  title = {Deep Learning Face Attributes in the Wild},
  author = {Liu, Ziwei and Luo, Ping and Wang, Xiaogang and Tang, Xiaoou},
  booktitle = ICCV_IEEE,
  year = {2015} 
}

@inproceedings{yu2022celebvtext,
  title={{CelebV-Text}: A Large-Scale Facial Text-Video Dataset},
  author={Yu, Jianhui and Zhu, Hao and Jiang, Liming and Loy, Chen Change and Cai, Weidong and Wu, Wayne},
  booktitle=CVPR_IEEE_CVF,
  year={2023}
}

@inproceedings{yao2023towards,
  title={Towards Understanding the Generalization of Deepfake Detectors from a Game-Theoretical View},
  author={Yao, Kelu and Wang, Jin and Diao, Boyu and Li, Chao},
  booktitle=ICCV_IEEE_CVF,
  pages={2031--2041},
  year={2023}
}

@inproceedings{shiohara2022detecting,
  title={Detecting deepfakes with self-blended images},
  author={Shiohara, Kaede and Yamasaki, Toshihiko},
  booktitle=CVPR_IEEE_CVF,
  pages={18720--18729},
  year={2022}
}

@inproceedings{li2020face,
  title={Face x-ray for more general face forgery detection},
  author={Li, Lingzhi and Bao, Jianmin and Zhang, Ting and Yang, Hao and Chen, Dong and Wen, Fang and Guo, Baining},
  booktitle=CVPR_IEEE_CVF,
  pages={5001--5010},
  year={2020}
}

@inproceedings{le2024grad,
  title={Gradient Alignment for Cross-Domain Face Anti-Spoofing},
  author={Le, Binh M and Woo, Simon S},
  booktitle=CVPR_IEEE_CVF,
  year={2024}
}

@inproceedings{zhao2021multi,
  title={Multi-attentional deepfake detection},
  author={Zhao, Hanqing and Zhou, Wenbo and Chen, Dongdong and Wei, Tianyi and Zhang, Weiming and Yu, Nenghai},
  booktitle=CVPR_IEEE_CVF,
  pages={2185--2194},
  year={2021}
}

@inproceedings{sun2022information,
  title={An information theoretic approach for attention-driven face forgery detection},
  author={Sun, Ke and Liu, Hong and Yao, Taiping and Sun, Xiaoshuai and Chen, Shen and Ding, Shouhong and Ji, Rongrong},
  booktitle=ECCV,
  pages={111--127},
  year={2022},
  organization={Springer}
}

@inproceedings{gidaris2021obow,
  title={Obow: Online bag-of-visual-words generation for self-supervised learning},
  author={Gidaris, Spyros and Bursuc, Andrei and Puy, Gilles and Komodakis, Nikos and Cord, Matthieu and P{\'e}rez, Patrick},
  booktitle=CVPR_IEEE_CVF,
  pages={6830--6840},
  year={2021}
}

@article{bardes2021vicreg,
  title={Vicreg: Variance-invariance-covariance regularization for self-supervised learning},
  author={Bardes, Adrien and Ponce, Jean and LeCun, Yann},
  journal={arXiv preprint arXiv:2105.04906},
  year={2021}
}

@inproceedings{chen2021exploring,
  title={Exploring simple siamese representation learning},
  author={Chen, Xinlei and He, Kaiming},
  booktitle=CVPR_IEEE_CVF,
  pages={15750--15758},
  year={2021}
}

@inproceedings{chen2020simple,
  title={A simple framework for contrastive learning of visual representations},
  author={Chen, Ting and Kornblith, Simon and Norouzi, Mohammad and Hinton, Geoffrey},
  booktitle=ICML,
  pages={1597--1607},
  year={2020},
  organization={PMLR}
}

@article{chen2020big,
  title={Big self-supervised models are strong semi-supervised learners},
  author={Chen, Ting and Kornblith, Simon and Swersky, Kevin and Norouzi, Mohammad and Hinton, Geoffrey E},
  journal=NIPS,
  volume={33},
  pages={22243--22255},
  year={2020}
}

@inproceedings{he2020momentum,
  title={Momentum contrast for unsupervised visual representation learning},
  author={He, Kaiming and Fan, Haoqi and Wu, Yuxin and Xie, Saining and Girshick, Ross},
  booktitle=CVPR_IEEE_CVF,
  pages={9729--9738},
  year={2020}
}

@article{chen2003improved,
  title={Improved baselines with momentum contrastive learning. arXiv 2020},
  author={Chen, Xinlei and Fan, Haoqi and Girshick, Ross and He, Kaiming},
  journal={arXiv preprint arXiv:2003.04297},
  year={2003}
}

@inproceedings{chen2021empirical,
  title={An empirical study of training self-supervised vision transformers},
  author={Chen, Xinlei and Xie, Saining and He, Kaiming},
  booktitle=ICCV_IEEE_CVF,
  pages={9640--9649},
  year={2021}
}

@inproceedings{he2022masked,
  title={Masked autoencoders are scalable vision learners},
  author={He, Kaiming and Chen, Xinlei and Xie, Saining and Li, Yanghao and Doll{\'a}r, Piotr and Girshick, Ross},
  booktitle=CVPR_IEEE_CVF,
  pages={16000--16009},
  year={2022}
}

@inproceedings{caron2021emerging,
  title={Emerging Properties in Self-Supervised Vision Transformers},
  author={Caron, Mathilde and Touvron, Hugo and Misra, Ishan and J\'egou, Herv\'e  and Mairal, Julien and Bojanowski, Piotr and Joulin, Armand},
  booktitle=ICCV_IEEE_CVF,
  year={2021}
}

@article{peng2208beit,
  title={{BEiT} v2: Masked image modeling with vector-quantized visual tokenizers},
  author={Peng, Z and Dong, L and Bao, H and Ye, Q and Wei, F},
  journal={arXiv preprint arXiv:2208.06366},
  year={2022}
}

@inproceedings{wang2023image,
  title={Image as a foreign language: Beit pretraining for vision and vision-language tasks},
  author={Wang, Wenhui and Bao, Hangbo and Dong, Li and Bjorck, Johan and Peng, Zhiliang and Liu, Qiang and Aggarwal, Kriti and Mohammed, Owais Khan and Singhal, Saksham and Som, Subhojit and others},
  booktitle=CVPR_IEEE_CVF,
  pages={19175--19186},
  year={2023}
}

@inproceedings{xie2022simmim,
  title={Simmim: A simple framework for masked image modeling},
  author={Xie, Zhenda and Zhang, Zheng and Cao, Yue and Lin, Yutong and Bao, Jianmin and Yao, Zhuliang and Dai, Qi and Hu, Han},
  booktitle=CVPR_IEEE_CVF,
  pages={9653--9663},
  year={2022}
}

@article{yi2022masked,
  title={Masked image modeling with denoising contrast},
  author={Yi, Kun and Ge, Yixiao and Li, Xiaotong and Yang, Shusheng and Li, Dian and Wu, Jianping and Shan, Ying and Qie, Xiaohu},
  journal=ICLR,
  year={2023}
}

@article{chen2024context,
  title={Context autoencoder for self-supervised representation learning},
  author={Chen, Xiaokang and Ding, Mingyu and Wang, Xiaodi and Xin, Ying and Mo, Shentong and Wang, Yunhao and Han, Shumin and Luo, Ping and Zeng, Gang and Wang, Jingdong},
  journal=IJCV,
  volume={132},
  number={1},
  pages={208--223},
  year={2024},
  publisher={Springer}
}

@inproceedings{zheng2021exploring,
  title={Exploring Temporal Coherence for More General Video Face Forgery Detection},
  author={Zheng, Yinglin and Bao, Jianmin and Chen, Dong and Zeng, Ming and Wen, Fang},
  booktitle=ICCV_IEEE_CVF,
  pages={15044--15054},
  year={2021}
}

@InProceedings{ni2022core,
    author    = {Ni, Yunsheng and Meng, Depu and Yu, Changqian and Quan, Chengbin and Ren, Dongchun and Zhao, Youjian},
    title     = {CORE: COnsistent REpresentation Learning for Face Forgery Detection},
    booktitle = {Proceedings of the IEEE/CVF Conference on Computer Vision and Pattern Recognition (CVPR) Workshops},
    year      = {2022},
    pages     = {12-21}
}

@article{wang2022deepfake,
  title={Deepfake forensics via an adversarial game},
  author={Wang, Zhi and Guo, Yiwen and Zuo, Wangmeng},
  journal=TIP,
  volume={31},
  pages={3541--3552},
  year={2022},
  publisher={IEEE}
}

@inproceedings{yan2023ucf,
  title={Ucf: Uncovering common features for generalizable deepfake detection},
  author={Yan, Zhiyuan and Zhang, Yong and Fan, Yanbo and Wu, Baoyuan},
  booktitle=ICCV_IEEE_CVF,
  pages={22412--22423},
  year={2023}
}

@inproceedings{dong2023implicit,
  title={Implicit identity leakage: The stumbling block to improving deepfake detection generalization},
  author={Dong, Shichao and Wang, Jin and Ji, Renhe and Liang, Jiajun and Fan, Haoqiang and Ge, Zheng},
  booktitle=CVPR_IEEE_CVF,
  pages={3994--4004},
  year={2023}
}

@inproceedings{wang2023altfreezing,
  title={Altfreezing for more general video face forgery detection},
  author={Wang, Zhendong and Bao, Jianmin and Zhou, Wengang and Wang, Weilun and Li, Houqiang},
  booktitle=CVPR_IEEE_CVF,
  pages={4129--4138},
  year={2023}
}

@inproceedings{jia2020single,
  title={Single-side domain generalization for face anti-spoofing},
  author={Jia, Yunpei and Zhang, Jie and Shan, Shiguang and Chen, Xilin},
  booktitle=CVPR_IEEE_CVF,
  pages={8484--8493},
  year={2020}
}

@inproceedings{wang2022domain,
  title={Domain generalization via shuffled style assembly for face anti-spoofing},
  author={Wang, Zhuo and Wang, Zezheng and Yu, Zitong and Deng, Weihong and Li, Jiahong and Gao, Tingting and Wang, Zhongyuan},
  booktitle=CVPR_IEEE_CVF,
  pages={4123--4133},
  year={2022}
}

@inproceedings{wang2022patchnet,
  title={Patchnet: A simple face anti-spoofing framework via fine-grained patch recognition},
  author={Wang, Chien-Yi and Lu, Yu-Ding and Yang, Shang-Ta and Lai, Shang-Hong},
  booktitle=CVPR_IEEE_CVF,
  pages={20281--20290},
  year={2022}
}

@article{long2023dual,
  title={Dual Sampling Based Causal Intervention for Face Anti-Spoofing with Identity Debiasing},
  author={Long, Xingming and Zhang, Jie and Wu, Shuzhe and Jin, Xin and Shan, Shiguang},
  journal=TIFS,
  year={2023},
  publisher={IEEE}
}

@inproceedings{liu2023towards,
  title={Towards unsupervised domain generalization for face anti-spoofing},
  author={Liu, Yuchen and Chen, Yabo and Gou, Mengran and Huang, Chun-Ting and Wang, Yaoming and Dai, Wenrui and Xiong, Hongkai},
  booktitle=ICCV_IEEE_CVF,
  pages={20654--20664},
  year={2023}
}

@article{kendall2017uncertainties,
  title={What uncertainties do we need in bayesian deep learning for computer vision?},
  author={Kendall, Alex and Gal, Yarin},
  journal=NIPS,
  volume={30},
  year={2017}
}

@inproceedings{bao2021evidential,
  title={Evidential deep learning for open set action recognition},
  author={Bao, Wentao and Yu, Qi and Kong, Yu},
  booktitle=ICCV_IEEE_CVF,
  pages={13349--13358},
  year={2021}
}

@article{he2020deformable,
  title={Deformable face net for pose invariant face recognition},
  author={He, Mingjie and Zhang, Jie and Shan, Shiguang and Kan, Meina and Chen, Xilin},
  journal=PR,
  volume={100},
  pages={107113},
  year={2020},
  publisher={Elsevier}
}

@article{bao2021beit,
  title={Beit: Bert pre-training of image transformers},
  author={Bao, Hangbo and Dong, Li and Piao, Songhao and Wei, Furu},
  journal={arXiv preprint arXiv:2106.08254},
  year={2021}
}

@inproceedings{radford2021learning,
  title={Learning transferable visual models from natural language supervision},
  author={Radford, Alec and Kim, Jong Wook and Hallacy, Chris and Ramesh, Aditya and Goh, Gabriel and Agarwal, Sandhini and Sastry, Girish and Askell, Amanda and Mishkin, Pamela and Clark, Jack and others},
  booktitle=ICML,
  pages={8748--8763},
  year={2021},
  organization={PMLR}
}

@article{sensoy2018evidential,
  title={Evidential deep learning to quantify classification uncertainty},
  author={Sensoy, Murat and Kaplan, Lance and Kandemir, Melih},
  journal=NIPS,
  volume={31},
  year={2018}
}

@inproceedings{rossler2019faceforensics++,
  title={Faceforensics++: Learning to detect manipulated facial images},
  author={Rossler, Andreas and Cozzolino, Davide and Verdoliva, Luisa and Riess, Christian and Thies, Justus and Nie{\ss}ner, Matthias},
  booktitle=ICCV_IEEE_CVF,
  pages={1--11},
  year={2019}
}

@inproceedings{li2020celeb,
  title={Celeb-df: A large-scale challenging dataset for deepfake forensics},
  author={Li, Yuezun and Yang, Xin and Sun, Pu and Qi, Honggang and Lyu, Siwei},
  booktitle=CVPR_IEEE_CVF,
  pages={3207--3216},
  year={2020}
}

@article{dolhansky2020deepfake,
  title={The deepfake detection challenge (dfdc) dataset},
  author={Dolhansky, Brian and Bitton, Joanna and Pflaum, Ben and Lu, Jikuo and Howes, Russ and Wang, Menglin and Ferrer, Cristian Canton},
  journal={arXiv preprint arXiv:2006.07397},
  year={2020}
}

@inproceedings{zhou2021face,
  title={Face forensics in the wild},
  author={Zhou, Tianfei and Wang, Wenguan and Liang, Zhiyuan and Shen, Jianbing},
  booktitle=CVPR_IEEE_CVF,
  pages={5778--5788},
  year={2021}
}

@inproceedings{thies2016face2face,
  title={Face2face: Real-time face capture and reenactment of rgb videos},
  author={Thies, Justus and Zollhofer, Michael and Stamminger, Marc and Theobalt, Christian and Nie{\ss}ner, Matthias},
  booktitle=CVPR_IEEE,
  pages={2387--2395},
  year={2016}
}

@article{thies2019deferred,
  title={Deferred neural rendering: Image synthesis using neural textures},
  author={Thies, Justus and Zollh{\"o}fer, Michael and Nie{\ss}ner, Matthias},
  journal=TOG,
  volume={38},
  number={4},
  pages={1--12},
  year={2019},
  publisher={ACM New York, NY, USA}
}

@article{you2019large,
  title={Large batch optimization for deep learning: Training bert in 76 minutes},
  author={You, Yang and Li, Jing and Reddi, Sashank and Hseu, Jonathan and Kumar, Sanjiv and Bhojanapalli, Srinadh and Song, Xiaodan and Demmel, James and Keutzer, Kurt and Hsieh, Cho-Jui},
  journal={arXiv preprint arXiv:1904.00962},
  year={2019}
}

@article{deng2019retinaface,
  title={Retinaface: Single-stage dense face localisation in the wild},
  author={Deng, Jiankang and Guo, Jia and Zhou, Yuxiang and Yu, Jinke and Kotsia, Irene and Zafeiriou, Stefanos},
  journal={arXiv preprint arXiv:1905.00641},
  year={2019}
}

@inproceedings{boulkenafet2017oulu,
  title={Oulu-npu: A mobile face presentation attack database with real-world variations},
  author={Boulkenafet, Zinelabinde and Komulainen, Jukka and Li, Lei and Feng, Xiaoyi and Hadid, Abdenour},
  booktitle={2017 12th IEEE international conference on automatic face \& gesture recognition (FG 2017)},
  pages={612--618},
  year={2017},
  organization={IEEE}
}

@inproceedings{zhang2012face,
  title={A face antispoofing database with diverse attacks},
  author={Zhang, Zhiwei and Yan, Junjie and Liu, Sifei and Lei, Zhen and Yi, Dong and Li, Stan Z},
  booktitle={2012 5th IAPR international conference on Biometrics (ICB)},
  pages={26--31},
  year={2012},
  organization={IEEE}
}

@inproceedings{chingovska2012effectiveness,
  title={On the effectiveness of local binary patterns in face anti-spoofing},
  author={Chingovska, Ivana and Anjos, Andr{\'e} and Marcel, S{\'e}bastien},
  booktitle={2012 BIOSIG-proceedings of the international conference of biometrics special interest group (BIOSIG)},
  pages={1--7},
  year={2012},
  organization={IEEE}
}

@article{wen2015face,
  title={Face spoof detection with image distortion analysis},
  author={Wen, Di and Han, Hu and Jain, Anil K},
  journal=TIFS,
  volume={10},
  number={4},
  pages={746--761},
  year={2015},
  publisher={IEEE}
}

@article{van2008visualizing,
  title={Visualizing data using t-SNE.},
  author={Van der Maaten, Laurens and Hinton, Geoffrey},
  journal={JMLR},
  volume={9},
  number={11},
  year={2008}
}

@article{wang2022disentangled,
  title={Disentangled representation learning},
  author={Wang, Xin and Chen, Hong and Tang, Si'ao and Wu, Zihao and Zhu, Wenwu},
  journal={arXiv preprint arXiv:2211.11695},
  year={2022}
}

@article{shapley1953value,
  title={A Value for N-Person Games},
  author={Shapley, Lloyd S},
  journal={Annals of Mathematics Studies},
  pages={307--317},
  year={1953},
  publisher={Princeton University Press Princeton, NJ}
}

@inproceedings{dong2022explaining,
  title={Explaining deepfake detection by analysing image matching},
  author={Dong, Shichao and Wang, Jin and Liang, Jiajun and Fan, Haoqiang and Ji, Renhe},
  booktitle=ECCV,
  pages={18--35},
  year={2022},
  organization={Springer}
}

@inproceedings{chen2022self,
  title={Self-supervised learning of adversarial example: Towards good generalizations for deepfake detection},
  author={Chen, Liang and Zhang, Yong and Song, Yibing and Liu, Lingqiao and Wang, Jue},
  booktitle=ICCV_IEEE_CVF,
  pages={18710--18719},
  year={2022}
}

@article{grill2020bootstrap,
  title={Bootstrap your own latent-a new approach to self-supervised learning},
  author={Grill, Jean-Bastien and Strub, Florian and Altch{\'e}, Florent and Tallec, Corentin and Richemond, Pierre and Buchatskaya, Elena and Doersch, Carl and Avila Pires, Bernardo and Guo, Zhaohan and Gheshlaghi Azar, Mohammad and others},
  journal=NIPS,
  volume={33},
  pages={21271--21284},
  year={2020}
}

@article{guo2016ms,
  title={Ms-celeb-1m: Challenge of recognizing one million celebrities in the real world},
  author={Guo, Yandong and Zhang, Lei and Hu, Yuxiao and He, Xiaodong and Gao, Jianfeng},
  journal={Electronic imaging},
  volume={28},
  pages={1--6},
  year={2016},
  publisher={Society for Imaging Science and Technology}
}

@article{miller2015megaface,
  title={Megaface: A million faces for recognition at scale},
  author={Miller, Daniel and Brossard, Evan and Seitz, S and Kemelmacher-Shlizerman, Ira},
  journal={arXiv preprint arXiv:1505.02108},
  year={2015}
}

@inproceedings{zheng2022general,
  title={General facial representation learning in a visual-linguistic manner},
  author={Zheng, Yinglin and Yang, Hao and Zhang, Ting and Bao, Jianmin and Chen, Dongdong and Huang, Yangyu and Yuan, Lu and Chen, Dong and Zeng, Ming and Wen, Fang},
  booktitle=CVPR_IEEE_CVF,
  pages={18697--18709},
  year={2022}
}

@article{radford2018improving,
  title={Improving language understanding by generative pre-training},
  author={Radford, Alec and Narasimhan, Karthik and Salimans, Tim and Sutskever, Ilya and others},
  year={2018},
  journal={OpenAI}
}

@inproceedings{choi2024exploiting,
  title={Exploiting Style Latent Flows for Generalizing Deepfake Video Detection},
  author={Choi, Jongwook and Kim, Taehoon and Jeong, Yonghyun and Baek, Seungryul and Choi, Jongwon},
  booktitle= CVPR_IEEE_CVF,
  pages={1133--1143},
  year={2024}
}

@InProceedings{yan2023transcending,
    author    = {Yan, Zhiyuan and Luo, Yuhao and Lyu, Siwei and Liu, Qingshan and Wu, Baoyuan},
    title     = {Transcending Forgery Specificity with Latent Space Augmentation for Generalizable Deepfake Detection},
    booktitle = CVPR_IEEE_CVF,
    month     = {June},
    year      = {2024},
    pages     = {8984-8994}
}

@article{
ozbulak2023know,
title={Know Your Self-supervised Learning: A Survey on Image-based Generative and Discriminative Training},
author={Utku Ozbulak and Hyun Jung Lee and Beril Boga and Esla Timothy Anzaku and Ho-min Park and Arnout Van Messem and Wesley De Neve and Joris Vankerschaver},
journal=TMLR,
issn={2835-8856},
year={2023},
}

@article{jing2020self,
  title={Self-supervised visual feature learning with deep neural networks: A survey},
  author={Jing, Longlong and Tian, Yingli},
  journal=PAMI,
  volume={43},
  number={11},
  pages={4037--4058},
  year={2020},
  publisher={IEEE}
}

@inproceedings{caron2018deep,
  title={Deep clustering for unsupervised learning of visual features},
  author={Caron, Mathilde and Bojanowski, Piotr and Joulin, Armand and Douze, Matthijs},
  booktitle=ECCV,
  pages={132--149},
  year={2018}
}

@inproceedings{van2020scan,
  title={Scan: Learning to classify images without labels},
  author={Van Gansbeke, Wouter and Vandenhende, Simon and Georgoulis, Stamatios and Proesmans, Marc and Van Gool, Luc},
  booktitle=ECCV,
  pages={268--285},
  year={2020},
  organization={Springer}
}

@article{caron2020unsupervised,
  title={Unsupervised learning of visual features by contrasting cluster assignments},
  author={Caron, Mathilde and Misra, Ishan and Mairal, Julien and Goyal, Priya and Bojanowski, Piotr and Joulin, Armand},
  journal=NIPS,
  volume={33},
  pages={9912--9924},
  year={2020}
}

@article{devlin2018bert,
  title={B{E}{R}{T}: Pre-training of deep bidirectional transformers for language understanding},
  author={Devlin, Jacob and Chang, Ming-Wei and Lee, Kenton and Toutanova, Kristina},
  journal={arXiv preprint arXiv:1810.04805},
  year={2018}
}

@misc{deepfakes2018,
    title = {Deepfakes github},
    url = {https://github.com/ deepfakes/faceswap},
    urldate      = {2018-10-29},
    year={2018}

}

@misc{faceswap2020,
    author={Marek Kowalski},
    title = {Faceswap github},
    url = {https://github.com/MarekKowalski/FaceSwap},
    urldate      = {2020-08-05},
    year={2020}

}

@article{radford2015unsupervised,
  title={Unsupervised representation learning with deep convolutional generative adversarial networks},
  author={Radford, Alec},
  journal={arXiv preprint arXiv:1511.06434},
  year={2015}
}

@inproceedings{Karras2021,
  author = {Tero Karras and Miika Aittala and Samuli Laine and Erik H\"ark\"onen and Janne Hellsten and Jaakko Lehtinen and Timo Aila},
  title = {Alias-Free Generative Adversarial Networks},
  booktitle = NIPS,
  year = {2021}
}

@inproceedings{rombach2022high,
  title={High-resolution image synthesis with latent diffusion models},
  author={Rombach, Robin and Blattmann, Andreas and Lorenz, Dominik and Esser, Patrick and Ommer, Bj{\"o}rn},
  booktitle=CVPR_IEEE_CVF,
  pages={10684--10695},
  year={2022}
}

@inproceedings{choi2020starganv2,
  title={StarGAN v2: Diverse Image Synthesis for Multiple Domains},
  author={Yunjey Choi and Youngjung Uh and Jaejun Yoo and Jung-Woo Ha},
  booktitle=CVPR_IEEE_CVF,
  year={2020}
}

@inproceedings{xu2018attngan,
  title={Attngan: Fine-grained text to image generation with attentional generative adversarial networks},
  author={Xu, Tao and Zhang, Pengchuan and Huang, Qiuyuan and Zhang, Han and Gan, Zhe and Huang, Xiaolei and He, Xiaodong},
  booktitle=CVPR_IEEE_CVF,
  pages={1316--1324},
  year={2018}
}

@inproceedings{liu2023fine,
  title={Fine-grained face swapping via regional gan inversion},
  author={Liu, Zhian and Li, Maomao and Zhang, Yong and Wang, Cairong and Zhang, Qi and Wang, Jue and Nie, Yongwei},
  booktitle=CVPR_IEEE_CVF,
  pages={8578--8587},
  year={2023}
}

@article{kim2022diffface,
  title={Diffface: Diffusion-based face swapping with facial guidance},
  author={Kim, Kihong and Kim, Yunho and Cho, Seokju and Seo, Junyoung and Nam, Jisu and Lee, Kychul and Kim, Seungryong and Lee, KwangHee},
  journal={arXiv preprint arXiv:2212.13344},
  year={2022}
}

@inproceedings{stypulkowski2024diffused,
  title={Diffused heads: Diffusion models beat gans on talking-face generation},
  author={Stypu{\l}kowski, Micha{\l} and Vougioukas, Konstantinos and He, Sen and Zi{\k{e}}ba, Maciej and Petridis, Stavros and Pantic, Maja},
  booktitle={IEEE/CVF Conf. Comput. Vis. Pattern Recognit. Workshops},
  pages={5091--5100},
  year={2024}
}

@inproceedings{prajwal2020lip,
  title={A lip sync expert is all you need for speech to lip generation in the wild},
  author={Prajwal, KR and Mukhopadhyay, Rudrabha and Namboodiri, Vinay P and Jawahar, CV},
  booktitle=ACMMM,
  pages={484--492},
  year={2020}
}

@InProceedings{Nguyen_2024_CVPR,
    author    = {Nguyen, Dat and Mejri, Nesryne and Singh, Inder Pal and Kuleshova, Polina and Astrid, Marcella and Kacem, Anis and Ghorbel, Enjie and Aouada, Djamila},
    title     = {LAA-Net: Localized Artifact Attention Network for Quality-Agnostic and Generalizable Deepfake Detection},
    booktitle = CVPR_IEEE_CVF,
    month     = {June},
    year      = {2024},
    pages     = {17395-17405}
}

@inproceedings{xu2022mobilefaceswap,
  title={Mobilefaceswap: A lightweight framework for video face swapping},
  author={Xu, Zhiliang and Hong, Zhibin and Ding, Changxing and Zhu, Zhen and Han, Junyu and Liu, Jingtuo and Ding, Errui},
  booktitle=AAAI,
  volume={36},
  number={3},
  pages={2973--2981},
  year={2022}
}

@inproceedings{zhang2023sadtalker,
  title={Sadtalker: Learning realistic 3d motion coefficients for stylized audio-driven single image talking face animation},
  author={Zhang, Wenxuan and Cun, Xiaodong and Wang, Xuan and Zhang, Yong and Shen, Xi and Guo, Yu and Shan, Ying and Wang, Fei},
  booktitle=CVPR_IEEE_CVF,
  pages={8652--8661},
  year={2023}
}

@article{yan2024df40,
  title={DF40: Toward Next-Generation Deepfake Detection},
  author={Yan, Zhiyuan and Yao, Taiping and Chen, Shen and Zhao, Yandan and Fu, Xinghe and Zhu, Junwei and Luo, Donghao and Yuan, Li and Wang, Chengjie and Ding, Shouhong and others},
  journal={arXiv preprint arXiv:2406.13495},
  year={2024}
}

@inproceedings{rosberg2023facedancer,
  title={Facedancer: Pose-and occlusion-aware high fidelity face swapping},
  author={Rosberg, Felix and Aksoy, Eren Erdal and Alonso-Fernandez, Fernando and Englund, Cristofer},
  booktitle=WACV,
  pages={3454--3463},
  year={2023}
}

@inproceedings{chen2020simswap,
  title={Simswap: An efficient framework for high fidelity face swapping},
  author={Chen, Renwang and Chen, Xuanhong and Ni, Bingbing and Ge, Yanhao},
  booktitle=ACMMM,
  pages={2003--2011},
  year={2020}
}

@inproceedings{siarohin2019first,
  title={First order motion model for image animation},
  author={Siarohin, Aliaksandr and Lathuili{\`e}re, St{\'e}phane and Tulyakov, Sergey and Ricci, Elisa and Sebe, Nicu},
  booktitle=NIPS,
  volume={32},
  year={2019}
}

@article{luo2023beyond,
  title={Beyond the prior forgery knowledge: Mining critical clues for general face forgery detection},
  author={Luo, Anwei and Kong, Chenqi and Huang, Jiwu and Hu, Yongjian and Kang, Xiangui and Kot, Alex C},
  journal=TIFS,
  volume={19},
  pages={1168--1182},
  year={2023},
  publisher={IEEE}
}

@misc{InSwapper,
  title        = {In{Swapper}},
  howpublished = {\url{https://github.com/haofanwang/inswapper}},
}

@inproceedings{xu2022designing,
  title={Designing one unified framework for high-fidelity face reenactment and swapping},
  author={Xu, Chao and Zhang, Jiangning and Han, Yue and Tian, Guanzhong and Zeng, Xianfang and Tai, Ying and Wang, Yabiao and Wang, Chengjie and Liu, Yong},
  booktitle=ECCV,
  pages={54--71},
  year={2022},
  organization={Springer}
}

@InProceedings{Liu_2024_CVPR,
    author    = {Liu, Ajian and Xue, Shuai and Gan, Jianwen and Wan, Jun and Liang, Yanyan and Deng, Jiankang and Escalera, Sergio and Lei, Zhen},
    title     = {CFPL-FAS: Class Free Prompt Learning for Generalizable Face Anti-spoofing},
    booktitle = CVPR_IEEE_CVF,
    month     = {June},
    year      = {2024},
    pages     = {222-232}
}

@inproceedings{yan2024transcending,
  title={Transcending forgery specificity with latent space augmentation for generalizable deepfake detection},
  author={Yan, Zhiyuan and Luo, Yuhao and Lyu, Siwei and Liu, Qingshan and Wu, Baoyuan},
  booktitle=CVPR_IEEE_CVF,
  pages={8984--8994},
  year={2024}
}

@inproceedings{liu2024forgery,
  title={Forgery-aware adaptive transformer for generalizable synthetic image detection},
  author={Liu, Huan and Tan, Zichang and Tan, Chuangchuang and Wei, Yunchao and Wang, Jingdong and Zhao, Yao},
  booktitle=CVPR_IEEE_CVF,
  pages={10770--10780},
  year={2024}
}

@article{dolhansky1910deepfake,
  title={The deepfake detection challenge (dfdc) preview dataset.},
  author={Dolhansky, B and Howes, R and Pflaum, B and Baram, N and Ferrer, CC},
  journal={arXiv preprint arXiv:1910.08854},
    year={2019}
}

@inproceedings{zi2020wilddeepfake,
  title={Wilddeepfake: A challenging real-world dataset for deepfake detection},
  author={Zi, Bojia and Chang, Minghao and Chen, Jingjing and Ma, Xingjun and Jiang, Yu-Gang},
  booktitle=ACMMM,
  pages={2382--2390},
  year={2020}
}

@misc{Deepfake21,
    author={Google, JigSaw},
    title = {Deep Fake Detection Dataset},
    url = {https://research.google/blog/contributing-data-to-deepfake-detection-research/},
    urldate      = {2021-11-13},
    year={2021}

}

@inproceedings{han2025towards,
  title={Towards More General Video-based Deepfake Detection through Facial Component Guided Adaptation for Foundation Model},
  author={Han, Yue-Hua and Huang, Tai-Ming and Hua, Kai-Lung and Chen, Jun-Cheng},
  booktitle=CVPR_IEEE_CVF,
  pages={22995--23005},
  year={2025}
}

@inproceedings{yan2025generalizing,
  title={Generalizing deepfake video detection with plug-and-play: Video-level blending and spatiotemporal adapter tuning},
  author={Yan, Zhiyuan and Zhao, Yandan and Chen, Shen and Guo, Mingyi and Fu, Xinghe and Yao, Taiping and Ding, Shouhong and Wu, Yunsheng and Yuan, Li},
  booktitle=CVPR_IEEE_CVF,
  pages={12615--12625},
  year={2025}
}

@inproceedings{guo2025face,
  title={Face Forgery Video Detection via Temporal Forgery Cue Unraveling},
  author={Guo, Zonghui and Liu, Yingjie and Zhang, Jie and Zheng, Haiyong and Shan, Shiguang},
  booktitle=CVPR_IEEE_CVF,
  pages={7396--7405},
  year={2025}
}

@inproceedings{yan2023deepfakebench,
    title={Deepfakebench: A comprehensive benchmark of deepfake detection},
  author={Yan, Zhiyuan and Zhang, Yong and Yuan, Xinhang and Lyu, Siwei and Wu, Baoyuan},
  booktitle=NIPS,
  year={2023}
}

@inproceedings{haliassos2021lips,
  title={Lips don't lie: A generalisable and robust approach to face forgery detection},
  author={Haliassos, Alexandros and Vougioukas, Konstantinos and Petridis, Stavros and Pantic, Maja},
  booktitle=CVPR_IEEE_CVF,
  pages={5039--5049},
  year={2021}
}

@inproceedings{haliassos2022leveraging,
  title={Leveraging real talking faces via self-supervision for robust forgery detection},
  author={Haliassos, Alexandros and Mira, Rodrigo and Petridis, Stavros and Pantic, Maja},
  booktitle=CVPR_IEEE_CVF,
  pages={14950--14962},
  year={2022}
}

@inproceedings{li2022blip,
      title={BLIP: Bootstrapping Language-Image Pre-training for Unified Vision-Language Understanding and Generation}, 
      author={Junnan Li and Dongxu Li and Caiming Xiong and Steven Hoi},
      year={2022},
      booktitle={ICML},
}

@INPROCEEDINGS{10377550,
  author={Zhai, Xiaohua and Mustafa, Basil and Kolesnikov, Alexander and Beyer, Lucas},
  booktitle={ICCV}, 
  title={Sigmoid Loss for Language Image Pre-Training}, 
  year={2023},
  volume={},
  number={},
  pages={11941-11952},
  doi={10.1109/ICCV51070.2023.01100}}

@article{liu2021video,
  title={Video Swin Transformer},
  author={Liu, Ze and Ning, Jia and Cao, Yue and Wei, Yixuan and Zhang, Zheng and Lin, Stephen and Hu, Han},
  journal={arXiv preprint arXiv:2106.13230},
  year={2021}
}

@inproceedings{tong2022videomae,
  title={Video{MAE}: Masked Autoencoders are Data-Efficient Learners for Self-Supervised Video Pre-Training},
  author={Zhan Tong and Yibing Song and Jue Wang and Limin Wang},
  booktitle={Advances in Neural Information Processing Systems},
  year={2022}
}

@inproceedings{liu2022umt,
  title={UMT: Unified Multi-modal Transformers for Joint Video Moment Retrieval and Highlight Detection},
  author={Liu, Ye and Li, Siyuan and Wu, Yang and Chen, Chang Wen and Shan, Ying and Qie, Xiaohu},
  booktitle={Proceedings of the IEEE/CVF Conference on Computer Vision and Pattern Recognition (CVPR)},
  pages={3042--3051},
  year={2022}
}

@misc{li2022uniformerv2,
      title={UniFormerV2: Spatiotemporal Learning by Arming Image ViTs with Video UniFormer}, 
      author={Kunchang Li and Yali Wang and Yinan He and Yizhuo Li and Yi Wang and Limin Wang and Yu Qiao},
      year={2022},
      eprint={2211.09552},
      archivePrefix={arXiv},
      primaryClass={cs.CV}
}

@inproceedings{ M2F2_Det_xiao,
  author = { Xiao Guo and Xiufeng Song and Yue Zhang and Xiaohong Liu and Xiaoming Liu },
  title = { Rethinking Vision-Language Model in Face Forensics: Multi-Modal Interpretable Forged Face Detector },
  booktitle = {Computer Vision and Pattern Recognition },
  year = { 2025 },
}

@inproceedings{laion5b,
author = {Schuhmann, Christoph and Beaumont, Romain and Vencu, Richard and Gordon, Cade and Wightman, Ross and Cherti, Mehdi and Coombes, Theo and Katta, Aarush and Mullis, Clayton and Wortsman, Mitchell and Schramowski, Patrick and Kundurthy, Srivatsa and Crowson, Katherine and Schmidt, Ludwig and Kaczmarczyk, Robert and Jitsev, Jenia},
title = {LAION-5B: an open large-scale dataset for training next generation image-text models},
year = {2022},
isbn = {9781713871088},
publisher = {Curran Associates Inc.},
address = {Red Hook, NY, USA},
booktitle = NIPS,
articleno = {1833},
numpages = {17},
location = {New Orleans, LA, USA},
series = {NIPS '22}
}

@inproceedings{cui2025forensics,
  title={Forensics adapter: Adapting clip for generalizable face forgery detection},
  author={Cui, Xinjie and Li, Yuezun and Luo, Ao and Zhou, Jiaran and Dong, Junyu},
  booktitle=CVPR_IEEE_CVF,
  pages={19207--19217},
  year={2025}
}

@inproceedings{lin2025standing,
  title={Standing on the shoulders of giants: Reprogramming visual-language model for general deepfake detection},
  author={Lin, Kaiqing and Lin, Yuzhen and Li, Weixiang and Yao, Taiping and Li, Bin},
  booktitle=AAAI,
  volume={39},
  number={5},
  pages={5262--5270},
  year={2025}
}

@inproceedings{kundu2025towards,
  title={Towards a universal synthetic video detector: From face or background manipulations to fully ai-generated content},
  author={Kundu, Rohit and Xiong, Hao and Mohanty, Vishal and Balachandran, Athula and Roy-Chowdhury, Amit K},
  booktitle=CVPR_IEEE_CVF,
  pages={28050--28060},
  year={2025}
}

@inproceedings{zhang2024common,
  title={Common sense reasoning for deepfake detection},
  author={Zhang, Yue and Colman, Ben and Guo, Xiao and Shahriyari, Ali and Bharaj, Gaurav},
  booktitle=ECCV,
  pages={399--415},
  year={2024},
  organization={Springer}
}

@String(CVPR_IEEE_CVF  = {CVPR})

@String(CVPR_IEEE  = {CVPR})

@String(CVPR  = {CVPR})

@String(ICCV_IEEE_CVF  = {ICCV})

@String(ICCV_IEEE  = {ICCV})

@String(ICCV  = {ICCV})

@String(ECCV  = {ECCV})

@String(NIPS  = {NeurIPS})

@String(ICML  = {ICML})

@String(WACV  = {WACV})

@String(FG  = {FG})

@String(ICLR  = {ICLR})

@String(PR = {Pattern Recognit.})

@article{kay2017kinetics,
  title={The kinetics human action video dataset},
  author={Kay, Will and Carreira, Joao and Simonyan, Karen and Zhang, Brian and Hillier, Chloe and Vijayanarasimhan, Sudheendra and Viola, Fabio and Green, Tim and Back, Trevor and Natsev, Paul and others},
  journal={arXiv preprint arXiv:1705.06950},
  year={2017}
}

@INPROCEEDINGS{UADFV,
  author={Yang, Xin and Li, Yuezun and Lyu, Siwei},
  booktitle=ICASSP, 
  title={Exposing Deep Fakes Using Inconsistent Head Poses}, 
  year={2019},
  volume={},
  number={},
  pages={8261-8265},
  doi={10.1109/ICASSP.2019.8683164}}

@InProceedings{Lin_2025_CVPR,
    author    = {Lin, Li and Santosh, Santosh and Wu, Mingyang and Wang, Xin and Hu, Shu},
    title     = {AI-Face: A Million-Scale Demographically Annotated AI-Generated Face Dataset and Fairness Benchmark},
    booktitle = CVPR_IEEE_CVF,
    month     = {June},
    year      = {2025},
    pages     = {3503-3515}
}

@article{Massey1951TheKT,
  title={The Kolmogorov-Smirnov Test for Goodness of Fit},
  author={Frank J. Massey},
  journal={Journal of the American Statistical Association},
  year={1951},
  volume={46},
  pages={68-78},
}

@inproceedings{FSGAN,
  title={{FSGAN}: Subject agnostic face swapping and reenactment},
  author={Nirkin, Yuval and Keller, Yosi and Hassner, Tal},
  booktitle=ICCV_IEEE_CVF,
  pages={7184--7193},
  year={2019}
}

@inproceedings{
LIA,
title={Latent Image Animator: Learning to Animate Images via Latent Space Navigation},
author={Yaohui Wang and Di Yang and Francois Bremond and Antitza Dantcheva},
booktitle=ICLR,
year={2022},
}


\begin{IEEEbiography}[{\includegraphics[width=0.6in,height=0.75in,clip,keepaspectratio]{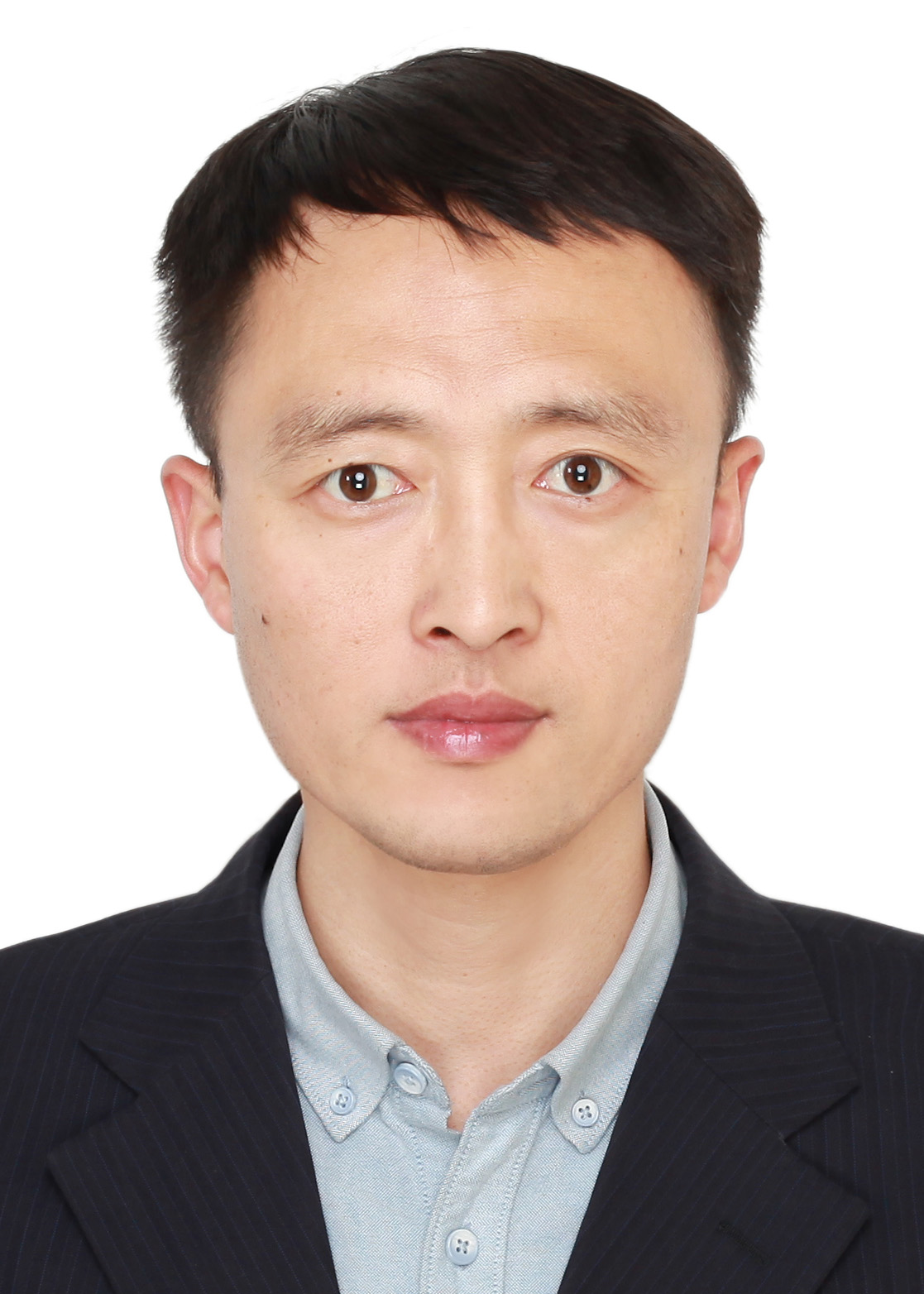}}]{Zonghui Guo} (Member, IEEE) received Ph.D. degree in intelligence information and communication systems from the Ocean University of China, Qingdao, China, in 2021. From 2021 to 2024, he was a Postdoctoral Researcher at the Institute of Computing Technology, Chinese Academy of Sciences (CAS), Beijing, China. In 2025, he joined the Ocean University of China, where he is currently an Associate Professor. His research interests include face forgery detection, low-level vision 
applications such as image/video harmonization and image/video enhancement.
\end{IEEEbiography}

\begin{IEEEbiography}[{\includegraphics[width=0.6in,height=0.75in,clip,keepaspectratio]{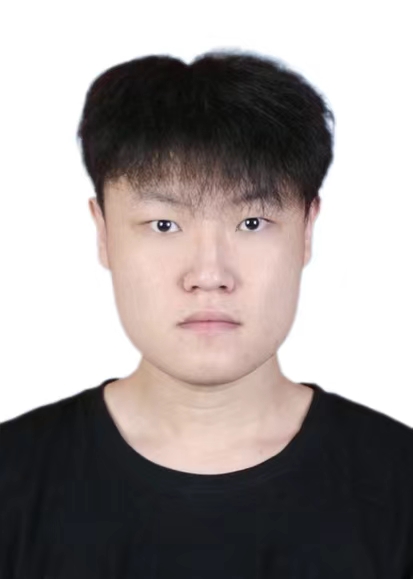}}]{Yingjie Liu} (Student Member, IEEE) received the B.E. degree in information engineering from the Qingdao University of Science and Technology, Qingdao, China, in 2022. He is currently a Postgraduate Student with the College of Electronic Engineering, Ocean University of China, Qingdao, China. His research interests include face forgery detection and presentation attack detection.
\end{IEEEbiography}

\begin{IEEEbiography}[{\includegraphics[width=0.6in,height=0.75in,clip,keepaspectratio]{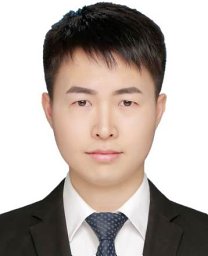}}]{Jie Zhang}
(Member, IEEE) is an associate professor with the Institute of Computing Technology, Chinese Academy of Sciences (CAS). He received the Ph.D. degree from the University of Chinese Academy of Sciences, Beijing, China. His research interests cover computer vision, pattern recognition, machine learning, particularly include face recognition, image segmentation, weakly/semi-supervised learning, domain generalization.
\end{IEEEbiography}

\begin{IEEEbiography}[{\includegraphics[width=0.6in,height=0.75in,clip,keepaspectratio]{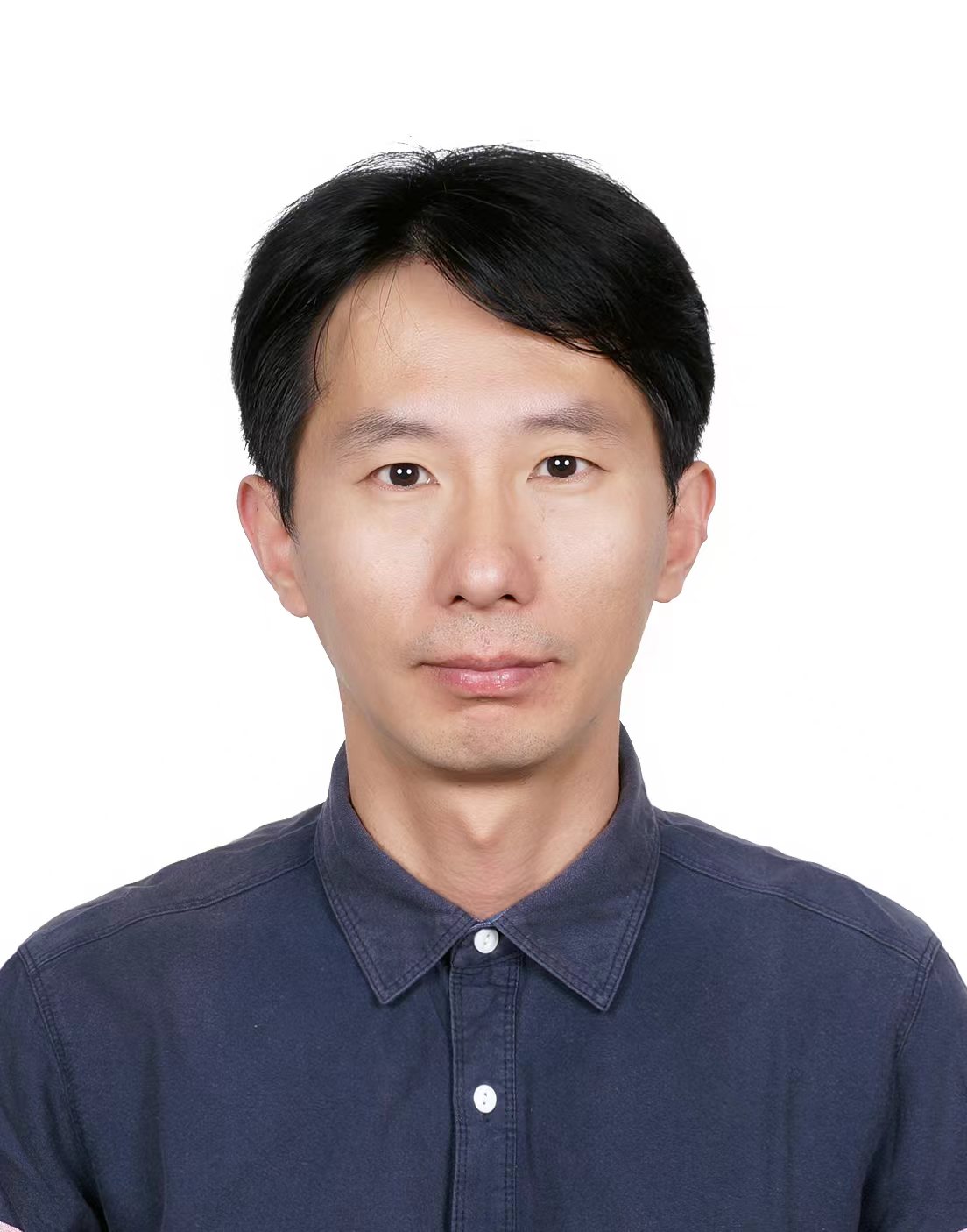}}]{Haiyong Zheng}
(Senior~Member, IEEE) received the B.E. degree in electronic information engineering and the Ph.D. degree in ocean information sensing and processing from the Ocean University of China, Qingdao, China, in 2004 and 2009, respectively. In 2009, he joined the College of Electronic Engineering, Ocean University of China, where he is currently a Professor. His research interests include computer vision, underwater vision, and deep learning.
\end{IEEEbiography}

\begin{IEEEbiography}[{\includegraphics[width=0.6in,height=0.75in,clip,keepaspectratio]{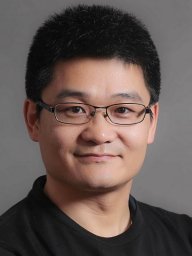}}]{Shiguang Shan}
(Fellow, IEEE) received Ph.D. degree in computer science from the Institute of Computing Technology (ICT), Chinese Academy of Sciences (CAS), Beijing, China, in 2004. He has been a full professor of this institute since 2010 and now the director of CAS Key Lab of Intelligent Information Processing. His research interests cover computer vision, pattern recognition, and machine learning. He has published more than 300 papers, with totally more than 38,000 Google scholar citations. He has served as Area Chair (or Senior PC) for many international conferences including ICCV11, ICPR12/14/20, ACCV12/16/18, FG13/18, ICASSP14, BTAS18, AAAI20/21, IJCAI21, and CVPR19/20/21. And he was/is Associate Editors of several journals including IEEE T-IP, Neurocomputing, CVIU, and PRL. He was a recipient of the China's State Natural Science Award in 2015, and the China's State S\&T Progress Award in 2005 for his research work.
\end{IEEEbiography}

\vfill

\end{document}


\title{Revisiting Face Forgery Detection: From Facial Representation to Forgery Detection (Supplementary Material)}
\author{Zonghui~Guo,~\IEEEmembership{Member,~IEEE},
        Yingjie~Liu,~\IEEEmembership{Student Member,~IEEE},
        Jie~Zhang,~\IEEEmembership{Member,~IEEE},\\
        Haiyong~Zheng,~\IEEEmembership{Senior Member,~IEEE},
        and~Shiguang~Shan,~\IEEEmembership{Fellow,~IEEE}


}

\markboth{Journal of \LaTeX\ Class Files,~Vol.~14, No.~8, August~2015}%
{Shell \MakeLowercase{\textit{et al.}}: Bare Demo of IEEEtran.cls for Computer Society Journals}

\maketitle

\section{Decorrelation Constraint with Different Tokens}



Our method utilizes $[\texttt{CLS}]$ token as classification head input and minimizes correlation between $[\texttt{CLS}]$ tokens of main and auxiliary branches. Actually, spatial patch tokens (Patches) in Transformers can also be subjected to decorrelation constraints. We conduct $4$ comparative experiments to verify effectiveness of minimizing correlation constrained on $[\texttt{CLS}]$ token or all tokens ($[\texttt{CLS}]$ \& Patches), as shown in Table~\ref{tab:ablation_decorr}. Results demonstrate that performance remains stable when constraints are applied directly to classification head input (Row 1 vs. Row 4). Conversely, inconsistent constraints slightly degrade performance (Rows 1 vs. 2 and 3). Specifically, applying decorrelation to $[\texttt{CLS}]$ token achieves highest average performance of $88.14$\% AUC.

\begin{table}[htbp]
    \centering
    \caption{Ablation study on decorrelation targets. Performance (AUC \%) is compared across token combinations.}
    \label{tab:ablation_decorr}
    \resizebox{\columnwidth}{!}{
    \footnotesize 
    \setlength{\tabcolsep}{1.2mm} 
    \renewcommand{\arraystretch}{1.1} 
    \begin{tabular}{llcccc}
        \toprule
        Head Input & Decorrelation Target & Celeb-DF & DFDC & FFIW & Avg. \\
        \midrule
        $[\texttt{CLS}]$ & $[\texttt{CLS}]$ (Ours) & 90.31 & \textbf{84.89} & \textbf{89.21} & \textbf{88.14} \\
        $[\texttt{CLS}]$ & Patches & \textbf{90.80} & 84.60 & 88.40 & 87.93 \\
        $[\texttt{CLS}]$ & $[\texttt{CLS}]$ \& Patches & 90.36 & 84.41 & 88.45 & 87.74 \\
        \midrule
        Avg. Tokens & $[\texttt{CLS}]$ \& Patches & 90.34 & 84.75 & 89.14 & 88.08 \\
        \bottomrule
    \end{tabular}
    }
\end{table}

Moreover, these experimental results indicate that multi-layer self-attention in ViTs causes spatial patch tokens and $[\texttt{CLS}]$ token to become highly correlated in final layer. Therefore, minimizing correlation between input features of main and auxiliary classification heads leads to similar performance, whether using $[\texttt{CLS}]$ token or all tokens.


\section{Further Analysis of Threshold Optimization Mechanism}

\subsection{Analysis of \texorpdfstring{$p' = p/u$}{p' = p/u} Normalization}
Since $p, u \in [0, 1)$ allows $p'$ to exceed 1, we apply $\tanh$ function to map scores into $[0, 1)$, managing large values without hard clipping as follows:
\begin{equation}
    p' = \tanh\left(\frac{\lambda}{2} \cdot \frac{p}{u+\epsilon}\right),
\label{eq:predict}
\end{equation}
where $\epsilon = 10^{-7}$ avoids division by zero, and scaling factor $\lambda = 0.02$ is approximated by $2.0/m$, with $m$ being median of raw ratio $r = p/(u+\epsilon)$, to align score distribution with near-linear region of $\tanh$ function and prevent score saturation. Indeed, choice of $\lambda$ and normalization function do not alter relative ranking of samples based on $p/(u+\epsilon)$, nor do they affect search for optimal threshold $\tau_{ot}^{UTOM}$ to achieve maximum accuracy. Once raw ratios are obtained, applying consistent $\lambda$ and normalization function naturally ensures that effectiveness of $\tau_{ot}^{UTOM}$ remains preserved. 

\subsection{Analysis of Cross-Domain Distribution Shift}

\begin{table}[htbp]
    \centering
    \caption{Cross-dataset transfer performance analysis using KS distance to measure the similarity of prediction and uncertainty distributions between source and target domains.}
    \label{tab:ks_transfer}
    \resizebox{\columnwidth}{!}{
    \begin{tabular}{l cc ccc}
        \toprule
        \textbf{Transfer} & \multicolumn{2}{c}{\textbf{KS Distance}} & \multicolumn{3}{c}{\textbf{Target Acc. (\%)}} \\
        \cmidrule(r){2-3} \cmidrule(l){4-6}
        \textbf{Path} & Prob. & Uncert. & $\tau_{0.5}$ & $\tau_{ot}$ & $\tau_{ot}^{UTOM}$ \\
        \midrule
        \multicolumn{6}{l}{\textit{Group A: Highly Similar ($KS < 0.15$)}} \\
        DFDC $\rightarrow$ FFIW & 0.10 & 0.07 & 81.20 & 82.20\scriptsize{(+0.99)} & \textbf{84.00}\scriptsize{(+2.80)} \\
        FFIW $\rightarrow$ DFDC & 0.10 & 0.07 & 76.00 & 76.76\scriptsize{(+0.76)} & \textbf{78.30}\scriptsize{(+2.30)} \\
        \midrule
        \multicolumn{6}{l}{\textit{Group B: Moderately Similar ($0.15 \le KS < 0.22$)}} \\
        CDF $\rightarrow$ FFIW & 0.19 & 0.19 & 81.20 & 81.40\scriptsize{(+0.20)} & \textbf{83.80}\scriptsize{(+2.60)} \\
        FFIW $\rightarrow$ CDF & 0.19 & 0.19 & 83.20 & 82.63\scriptsize{(-0.57)} & \textbf{83.40}\scriptsize{(+0.19)} \\
        \midrule
        \multicolumn{6}{l}{\textit{Group C: Dissimilar ($KS \ge 0.22$)}} \\
        CDF $\rightarrow$ DFDC & 0.26 & 0.11 & 76.00 & 75.94\scriptsize{(-0.06)} & \textbf{77.54}\scriptsize{(+1.54)} \\
        DFDC $\rightarrow$ CDF & 0.26 & 0.11 & 83.20 & 81.66\scriptsize{(-1.54)} & \textbf{84.75}\scriptsize{(+1.55)} \\
        \bottomrule
    \end{tabular}
    }
\end{table}

Since distribution shifts across datasets and manipulation approaches affect performance of both OT and our UTOM, with such shifts standing as critical bottleneck and fundamental challenge for generalization of FFD task. To investigate impact of these distribution shifts, we use Kolmogorov-Smirnov (KS) distance~\cite{Massey1951TheKT} to quantify gap between respective distributions of probability $p$ and uncertainty $u$ across Celeb-DF, DFDC, and FFIW. We then evaluate performance of three different thresholds ($\tau_{0.5}$, $\tau_{ot}$, and $\tau_{ot}^{UTOM}$) based on relative distance between these distributions, as shown in Table~\ref{tab:ks_transfer}. Experimental results demonstrate that as discrepancy in prediction distributions increases, $\tau_{ot}$ leads to decrease in accuracy compared to $\tau_{0.5}$ baseline, whereas our $\tau_{ot}^{UTOM}$ consistently outperforms $\tau_{0.5}$ threshold, confirming that UTOM with uncertainty achieves more stable performance under varying distribution shifts.

\subsection{Analysis of Class Imbalance in Source Datasets}

\begin{figure*}[t]
  \centering
  \includegraphics[width=1.0\linewidth]{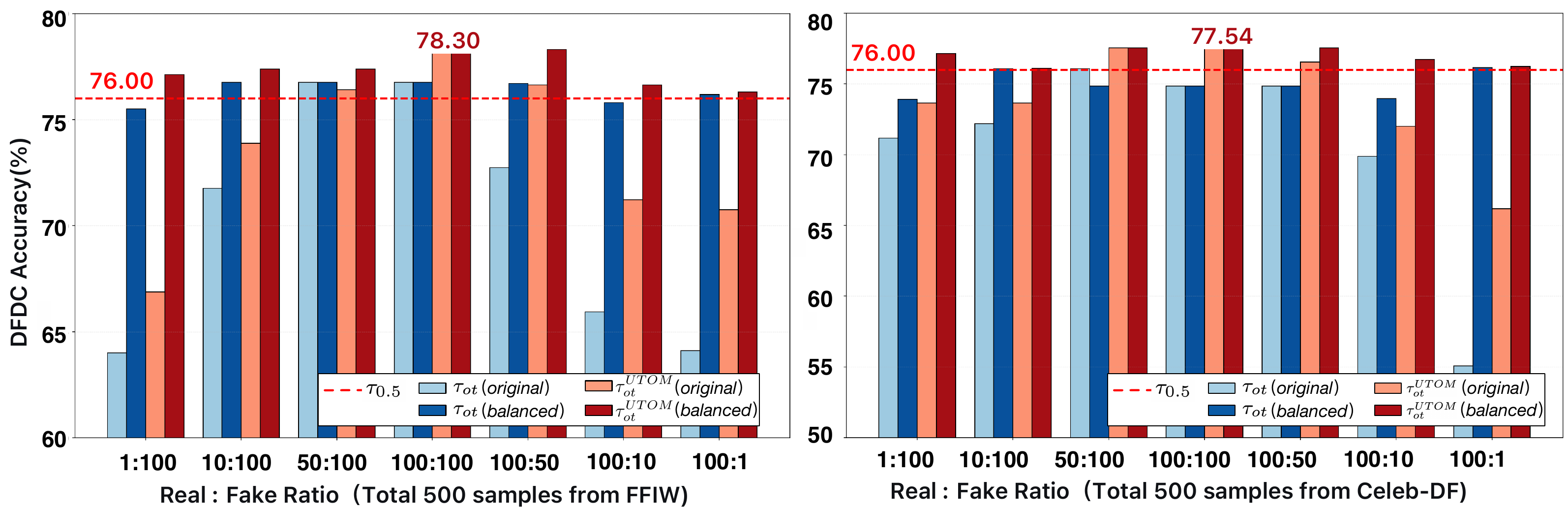}
  \caption{Impact of class imbalance on threshold optimization. Performance of $\tau_{ot}$ and $\tau_{ot}^{UTOM}$ is evaluated under various real-to-fake ratios on FFIW (left) and Celeb-DF (right). Here, ``original'' denotes thresholds calculated using original imbalanced ratios, while ``balanced'' represents results after re-sampling minority class to achieve balanced source classes.}
  \label{fig:imbalanced}
\end{figure*}

Actually, OT and our UTOM calculate optimal thresholds for source datasets, which is affected by severe imbalance between real and fake samples. Accordingly, we simulate impact of imbalance on target dataset (DFDC) accuracy by sampling Celeb-DF and FFIW respectively with different class ratios to evaluate three different thresholds ($\tau_{0.5}$, $\tau_{ot}$, and $\tau_{ot}^{UTOM}$), as shown in Figure~\ref{fig:imbalanced}.

Experimental results demonstrate that when class ratios reach $10:100$ and $1:100$ (or $100:10$ and $100:1$), performance of $\tau_{ot}$ and $\tau_{ot}^{UTOM}$ is significantly impacted, with accuracy falling below $\tau_{0.5}$. However, once ratio reaches $50:100$ (or $100:50$), both exceed $\tau_{0.5}$, proving that severe class imbalance in source dataset weakens capability of $\tau_{ot}$ and $\tau_{ot}^{UTOM}$. Furthermore, when we apply re-sampling to minority class to achieve balance, both $\tau_{ot}$ and $\tau_{ot}^{UTOM}$ improve substantially, with our $\tau_{ot}^{UTOM}$ specifically consistently outperforming $\tau_{0.5}$. Therefore, class balance in source dataset should be maintained as much as possible when using $\tau_{ot}^{UTOM}$. If source dataset exhibits extreme imbalance between real and fake samples, applying re-sampling to minority class before calculating $\tau_{ot}^{UTOM}$ serves as effective solution.

\subsection{Analysis of Failure Cases}
\begin{figure*}[t]
  \centering
  \includegraphics[width=1.0\linewidth]{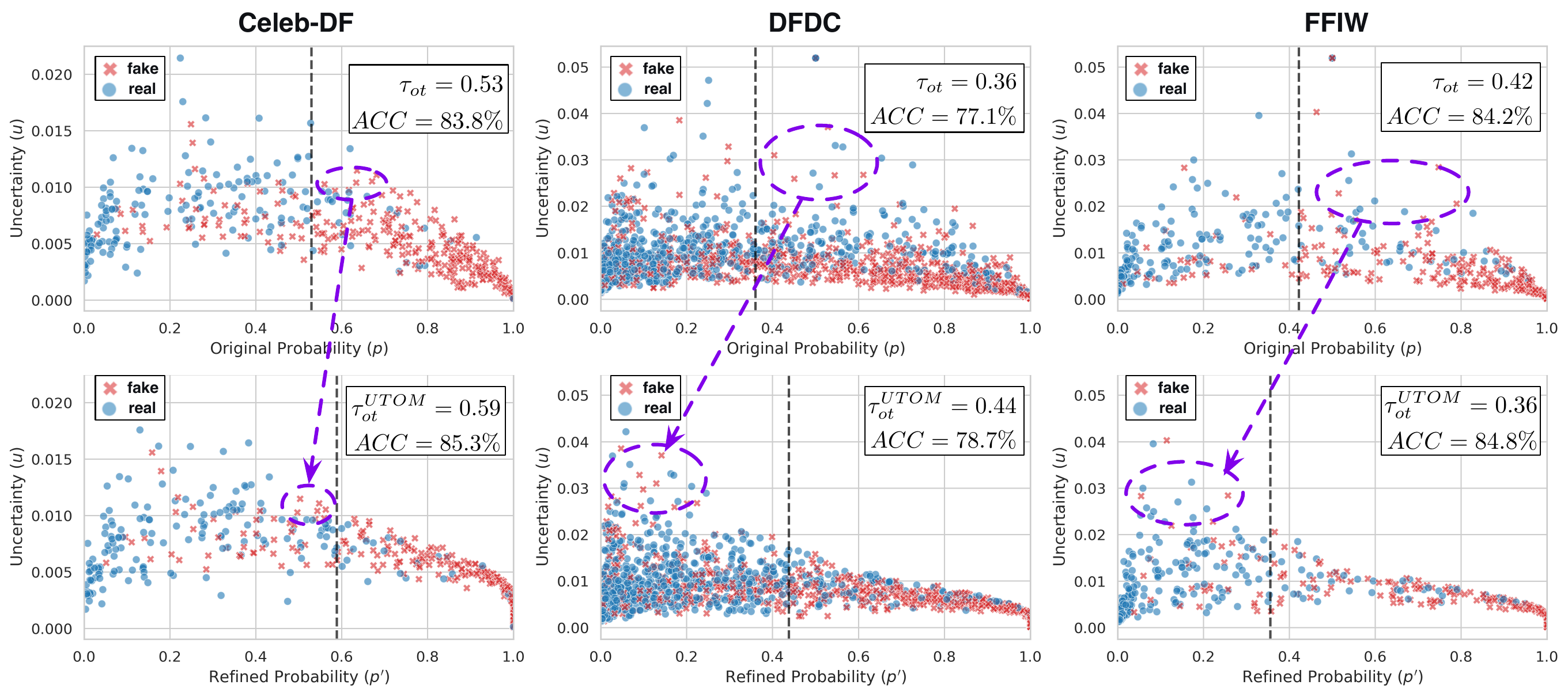}
  \caption{Joint distribution of prediction $p$ and uncertainty $u$ on Celeb-DF, DFDC, and FFIW datasets. Top-row shows original distributions while bottom-row illustrates refined probability $p'$ adjusted via Equation~\ref{eq:predict}. Purple elliptical regions show failure cases where high uncertainty $u$ leads to leftward shift and subsequent misclassification of fake samples as real.}
  \label{fig:fcase}
\end{figure*}
We analyze changes in prediction probability $p'$ after applying Equation~\ref{eq:predict} of UTOM through the joint distribution of $p$ and $u$. Figure~\ref{fig:fcase} presents results on Celeb-DF, DFDC, and FFIW. Ideally, real samples should cluster near bottom-left corner, while fake samples should cluster near bottom-right. As shown in top-row, majority of samples align with this ideal case. However, some samples yield prediction probabilities between $0.4$ and $0.6$ with high uncertainty, which is consistent with expectations, as samples difficult to predict typically exhibit higher uncertainty.

As shown in bottom-row, refined probabilities $p'$ calculated via Equation.~\ref{eq:predict} achieve higher accuracy and effectively shift toward respective ends of distribution, \ie, real (left) and fake (right). Nevertheless, certain failure cases emerge within refined scores. As indicated by red crosses within purple elliptical regions, some fake samples originally predicted as fake in top-row shift leftward and are misclassified as real in bottom-row.  Such shifts are driven by large uncertainty $u$, where dividing original probability $p$ by high $u$ values suppresses final score $p'$ below decision threshold. These cases do not negate the superior correct judgment of our method for the majority of samples, and such issues are expected to be mitigated by further enhancing the discriminative generalization of FFD models in future work.

\bibliographystyle{IEEEtran}

\bibliography{deepfake.bib}

\vfill